\documentclass[preprint,3p,times,twocolumn,authoryear,number]{elsarticle}

\usepackage{amssymb}
\usepackage{amsmath}
\usepackage{multicol}
\usepackage{multirow}
\usepackage{tabularx}
\usepackage{graphicx}
\usepackage{wrapfig}
\usepackage{enumitem}
\usepackage[table]{xcolor}%
\usepackage[utf8]{inputenc} %
\usepackage[T1]{fontenc}    %
\usepackage[colorlinks=true, allcolors=RoyalBlue]{hyperref}
\usepackage{url}            %
\usepackage{booktabs}       %
\usepackage{amsfonts}       %
\usepackage{nicefrac}       %
\usepackage{amsmath}
\usepackage{microtype}      %
\usepackage{xcolor}         %
\usepackage{pifont}
\usepackage{float}
\usepackage{algorithm}
\usepackage{algpseudocode}
\usepackage{adjustbox}  %

\definecolor{best_color}{HTML}{FB9999}
\definecolor{better_color}{HTML}{FDCC99}
\definecolor{good_color}{HTML}{FFF8AE}

\usepackage{subcaption}
\usepackage[capitalize,nameinlink]{cleveref}
\renewcommand{\Cref}[1]{\cref{#1}}

\newcommand{\editcolor}{black}

\newcommand{\cmark}{\ding{51}}%
\newcommand{\xmark}{\ding{55}}%
\newtheorem{theorem}{Theorem}
\newcommand{\method}{DIO}
\newcommand{\edit}[1]{{#1}}
\newcommand{\editv}[1]{{\textcolor{\editcolor}{#1}}}

\usepackage{lineno}

\makeatletter
\renewcommand{\paragraph}{%
  \@startsection{paragraph}{4}{\z@}%
  {3.25ex \@plus1ex \@minus.2ex}%
  {-1em}%
  {\normalfont\normalsize\bfseries}%
}
\makeatother

\journal{Medical Image Analysis}

\begin{document}

\begin{frontmatter}

\title{Deep Implicit Optimization enables Robust Learnable Features for Deformable Image Registration}
\author{Rohit Jena\corref{cor1}\fnref{cs,picsl}}
\ead{rjena@upenn.edu}
\author{Pratik Chuadhari\fnref{cs,ese}}
\author{James C. Gee\fnref{picsl,radio}}
\fntext[cs]{Computer and Information Science}
\fntext[picsl]{Penn Image Computing and Science Laboratory}
\fntext[ese]{Electrical and Systems Engineering}
\fntext[radio]{Radiology}
\cortext[cor1]{Corresponding author}
\affiliation{organization={University of Pennsylvania},
           addressline={},
           city={Philadelphia},
           postcode={19104},
           state={PA},
           country={USA}}
\begin{abstract}
Deep Learning in Image Registration (DLIR) methods have been tremendously successful in image registration due to their speed and ability to incorporate weak label supervision at training time.
However, existing DLIR methods forego many of the benefits and invariances of optimization methods.
The lack of a task-specific inductive bias in DLIR methods leads to suboptimal performance, especially in the presence of domain shift.
Our method aims to bridge this gap between statistical learning and optimization by explicitly incorporating optimization as a layer in a deep network.
A deep network is trained to predict multi-scale dense feature images that are registered using a black box iterative optimization solver.
This optimal warp is then used to minimize image and label alignment errors.
By \textit{implicitly} differentiating end-to-end through an iterative optimization solver, we \textit{explicitly} exploit invariances of the correspondence matching problem induced by the optimization, while learning registration and label-aware features, and guaranteeing the warp functions to be a local minima of the registration objective in the feature space.
Our framework shows excellent performance on in-domain datasets, and is agnostic to domain shift such as anisotropy and varying intensity profiles.
For the first time, our method allows switching between arbitrary transformation representations (free-form to diffeomorphic) at test time with zero retraining.
End-to-end feature learning also facilitates interpretability of features and arbitrary test-time regularization, which is not possible with existing DLIR methods.
\end{abstract}

\onecolumn

\twocolumn


\end{frontmatter}

\defcitealias{wang2023robust}{keymorph}

\section{Introduction}

The success of deep learning methods over the past decade has radically transformed various disciplines including computer vision, natural language processing, robotics and biomedical and biological sciences.
A lot of empirical evidence in the deep learning literature points to the fact that incorporating invariance or equivariance to the task at hand is a key factor for good model performance.
This explains why convolutional networks, for example, excel at tasks like image classification and segmentation even with little data by  exploiting the translation equivariance of the task.
Learning this inductive bias from scratch (for example, transformer architectures are not translation-equivariant) requires significantly more data (\cite{dosovitskiy2020image}) and compute (\cite{he2022masked}).
Even in the large data regime, the right inductive bias can lead to superior performance compared to a model without the inductive bias (\cite{liu2022convnet,woo2023convnext}).
Moreover, well-modeled inductive biases in the network design can lead to good generalization to unseen data, even when trained on purely synthetic data (\cite{fischer2015flownet,Yang_2021_CVPR}).

\paragraph{Motivation.}
Deformable Image Registration (DIR) pertains to the local, non-linear alignment of images by estimating a dense displacement field.
Deep Learning for Image Registration (DLIR) has emerged as a promising paradigm to use deep learning to directly predict a warp field that performs dense correspondence matching between images.
DLIR methods aim to resolve the limitations of traditional optimization-based methods by performing \textit{amortized optimization} and learning features to incorporate additional labelmap or keypoint overlap signals.
However, most existing DLIR architectures do not explicitly incorporate the invariances required for dense correspondence matching (more in ~\cref{sec:reg-invariance}), necessitating the use of vast amounts of real or synthetic data to learn these invariances (~\cite{hoffmann2021synthmorph}).
Since large amounts of data is hard to come by in medical imaging, existing DLIR methods exhibit brittle performance under minor domain shift.
Generalization to domain shift is imperative to biomedical and clinical imaging where volumes are acquired with different scanners, protocols, and resolutions, where the applicability of DLIR methods is limited to the training domain.

\edit{Moreover, the current paradigm of learning deep parameterized warp fields leads to a fixed warp representation and a lack of flexibility to switch between different warp representations at test time.}
Typical registration workflows require a practitioner to compare different parameterizations of the transformation (SVF~\cite{ashburner2007fast}, geodesic~\cite{niethammer2011geodesic}, LDDMM~\cite{beg2005computing}, B-Splines~\cite{tustison2013explicit}, or affine) to determine the representation most suitable for their downstream application and additional retraining of DLIR methods in this context becomes computationally prohibitive.
Hyperparameter tuning for regularization is also expensive for DLIR methods. 
Although recent methods propose conditional registration~\cite{hoopes2021hypermorph,mok2021conditional} to amortize over the regularization hyperparameter during training, the family of regularization is fixed in such cases, and the combinatorial nature of hyperparameter spaces exacerbates the complexity when considering multiple \edit{or unseen} regularizations. 

\edit{Finally, tasks like correspondence matching can benefit from an inductive bias that exploits the instrinsic error-correcting nature of optimization-based methods.}
\edit{
Although some prior attempts have emulated the flavor of iterative optimization using recurrent formulations~\cite{rwcnet,gradirn,suits}, they are limited in their expressive capacity due to the high memory and computational demands of storing the entire computational graph.  
}

\paragraph{Contributions.} \footnote{\editv{Source Code is available at \url{https://github.com/rohitrango/DIO}}}
The aforementioned limitations urge a departure from the prevailing parameterized warp field paradigm for deep learning in image registration.
\edit{Specifically, our goal is to \textit{synergistically} perform feature learning and optimization for image registration by enabling the ability to differentiate task-specific image and label matching objectives through \textit{arbitrary black box optimization-based solvers}.}
This retains the advantages inherent in classical optimization-based methods - namely, the flexibility of arbitrary warp field representations, invariance to resolutions of fixed and moving images, \edit{the intrinsic error-correcting nature of optimization-based methods, task-specific appearance invariance induced by the optimization objective, the} ability to integrate arbitrary regularization to the warp fields 
, and resilience to domain shifts.
To this end, we introduce \textit{\method}, a generic \textit{differentiable implicit optimization} layer integrated with learnable feature network for image registration. 
By explicitly decoupling feature learning and optimization, our framework bakes in additional apperance invariance and {incorporates weak supervisory signals like anatomical landmarks into the learned features} during training, improving the fidelity of the feature images for simultaneous image and landmark registration at inference. %
Feature learning also leads to \textit{dense} feature images, which smoothens the optimization landscape compared to intensity-based registration (\cref{sec:toyexample}) wherein intensity-level image heterogeneity hinders optimization in most medical imaging modalities. 
Since optimization frameworks are also \textit{discretization invariant} (agnostic to spatial resolutions and voxel sizes), {\method} is robust to domain shifts like varying anisotropy, difference in sizes of fixed and moving images, and different image acquisition and preprocessing protocols, even compared to models trained on contrast-agnostic synthetic data~\cite{hoffmann2021synthmorph}. 
Moreover, our framework allows \textit{zero-cost plug-and-play} of arbitrary transformation representations (free-form, geodesics, B-Spline, affine, etc.) and regularization at test time without additional training and loss of accuracy.
Furthermore, this paradigm for feature learning allows arbitrary regularization to be incorporated at test time, avoiding amortization costs for regularization at training time.

\section{Related Work}
\begin{figure*}[ht]
    \centering
    \begin{subfigure}[t]{\linewidth}
        \centering
        \includegraphics[width=\linewidth]{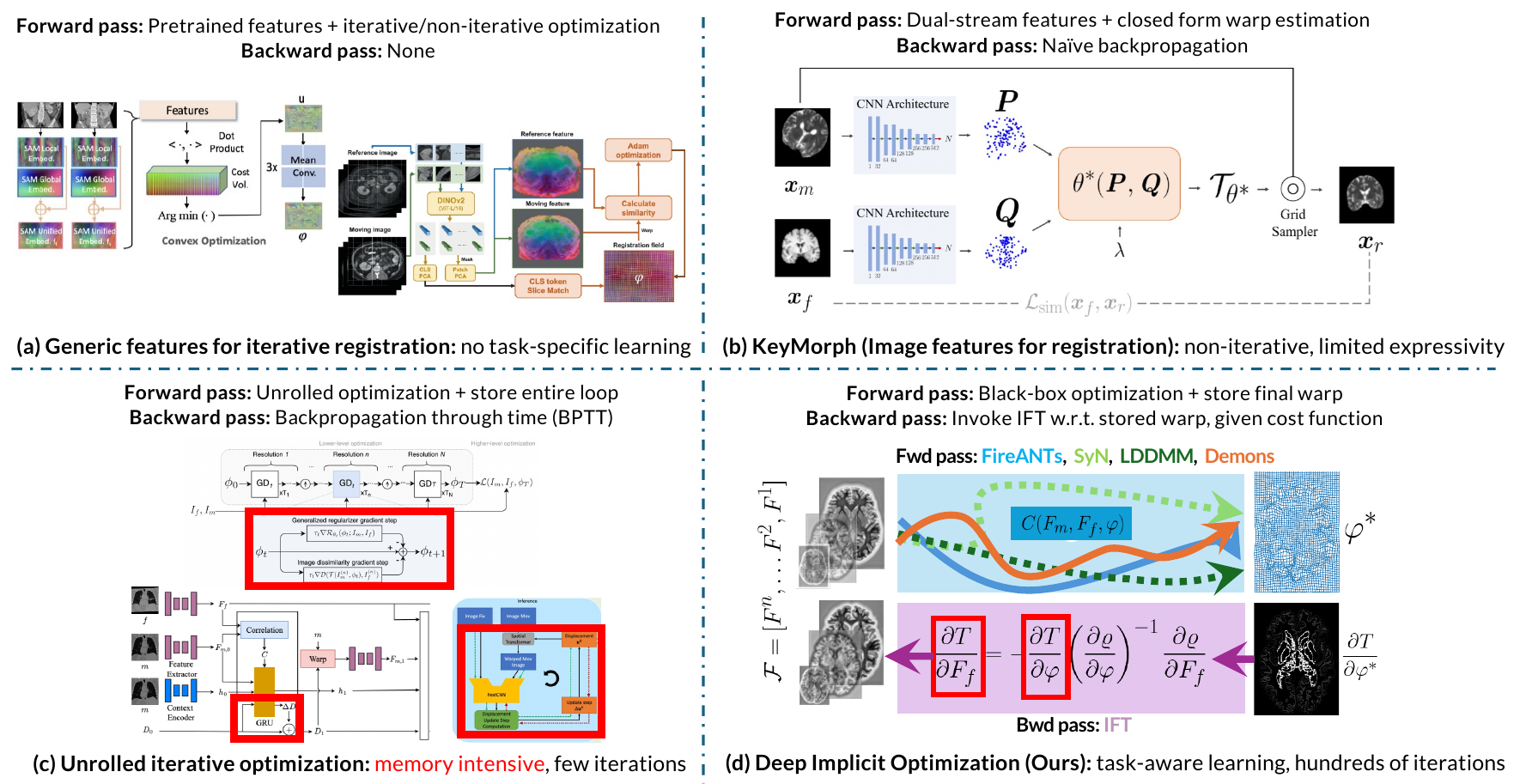}
    \end{subfigure}

    \caption{\textbf{An illustrative comparison of existing methods and our method}. \textbf{(a)} Generic features for image registration leverage the expressiveness and robustness of iterative optimization but do not incorporate task-specific learning, leading to suboptimal asymptotic performance on the in-distribution task. \textbf{(b)} Feature learning for closed-form parametric warp representations enable task-aware image features for registration, but are limited in expressiveness due to limited families of closed-form transforms and lack of error-correcting nature intrinsic to iterative optimization. \textbf{(c)} Unrolled iterative optimization using recurrent modules mimic the flavor of traditional optimization and enable task-aware image features. However, they are limited in expressivity because they can run only for a few number of iterations due to infeasible computational requirements. \textbf{(d)} {\method} (our method) synergizes the expressivity of advanced iterative solvers and task-aware image feature learning by defining a custom backward pass that does not require unrolling or iteration. {\method} provides the best of both worlds by inheriting the accuracy, expressivity, and robustness of iterative solvers, and asymptotic performance of learnable features.}
    \label{fig:cmp-overview}
\end{figure*}

\edit{
Existing methods have so far been limited in their ability to synergistically perform feature learning and iterative optimization.
This is primarily due to two reasons:
\begin{itemize}
    \item Backpropagation through the iterative optimization process requires storing the entire computation graph of the optimization process. For 3D images, each iteration stores a 3D warp field, which is infeasible due to its large memory footprint.
    \item Existing methods for image registration do not have the ability to backpropagate features from a generic iterative optimization-based solver to learnable, task-aware features of images
\end{itemize}
Here we discuss existing approaches that work around this gap between learning-based and optimization-based methods, and highlight the limitations of these methods.
An illustrative comparison is also presented in \cref{fig:cmp-overview}.
}

\subsection{Parametric learning-based methods for Image Registration} 
\edit{Deformable} Image Registration (DIR) refers to the alignment of a fixed image $I_f$ with a moving image $I_m$ using a transformation $\varphi \in T$, where $T$ is a family of transformations. 
Earliest deep learning for image registration (DLIR) methods like ~\cite{cao2017deformable,krebs2017robust,rohe2017svf,sokooti2017nonrigid} used supervised learning to predict the transformation $\varphi$.
~\cite{balakrishnan_voxelmorph_2019} was one of the first unsupervised method utilizing a UNet for unsupervised registration on neuroimaging data.
Since then, 
a variety of architectural innovations have emerged, including  ~\cite{chen_transmorph_2022-1,lebrat_corticalflow_2021,lku,mok2022affine} showing network design, ~\cite{zhao2019unsupervised,Zhao_2019_ICCV,joshi_diffeomorphic_nodate,de2019deep,mok_large_2020,zhang2021cascaded,qiu2021learning,chen2022unsupervised} using cascade-based architectures and loss functions, and ~\cite{mok2020fast,kim2021cyclemorph,kim2019unsupervised,tian2023gradicon,zhao2019unsupervised} formulating symmetric or inverse consistency-based formulations.
To address the challenge of dynamic hyperparameter incorporation into learning-based methods, ~\cite{mok2021conditional,hoopes2021hypermorph} inject the hyperparameter as input, and modulate the network to perform additional amortized optimization over different values of the hyperparameter. 
\edit{
~\cite{bigalke2023unsupervised} use a cyclical self-training framework and inverse-consistency loss to improve the performance of unsupervised registration methods.
}
However, most of these approaches are performant only in the training domain, and do not generalize to even small domain shifts as shown by ~\cite{hoffmann2021synthmorph,mok2022affine,neuralinstanceopt,jena2024deep,jian2024mamba}, limiting their applicability in clinical settings.
This is a fundamental prerequisite in biomedical imaging since different institutions follow varying acquisition and preprocessing pipelines, scanners from various manufacturers or different models.
To address this shortcoming, ~\cite{hoffmann2021synthmorph,uzunova2017training,perez2023learning,fu2020synthetic} adopt domain randomization and finetuning approaches to improve robustness of registration to domain shift. 
~\cite{tian2024unigradicon} propose a foundation model to improve registration accuracy.
Moreover, ~\cite{vxmpp,neuralinstanceopt,tian2024unigradicon} use instance optimization as a postprocessing step to improve performance of learning-based methods by refining the predicted warp field at inference time.
~\cite{wu_nodeo_2022,wolterink_implicit_nodate,joshi_diffeomorphic_nodate,hu_plug-and-play_2024} propose using implicit priors for deep learning within an optimization framework.
We refer the reader to ~\cite{gholipour2007brain,haskins_deep_2020,fu_deep_2020} for other detailed reviews.

\edit{
The limitation of these methods is their fixed parameterization of the warp field, which restrains their flexibility to switch between different warp field representations at test time.
Their end-to-end nature also impedes interpretability of these models. %
Moreover, we show that the learned features are not robust to domain shift, and require additional instance optimization to improve performance.
The inevitable necessity of instance optimization motivates our method to synergize feature learning and optimization end-to-end, instead of using instance optimization simply as a post-hoc step.
}

\subsection{Iterative methods for learning-based registration}
Learning-based registration methods typically predict \textit{imperfect} warp fields that need to be further tuned using an instance optimization step to improve registration quality(~\cite{balakrishnan2019voxelmorph,vxmpp,neuralinstanceopt,tian2024unigradicon}).
However, ~\cite{neuralinstanceopt} show marginal performance gains in instance optimization due to the high degrees of freedom and absence of robust initialization of deformations, urging a re-evaluation of the paradigm used for incorporating deep learning for registration.
Moreover, recent work by ~\cite{jian2024mamba} has shown that methods incorporating \textit{registration-aware} designs such as motion pyramids, correlation volumes, and iterative optimization significantly outperform prediction-based methods regardless of architecture.
Owing to the success of iterative optimization methods, recent DLIR solutions also propose emulating the iterative optimization within a network cascade. 
~\cite{Zhao_2019_ICCV,zhao2019unsupervised} use a cascade of networks to iteratively predict a warp field, and use the warped moving image as the input to the next layer in the cascade.
~\cite{chen2022unsupervised} uses a recurrent transformer network to predict a time-dependent velocity field.
~\cite{zhang2021cascaded} use a shared weights encoder to output feature images at multiple scales, and a deformation field estimator utilizing a correlation layer.
~\cite{teed_raft_2020} similarly build a 4D correlation volume from two 2D feature maps, and update the optical flow field using a recurrent unit that performs lookup on the correlation volume.
\edit{
In the context of medical image registration, ~\cite{gradirn} use an unrolled multi-resolution gradient-based energy optimization in its forward pass, which explicitly enforces image dissimilarity minimization in its update steps.
~\cite{rwcnet} use a recurrent network with a cost volume to emulate an iterative optimization process.
~\cite{suits} aim to disentangle appearance-based feature learning and deformation estimation using a Y-shaped FeatCNN architecture.
} \\

However, such recursive formulations have a large memory footprint due to explicit backpropagation through the entire cascade (shown by ~\cite{bai_deep_2022-1}), and are not adaptive or optimal with respect to the inputs. %
\edit{
For example, ~\cite{gradirn} runs for only 3 iterations at each scale of $4\times, 2\times, 1\times$, while ~\cite{suits} runs for at most 15 iterations at the original resolution.
In contrast, optimization-based methods need to run for hundreds of iterations at multiple scales to achieve good performance and robustness.
}
\edit{Our method {\method} uses implicit backpropagation through optimization} -- guaranteeing convergence to a local minima, and \textit{implicit backpropagation} avoids storing the entire computation graph of iterative optimization.
\edit{
This allows our method to run hundreds of iterative optimization steps at multiple scales with a constant memory footprint.
}

\subsection{Feature Learning for Image Registration}
\edit{
Since backpropagating through generic iterative optimization is difficult, many methods propose using non-iterative parameterizations of the transformation.
Other methods propose using generic feature extractors with the expectation that they contain the features necessary for iterative optimization.
} 

Registration-aware feature learning can facilitate the learning of label-aware features for registration, and is a promising alternative to prediction-based registration.
~\cite{wang2023robust,billot2023se,moyer2021equivariant} learn keypoints from images which is then used to compute the optimal affine transform using a closed form solution.
\cite{billot2023se} proposes learning center-of-mass keypoints from dense volumes that can be used for computing closed-form rigid transforms.
\cite{moyer2021equivariant} similarly uses a bank of equivariant filters and proposes optimizing a closed-form transformation.
The closest related work to our method is ~\cite{wang2023robust}, which learns sparse keypoints from images and uses a closed-form solution to compute the optimal affine or deformable thin-plate spline transform.
Although these methods have both appearance and discretization invariance, they are restricted to transformations that can only be represented by differentiable \textit{closed-form} analytical solutions, like rigid, affine or thin-plate splines. %
In contrast, dense deformable registration (diffeomorphic or free-form) is almost universally solved with non-closed-form representations using iterative optimization methods, owing to their unparalleled flexibility and accuracy. %
Since these representations do not admit closed-form solutions, it strongly motivates the need to perform \textit{implicit differentiation} through an iterative optimization solver to perform feature learning for registration.

\edit{Other approaches like ~\cite{dinoreg,samconvex,convexadam} leverage pretrained feature extractors like DINO, SAM, and nnUNet respectively as feature images for iterative optimization.
~\cite{wu2015scalable,ma2021image,wu2013unsupervised,quan2022self} use unsupervised learning to extract image features for registration. 
However, these methods} do not perform feature learning and registration end-to-end, i.e., the features obtained are not task-aware (registration-aware) and may not be optimal for registration, especially for domain or application specific anatomical landmark alignment.
In these approaches, learned features are either used as inputs into a parameteric form to compute the transformation end-to-end, or are learned using unsupervised learning in a stagewise manner.
\edit{In these approaches, either the benefits of instance optimization are lost by using a non-iterative parameterization, or the synergization of feature learning and optimization is lost by not learning features end-to-end through an iterative optimization solver.
}

In contrast, by implicitly differentiating through a black-box iterative solver, and minimizing the image and label alignment losses end-to-end, {\method} learns features that are \textit{registration-aware}, \textit{label-aware}, and \textit{dense}.
\edit{There is no constraint on the nature or parameterization of the transformation induced by the feature learning.}
The optimization routine also guarantees that the transformation is a local minima of the alignment of high-fidelity feature images.

\subsection{Deep Equilibrium models}
~\cite{bai_deep_2019,geng_torchdeq_2023} propose Deep Equilibrium (DEQ) models that have emerged as an interesting alterative to recurrent architectures.
DEQ layers solve a fixed-point equation of a layer to find its equilibrium state without unrolling the entire computation graph.
\edit{DEQ layers therefore form the cornerstone for differentiating through black-box iterative solvers.} 
~\cite{bai2020multiscale,bai_deep_2022-1,fung_jfb_2021,pokle2022deep,gilton2021deep,yang2022closer} adopt this approach that leads to high expressiveness of the model without the need for memory-intensive backpropagation through time.
~\cite{hu_plug-and-play_2024} uses a DEQ formulation to finetune the PnP denoiser network for registration, but unlike our work, the data-fidelity term comes from the intensity images. %
However, these methods use the DEQ framework to emulate an infinite-layer network, which typically consists of learnable parameters within the recurrent layer.
Contrary to this, {\method} uses DEQ to \edit{leverage \textit{existing} advanced} multi-scale optimization as a layer in a deep network, with no learnable parameters within the optimization layer itself.
This allows us to compute gradients with respect to the feature images, and backpropagate them through the optimization layer, making the learned features registration-aware.
\textbf{Conceptually, our work does not aim to simply emulate such an infinite cascade, but rather use the DEQ framework to \textit{synergize} feature learning and optimization in an end-to-end registration framework.}
This paradigm inherits all the task-specific invariances of optimization-based methods, while leveraging the fidelity of labelmap overlap into learned features.
DEQ allows us to avoid the memory-intensive layer-stacking paradigm for cascades, and use optimization as a black box layer without storing the entire computation graph, leading to constant memory footprint and faster convergence.
This allows learnable features to be registration-aware since gradients are backpropagated to the feature images through the optimization itself.

\subsection{Optical Flow}
\label{sec:optflow}
Optical flow is very similar to deformable image registration in terms of its dense correspondence nature.
Since the optical flow equation is also modelled as an optimization problem, it is dominated by methods that leverage this task-specific inductive bias to learn to perform local correspondence in some learned feature space.
\cite{fischer2015flownet} uses a correlation volume to perform dense feature matching and shows that it generalizes well even when trained exclusively on simple synthetic data.
\cite{pwcnet} uses pyramidal processing, warping, and a cost volume, \cite{xu2022gmflow} uses a global feature matching layer, \cite{jiang2021learning} computes a correlation volume, \cite{teed_raft_2020,bai_deep_2022-1} also uses a correlation matrix, \cite{liu2020learning} uses PWCNet as a base architecture. 
These methods explicitly model the dense correspondence problem via an explicit cost volume that is invariant to a null-space kernel (see more in \cref{sec:reg-invariance}), which is much harder to model with a stack of weighted convolutional layers.
Consequently, these models dominate leaderboards in Sintel, KITTI, and Spring optical flow benchmarks.
We propose a similar approach to these frameworks to explicitly imbue the model with apperance and discretization invariances of the correspondence matching problem, and show that this leads to state-of-the-art performance on a variety of datasets.

\section{Methods}

\subsection{Preliminaries}

Deformation Image Registration (DIR) is typically formulated as a variational optimization problem: %
\begin{equation}
    \label{eq:reg}
    \varphi^* = \arg\min_{\varphi} {L}(I_f, I_m\circ\varphi) + R(\varphi) = \arg\min_{\varphi} {C}(\varphi, I_f, I_m)
\end{equation}
where $I_f$ and $I_m$ are fixed and moving images respectively, $L$ is a loss function that measures the dissimilarity between the fixed image and the transformed moving image, and $R$ is a suitable regularizer that enforces desirable properties of the transformation $\varphi$. 
We call this the \textit{image matching} objective. If the images $I_f$ and $I_m$ are supplemented with anatomical \edit{label maps $S_f$ and $S_m$}, we call this the \textit{label matching} objective.
Classical methods perform image matching on the intensity images, but the label matching performance is bottlenecked by the fidelity of image gradients with respect to the label matching objective. %

Deep learning methods mitigate this \edit{shortcoming} by injecting label matching objectives (for example, Dice score \edit{or landmark distances}) into the objective \cref{eq:reg} and using a deep network with parameters $\theta$ to predict $\varphi$ for every image pair as input.  %
In essence, learning-based problems solve the following objective:
\begin{align}
    \label{eq:reglab}
    \theta^* &= \arg\min_{\theta} \sum_{f,m} {L}(I_f, I_m\circ\varphi_\theta) + {D}(S_f, S_m\circ\varphi_\theta) + R(\varphi_\theta)  \\
    &= \arg\min_{\theta} \sum_{f,m} {T}(\varphi_\theta, I_f, I_m, S_f, S_m)
\end{align}
where  $\varphi_\theta(I_f, I_m)$ is abbreviated to $\varphi_\theta$.
This leads to learned transformations $\varphi_\theta$ that perform both good image and label matching.
However, the feature learning and optimization are coupled, and features are learned implicitly to produce deformation fields. %
Moreover, this formulation does not explicitly imbue any task-specific invariance into the learning framework, and the learned features are optimized only for a specific training domain, leading to poor generalization to domain shift.
In the following text, we discuss the task-specific invariances followed by our model that incorporates them.

\begin{figure*}[ht]
    \centering
    \begin{subfigure}[t]{\linewidth}
        \centering
        \includegraphics[width=0.95\linewidth]{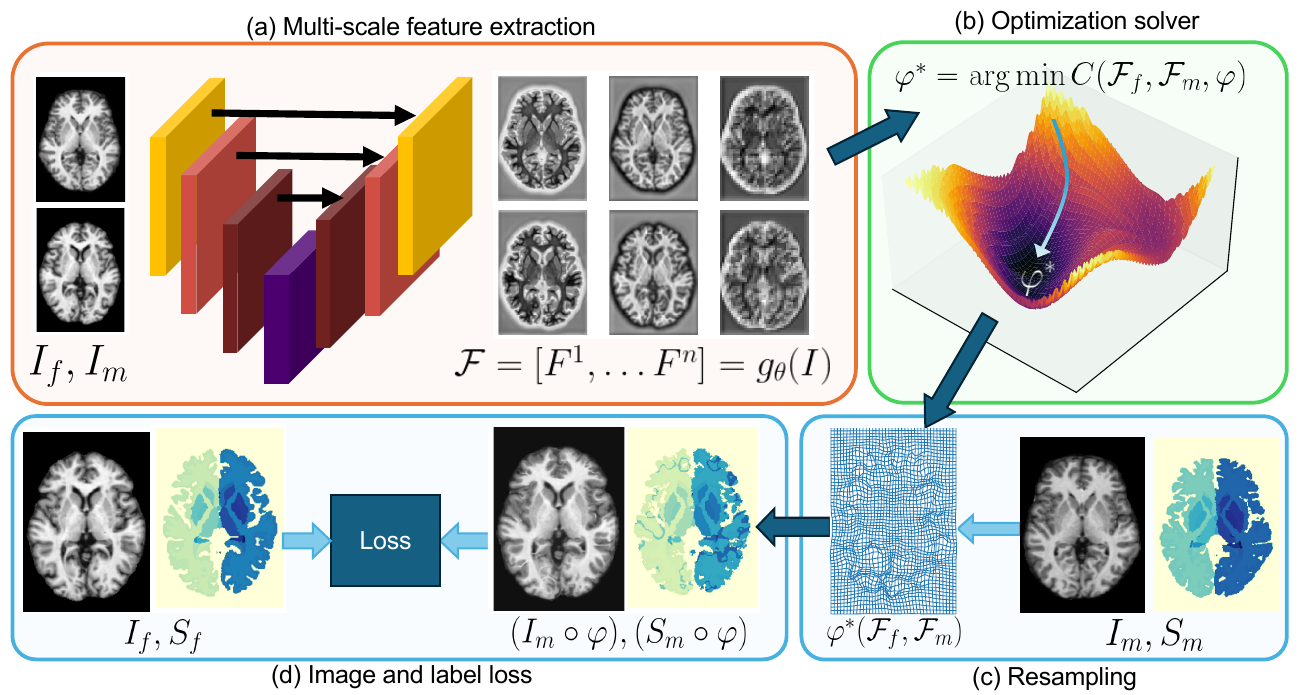}
    \end{subfigure}

    \caption{\textbf{Overview of our framework}. \textbf{(a)} A neural network extracts \textit{dense} multi-scale features from the input images. \textbf{(b)} These features are used to optimize warp fields using a multi-scale differentiable optimization solver. \textbf{(c)} The optimized transform is used to warp the moving image and labels. \textbf{(d)} The warped image/label are compared with the fixed image/label using a similarity metric.}
    \label{fig:overview}
\end{figure*}

\subsubsection{Task-specific invariances}
\label{sec:reg-invariance}
\textit{Correspondence Matching} is fundamentally a geometric problem where we aim to align matching physiological structures.
However, we only have access to observations (images) that are appearance-based.
An ideal correspondence matching algorithm should be invariant to the appearance of the images.
However, factoring out the apperance of an image (or even defining it mathematically) is a challenging problem.
Nevertheless, a model like \cref{eq:reg} is invariant to the following transformations (viewed as proxies of appearance) of the image:

\begin{itemize}
    \item \textbf{Global intensity scaling and translation}: If mean squared error is the loss function, the loss with the original images is given by $\varphi^* = \arg\min_{\varphi} \|I_f - I_m\circ\varphi \|_2^2$.
    When the image intensities are scaled and translated, i.e. $I_f'(x) = sI_f(x) + b, I_m'(x) = sI_m(x)+b$, the loss function is: $\varphi^* = \arg\min_{\varphi} \|I_f' - I_m'\circ\varphi \|_2^2 = $ $\arg\min_{\varphi} \|sI_f - sI_m\circ\varphi + b - b\|_2^2 = \arg\min_{\varphi} \|I_f - I_m\circ\varphi \|_2^2$. 
    Therefore, the optimization problem is identical to the original problem.
    Global scaling and translation of intensities can be a common occurence for MRI images with non-standard units of measurement, or PET scans where the standardized uptake values can vary across institutions.

    \item \textbf{Local intensity scaling and translation}: If local normalized cross-correlation is used as the loss function, then the loss is (nearly) invariant to local scaling and translation of intensity images. This kind of behavior may be seen in images with bias fields, shading artifacts due to RF coil sensitivity, gradient-driven eddy currents, or other imaging inhomogenities.

    \item \textbf{Monotonic intensity transforms}: If mutual information is used as the loss function, the loss is invariant to monotonic intensity transforms of the images. This can be seen in images with varying intensity profiles due to different acquisition protocols, or different scanners where the intensity profiles are monotonic but not identical.

    \item \textbf{Kernel of a linear operator}: A cost correlation volume is used like ~\cite{xu2022gmflow,teed_raft_2020}, i.e. $C = \left< W F_f, W F_m \right>$, where $F_f = g_\theta(I_f)$ and $F_m = g_\theta(I_m)$ are the features from a network, and $W$ is a projection matrix before computing the cost volume.
    In this case, any changes to the feature images that lie in the kernel of $W$ do not affect the cost volume, and therefore the subsequent optimization problem.
    This can be a learned way of delineating apperance related features from the images into the kernel of $W$, factoring out its effect on the optimization problem.
\end{itemize}
Moreover, the optimization problem \cref{eq:reg} is invariant to the discretization of the optimization algorithm by definition.

On the other hand, deep learning methods are not inherently invariant to these transformations.
For example, consider a convolutional network architecture with convolutions, ReLU/LeakyReLU activations and batch normalization with negligible bias terms in all layers.
In this case, the network propagates the scale of the inputs to the output, i.e. if $I_f'(x) = sI_f(x)$, then $\phi_\theta(I_f', I_m') = \phi_\theta(sI_f, sI_m) = s \phi_\theta(I_f, I_m)$.
However, we want $\phi_\theta(sI_f, sI_m) = \phi_\theta(I_f, I_m)$.
This inductive bias therefore has to be learned from scratch requiring a large amount of data and does not guarantee generalization to domain shift.
Moreover, ~\cite{kovachki2023neural} mention that convolutional networks are not discretization invariant due to their fixed resolution kernels.
These invariances are crucial for the success of classical optimization-based registration methods, and are not guaranteed by deep learning methods.
In the following section, we discuss how we incorporate these invariances into our model by using \cref{eq:reg} as a layer in a deep network.

\cref{fig:overview} shows the overview of our method.
Our goal is to learn feature images such that \textbf{registration in this feature space corresponds to both good image and label matching performance}, while retaining the invariances of \cref{eq:reg}. %
We do this by using a feature network to extract dense features from the intensity image, that now parameterizes \cref{eq:reg}.
Using a black-box solver, we solve \cref{eq:reg} to obtain an optimal transform $\varphi^*$.
This optimal $\varphi^*$ is plugged into \cref{eq:reglab} to obtain gradients with respect to $\varphi^*$ to maximize both image and label matching. 
Since $\varphi^*$ is a function of the feature images, we \textit{implicitly differentiate} through the optimizer to backpropagate gradients to the feature images and to the deep network.
\edit{This framework synergizes feature learning and optimization by backpropagating through the black-box optimizer, and allowing the optimizer to run hundreds of iterations at multiple scales with a constant and modest computational budget.}
We discuss the building blocks of our framework in the following sections.

\subsection{\edit{Dual-stream} Feature Extractor Network}
The first component of our framework is a \edit{dual stream} feature network that extracts dense features from the intensity images.
This network is parameterized by $\theta$, and takes an image $I \in \mathbb{R}^{H\times W \times D \times C_{in}}$ as input and outputs a feature map $F \in \mathbb{R}^{H\times W\times D\times C}$, where $C$ is the number of feature channels, i.e. $F = g_\theta(I)$.
\edit{Unlike parameteric DLIR methods where moving and fixed images are concatenated and passed to the network to estimate a parameteric warp representation, our feature network processes the images \textit{independently}.}
This allows the fixed and moving images to be of different voxel sizes.
The feature network can also output multi-scale feature maps $\mathcal{F} = g_\theta(I) = [F^0, F^1, \ldots, F^N]$, where $F^k \in \mathbb{R}^{H/2^k\times W/2^k \times D/2^k\times C_k}$, which can be used by multi-scale optimization solvers.
\edit{The overall framework does not dictate a particular choice of architecture, and we ablate on different popular architectures in the experiments.}

Note that in contrast to \cref{eq:reg} that has fixed dynamics because of fixed $I_f$ and $I_m$, the learned features induce a modified learned optimization described as follows:
\begin{align}
    \label{eq:regfeat}
    \arg\min_\varphi {C}(\varphi, \textcolor{green}{F_f}, \textcolor{green}{F_m})
\end{align}
Since $F_f$ and $F_m$ are learned using a deep network, we can now \textit{explicitly} imbue the task-specific inductive biases into arbitrary learned features.

\edit{In this work, we focus on the iterative refinement stage for end-to-end optimization of the learned image features from a task-specific training dataset}.

\subsection{Implicit Differentiation through Optimization}

\edit{
Due to the inherent limitations of parameteric warp field representation, prior works like ~\cite{gradirn,rwcnet,suits} have proposed to use recurrent architectural designs to mimic the flavor of traiditonal iterative optimization algorithms.
The iterative optimization is designed to minimize the image and label matching dissimilarity from the training data.
However, these methods are still different from traditional optimization in a few fundamental ways.
First, traditional optimization algorithms have a well-defined stopping criteria (i.e. convergence to a local minima).
This usually requires hundreds of iterative optimization steps over multiple resolutions.
In contrast, existing works employing recurrent architectures have a few number of fixed iterations at each step.
Second, an instance optimization solver does not require the entire optimization path, in contrast to recurrent architectures where the entire unrolled path must be stored to perform backpropagation-through-time (BPTT). 
These reasons limit the expressive capacity of iterative optimization while allowing backpropagation through the solver.}

\edit{We propose to close the gap using an implicit differentiation approach to leverage powerful image registration solver toolkits.}
\edit{Specifically}, given the feature maps $F_f$ and $F_m$ extracted from the fixed and moving images \edit{using a neural network, a gradient-based iterative} solver optimizes \cref{eq:regfeat} to obtain the optimal transformation $\varphi^*$.
\edit{
}
The minimization objective converges when \edit{the gradient of the dissimilarity is zero:}
\begin{equation}
   \label{eq:gradient}
   \varrho(\varphi^*, F_f, F_m) = \nabla_{\varphi}C = \frac{\partial C}{\partial \varphi}\Bigg|_{\varphi^*} = 0
\end{equation}
\edit{
At this point, subsequent iterations of the optimization do not change the value of $\varphi^*$.
Therefore, $\varphi^*$ can be thought of as the fixed point of an `infinite-layer' iterative optimization solver. 
}
This \edit{value of} $\varphi^*$ is then used to compute the loss \cref{eq:reglab} to minimize image and label matching objective.

Note that the \edit{analytical form of the} vector-valued function $\varrho$ is induced by the choice of scalar-valued loss function $C$ \edit{used to run the optimization in ~\cref{eq:regfeat}}.
For example, \edit{choosing to minimize the sum of squared distance loss} $C(F_f, F_m \circ \varphi) = \|F_m \circ \varphi - F_f\|_2^2$ \edit{induces} $\varrho(\varphi, F_f, F_m) = (F_m\circ\varphi - F_f)(\nabla F_m\circ\varphi)$.
To propagate derivatives from $\varphi^*$ to the feature images $F_f, F_m$, we invoke the Implicit Function Theorem~\cite{krantz2002implicit}:
\begin{theorem}
   \label{thm:implicit} 
   For a function $\varrho: \mathbb{R}^n \times \mathbb{R}^{m_1 + m_2} \rightarrow \mathbb{R}^n$ that is continuously differentiable, if $\varrho(\varphi^*, F_f, F_m) = 0$ and $\Big| \frac{\partial \varrho}{\partial \varphi}\Big||_{\varphi^*} \neq 0$, then there exist open sets $U, V_f, V_m$ containing $\varphi^*, F_f, F_m$, and a function $\varphi^*(F_f, F_m)$ defined on these open sets such that $\varrho(\varphi^*(F_f, F_m), F_f, F_m) = 0$.
\end{theorem}

Given the Implicit Function Theorem (IFT), we write $\varrho(\varphi^*(F_f, F_m), F_f, F_m) = 0$ and differentiate with respect to $F_f$ to obtain:
\begin{align}
    \frac{d\varrho}{dF_f} &= \frac{\partial \varrho}{\partial \varphi}\frac{\partial \varphi}{\partial F_f} + \frac{\partial \varrho}{\partial F_f} = 0 \\
    \implies \frac{\partial \varphi}{\partial F_f} &= -\left(\frac{\partial \varrho}{\partial \varphi}\right)^{-1}\frac{\partial \varrho}{\partial F_f} \\
    \implies \frac{\partial T}{\partial F_f} &= \frac{\partial T}{\partial \varphi} \frac{\partial \varphi}{\partial F_f} \\
    \label{eq:ift}
    &= -\frac{\partial T}{\partial \varphi}\left(\frac{\partial \varrho}{\partial \varphi}\right)^{-1}\frac{\partial \varrho}{\partial F_f}
\end{align}
\edit{The forward pass of the layer is simply the iterative solver run without any unrolling or storing of any intermediate steps.
During the backward pass, ~\cref{eq:ift} provides the analytical form for computing the derivative of $\varphi$ with respect to the feature images.
We explain how to use the result of ~\cref{eq:ift} to compute the gradients of the network with respect to the training loss in ~\cref{sec:iftimpl}.}

\edit{This design allows maximal expressivity of the iterative optimization solver by allowing hundreds of iterations until convergence, while being agnostic to the nature of the solver.
Moreover, there are no additional memory overheads for optimization.
In contrast, explicit T-step unrolling in prior work requires an $O(T)$ memory overhead for BPTT, rendering it infeasible for 3D image registration.
}

\edit{
To summarize, the implicit optimization's forward pass directly runs the optimization without additional overhead.
During the backward pass, the optimal $\varphi^*$ is used to compute the gradients with respect to the feature images $F_f, F_m$.
The gradients are subsequently passed back to the weights of the neural network.
}

\edit{\subsection{Computing the Implicit Gradient}
\label{sec:iftimpl}
There are two parts to computing the feature gradients:
\begin{itemize}[nosep]
    \item Computing the modified gradient $v^T = \frac{\partial T}{\partial \varphi}\left(\frac{\partial \varrho}{\partial \varphi}\right)^{-1}$
    \item Computing the gradient w.r.t. feature image $v^T\frac{\partial \varrho}{\partial F_f}$
\end{itemize}
We describe how to compute these gradients in the following sections.
}

\edit{
\subsubsection{Computing the Inverse Jacobian}
An important component of the implicit differentiation is the computation of the inverse Jacobian $\left(\frac{\partial \varrho}{\partial \varphi}\right)^{-1}$.
~\cite{bai_deep_2019} propose using a quasi-Newton approach to solve the linear system $\left(\frac{\partial \varrho}{\partial \varphi}\right) v = -\frac{\partial T}{\partial \varphi}$.
This requires solving another iterative optimization in the backward pass, that can be slow.
For general problems, there are typically no alternatives to performing iterative optimization, since the Jacobian does not have a reduced form.
}

\edit{
However, we exploit a special structure of the Jacobian that allows us to compute the inverse Jacobian efficiently without any iterative methods.
First, we note that since $\varrho = \frac{\partial C}{\partial \varphi}$, the Jacobian $\frac{\partial \varrho}{\partial \varphi}$ is the Hessian of the loss function $\nabla^2_{\varphi} C(\varphi)$.
This quantity is a $(n_v \cdot d)\times (n_v \cdot d)$ matrix, where $n_v$ is the number of voxels in $\varphi$, and $d$ is the spatial dimension, 
In a typical 3D registration scenario, $n_v$ is of the order of $10^7$, making this quantity hard to compute in general.
However, for the mean squared error, i.e. $C(\varphi, F_f, F_m) = \|F_f - F_m\circ\varphi\|_2^2$, the Hessian $\frac{\partial \varrho}{\partial \varphi}$ is a block-diagonal matrix, since there are no terms in $C$ containing both $\varphi(x_p)$ and $\varphi(x_q)$ for voxel indices $p \neq q$.
Specifically, we have 
\begin{align}
   (\varrho)(\varphi(x_p)) &= \nabla_{\varphi(x_p)} C(\varphi, F_f, F_m) \\
    &= (F_m(\varphi(x_p)) - F_f(x_p))\nabla F_m(\varphi(x_p))
\end{align}
This quantity is a vector of size $d$ due to the term $\nabla F_m(\varphi(x_p))$, and has no terms involving $\varphi(x_q)$ for $q \neq p$.
We now consider the scalar 
\begin{align}
g_i = \sum_p (\varrho)(\varphi(x_p))[i]
\end{align}
, where $[i]$ is the $i^{\text{th}}$ index of a vector.
The gradient of $g_i$ with respect to $\varphi(x_q)$ is therefore 
\begin{align}
    \nabla_{\varphi(x_q)}{\left(g_i\right)} &= \nabla_{\varphi(x_q)} \sum_p \left(\varrho(\varphi(x_p))\right)[i] \\
    &= \nabla_{{\varphi(x_q)}}\left({\varrho(\varphi(x_q))}[i]\right) \\ 
    &= \left(\nabla^2_{{\varphi(x_q)}}{C(\varphi)}\right)[i]
\end{align}
which is the $i^{th}$ row of the Hessian block corresponding to the voxel $x_q$.
}

\edit{
All the aforementioned operations can be performed efficiently using automatic differentiation libraries.
We compute the gradients of each $g_i$ ($i = 1, 2 \ldots d$) and stack them to obtain the full blockwise Hessian of size $n_v\times d \times d$.
Next, we can solve the following $d\times d$ system of equations for $v_p$ for each voxel $p$ independently:
\begin{align}
    \label{eq:smallsolver}
    \frac{\partial T}{\partial \varphi(x_p)} &= [\nabla_{\varphi(x_p)}{\left(g_1\right)}; \ldots ; \nabla_{\varphi(x_p)}{\left(g_d\right)}] v_p
\end{align}
Since $d$ is 2 or 3, \cref{eq:smallsolver} can be solved efficiently using standard linear algebra methods.
\cref{eq:smallsolver} allows us to compute the modified gradient $v^T = \frac{\partial T}{\partial \varphi} \left(\frac{\partial \varrho}{\partial \varphi}\right)^{-1}$.
}

\edit{\subsubsection{Computing the Feature Gradients}}
\edit{
Note that $\frac{\partial \varrho}{\partial F_f}$ is a matrix of size $n\times m_1$.
Here, $n = n_v\cdot d$ is the number of parameters in $\varphi$ and $m_1 = n \cdot C$ is the number of parameters in $F_f$.
Similar to $\frac{\partial \varrho}{\partial \varphi}$, this matrix is infeasible to compute in general.
}

\edit{
Fortunately, similar to most automatic differentiation libraries, this quantity is not computed explicitly.
Instead, Vector-Jacobian Products ~\cite{jax} are used to compute the quantity ${\frac{\partial T}{\partial F_f}} = -v^T \frac{\partial \varrho}{\partial F_f}$ directly.
The quantity $\varphi^*$ is used during the forward pass to compute the training loss ~\cref{eq:reglab} and the backward gradient $\frac{\partial T}{\partial \varphi}$.
This gradient is modified using \cref{eq:smallsolver} to obtain the modified gradient $v$.
The backward pass for feature gradients is then obtained by first computing the scalar quantity  $h = v^T \cdot \varrho(F_f, F_m, \varphi)$.
The derivative of the scalar $h$ with respect to $F_f$ is $\frac{\partial h}{\partial F_f} = v^T \frac{\partial \varrho}{\partial F_f}$. 
This is an application of the chain rule to compute vector-Jacobian product $\frac{\partial h}{\partial F_f}$ without explicitly computing the full matrix $\frac{\partial \varrho}{\partial F_f}$. 
The gradients of $F_m$ are obtained similarly.
\ref{alg:pseudocode-bwd} outlines the pseudocode for computing the gradients of the feature images with respect to the training loss.
These features can then be propagated back to the network to update the weights.
}

\subsection{Multi-scale optimization}
Iterative optimization based methods typically use a multi-scale approach to improve convergence and avoid local minima with the image matching objective~\cite{avants_lagrangian_2006,avants_symmetric_2008,ashburner2007fast,beg2005computing}.
However, the downsampling of intensity images leads to indiscriminate blurring and loss of details at the coarser scales.
We adopt a multi-scale approach by using pyramidal features from the network, which are naturally built into many UNet-like architectures.
\edit{
We consider two network designs for extracting multi-scale features:
\begin{itemize}
    \item \textbf{Shared decoder}: We consider UNet-like architectures, and use the features from the decoder layers at each resolution as multi-scale features. 
    The decoder features from each resolution are fed to an additional convolutional layer to obtain multi-scale feature maps for the fixed and moving images. 
    This allows propagation of dense gradients to multiple decoder layers and feature sharing within the network.
    This architecture is illustrated in ~\cref{fig:architectures}(a).
    \item \textbf{Independent decoders}: We consider a cascade of networks with separate decoders, where each decoder processes the images at different resolutions.
    We hypothesize that for multi-scale optimization, independent consideration of different scales may be necessary to extract relevant features at each scale.
    This architecture is illustrated in ~\cref{fig:architectures}(b).
\end{itemize}
We perform an ablation study on the choice of network architecture in \cref{sec:zeroshot}.
}

\edit{
Given these multi-scale features $\mathcal{F}_f = [F_f^n \ldots F_f^2, F_f^1]$ and $\mathcal{F}_m = [F_m^n \ldots F_m^2, F_m^1]$, we first perform optimization at the coarsest scale $n$, and store the result $\varphi^{*(n)}$.
For each subsequent level $k < n$, we first upsample $\varphi^{*(k+1)}$ to the resolution of $F_f^k$, and use this as initialization for the optimization at the next finer scale.
Finally, all the upsampled $\varphi^{*(k)}$ are used to compute the training loss ~\cref{eq:reglab}.
This mimics traditional multi-scale optimization methods while storing the result of the optimization at each scale for backpropagating to all feature maps.
This asymmetry of the multi-scale features allows the network to learn different features at different scales, for example, large ventricles at coarser scales and small sulci structure at finer scales.
}
A comparison of classical registration algorithm and our algorithm is highlighted in \cref{alg:classical-pcode,alg:dio-pcode}.

\subsection{Implementation Details}
Formulating arbitrary iterative solvers using implicit differentiation allows full expressivity of powerful solvers for learning-based image registration.
We elaborate on the implementation details that make this framework practical and scalable.

\subsubsection{Jacobian-Free Backprop}  
\label{sec:jfb}
\edit{
In practice, the ill-conditioned nature of the inverse Hessian leads to poor training performance.
To avoid the ill-conditioning, we follow~\cite{fung_jfb_2021} and substitute the Jacobian to identity, to compute $\hat{\frac{\partial {T}}{\partial F_f}} \approx -\frac{\partial T}{\partial \varphi}\frac{\partial \varrho}{\partial F_f}$.
This leads to lesser memory and compute requirements during the backward pass, and stable training dynamics compared to other estimates of Jacobian like phantom gradients, damped unrolling, or Neumann series ~\cite{geng2021training,geng_torchdeq_2023}.
We perform an ablation on using full blockwise Hessian and unrolling-based phantom gradient~\cite{geng2021training} in \cref{sec:ablationimplicitgradient}.
}

\subsubsection{Double Backward through \texttt{grid\_sample}}
\edit{
Note that in \cref{eq:gradient}, $\varrho$ contains a $\nabla F_m \circ \varphi$ term, and the quantity $\frac{\partial \varrho}{\partial F_m}$ will require the double-backward pass of the \texttt{grid\_sample} operator in PyTorch. Since this operation is not implemented in the PyTorch C backend, a backward pass for the gradient operation does not exist in PyTorch.
We use the \texttt{gridsample\_grad2} library~\cite{siarohin2023cudagridsamplegrad2} to compute the double-backward pass of the \texttt{grid\_sample} operator in \cref{eq:gradient}.
}

\subsubsection{Other details}
\edit{
For all experiments, we use multi-scale features with $4\times, 2\times, 1\times$ downsampling for multi-scale optimization, unless otherwise mentioned.
We run the solver for a maximum of $200, 100, 50$ iterations for each scale respectively, with an early stopping criteria if the relative loss does not change by more than $10^{-4}$ for 5 iterations. 
We choose the MSE loss for the feature matching objective within the solver. 
For the non-diffeomorphic iterative optimizer, we use a simple nonparameteric displacement field representation and an SGD-based solver with a learning rate of $0.003$.
For the diffeomorphic optimizer, we use the FireANTs library with Adam optimizer and a learning rate of $0.5$.
For learning the parameters of the feature network, we use the AdamW optimizer with a learning rate of $0.0003$.
All methods are implemented in PyTorch, and all experiments are performed on a single NVIDIA A6000 GPU.
}

\section{Experiments}

We show the efficacy of {\method} on a comprehensive experiment setup.
First, we show that our method can synthesize dense feature maps from sparse intensity images, facilitating sparse or dense registration.
We illustrate this on a toy dataset where classical optimization methods fail due to the lack of gradients in the loss landscape.
This is especially relevant for incorporating sparse anatomical landmark losses into registration, where classical methods typically do not provide meaningful gradients.
\edit{Second, we compare the in-distribution performance and flexibility of our learned representations with existing methods that aim to leverage either (a) pretrained features or intensity images for iterative optimization, (b) end-to-end or learned image features for parametric warp field regression, and (c) learning-based explicit unrolled iterative methods.
We choose two community-standard datasets for this comparison -- the OASIS dataset for inter-subject brain MRI registration, and the NLST dataset for intra-subject lung CT registration.
}
\edit{Qualitatively, we show our multi-scale features are task-aware, interpretable and agnostic to choice of solver, and the implicit differentiation framework allows high expressive capacity for optimization than baselines.}
Third, to substantiate the robustness of {\method}, we evaluate its performance on three out-of-distribution (OOD) neuroimaging datasets.
Our method demonstrates remarkable robustness to domain shift, outperforming other prediction-based methods.
This robustness is important in the context for DLIR since domain-shift leads 
to a shift in the distribution of warps, subseqeuntly resulting in poor generalization ~\cite{fu2020lungregnet,wolterink_implicit_nodate,mok2022affine,bigalke2022adapting,hansen_graphregnet_2021}
, limiting deployment in clinical settings.
Furthermore, we show that our method allows \textit{zero-shot} test-time switching of optimizers \edit{and efficacy across architectures}, enabling arbitrary transformation representations and constraints at test time.
We also evaluate the inference time of our method \edit{and compare it to explicit recurrent architectures that emulate iterative optimization}, and show that our method is fast, \edit{compute-efficient} and amenable to rapid experimentation and hyperparameter tuning.
\edit{
Finally, we examine the effect of choosing different implicit differentiation backends, and show that Jacobian-free backprop is the most well-conditioned and efficient for our task.
}

\begin{figure*}[ht!] %
\centering
\begin{minipage}{0.9\linewidth}
    \includegraphics[width=0.23\linewidth]{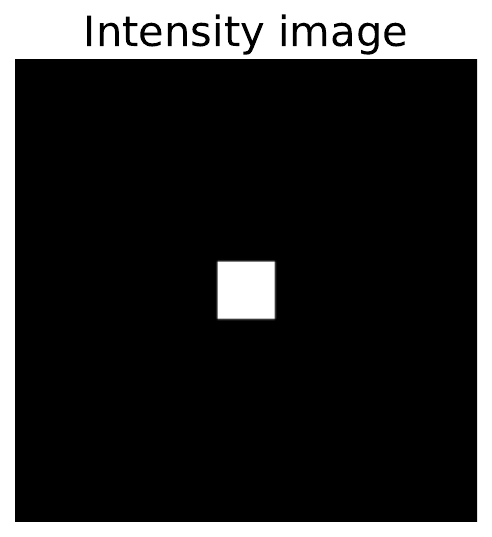}
    \includegraphics[width=0.75\linewidth]{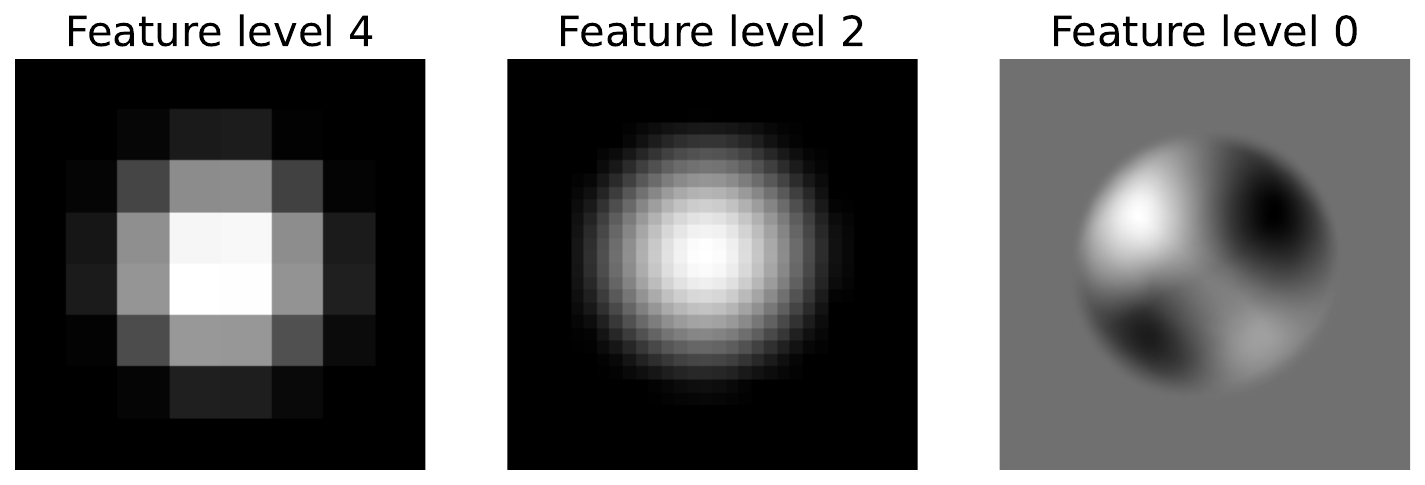} 
\end{minipage}
\begin{minipage}{0.92\linewidth}
    \includegraphics[width=0.25\linewidth]{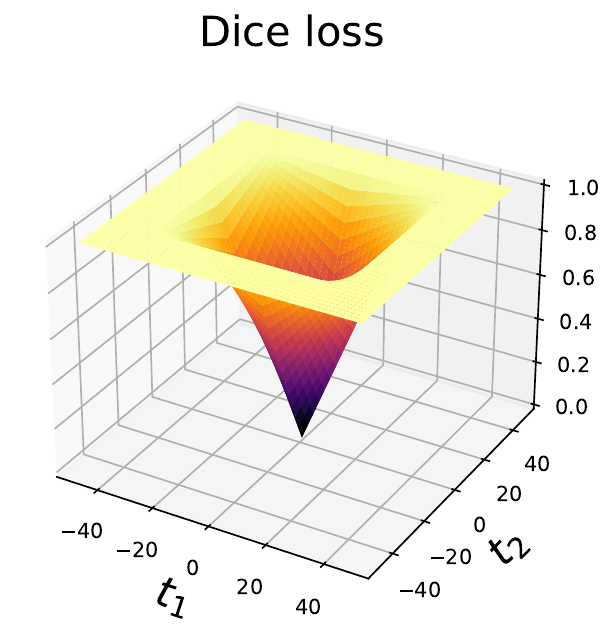}
    \includegraphics[width=0.75\linewidth]{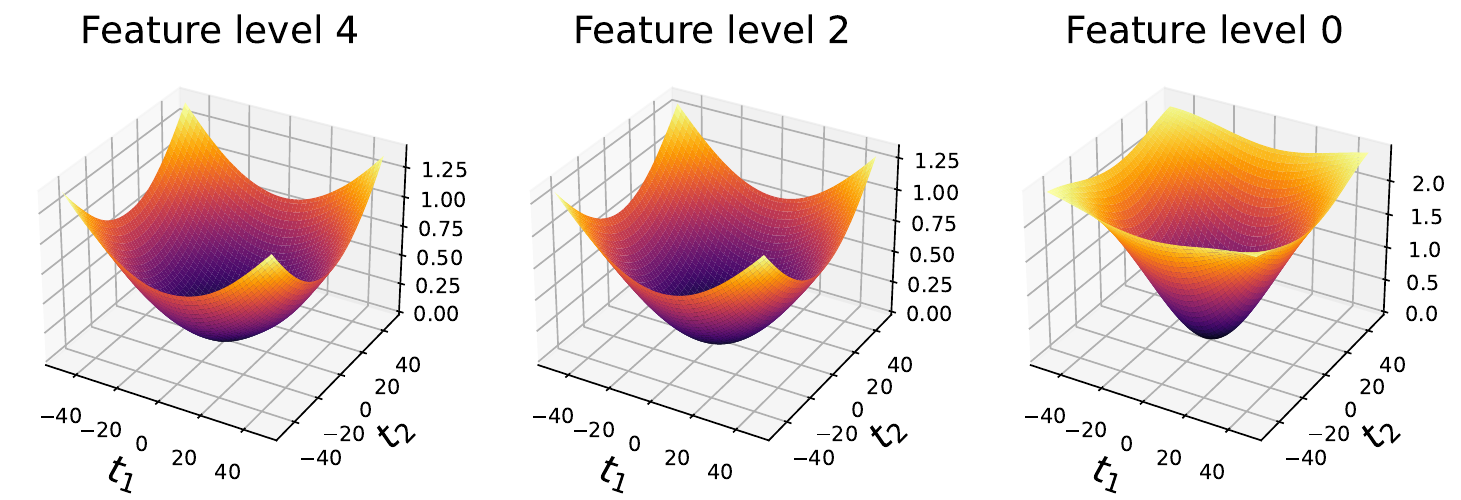}
\end{minipage}
\caption{\footnotesize \textbf{Dense feature learning leads to flatter loss landscapes}. \textit{Top row} shows the intensity image with the corresponding multi-scale features predicted by the deep network, where the $L^{\text{th}}$ level denotes a feature of size $H/2^k{\times}W/2^k{\times}C_k$. \textit{Bottom row} shows the loss landscape as a function of the relative translation between the squares in the fixed and moving image. Note the flat maxima which occurs when there is no overlap between the fixed and moving image, making optimization impossible if there is no overlap of the squares at initialization. On the contrary, the loss landscape for learned features is smooth, even at the finest scale, leading to much faster convergence even when there is no overlap between the intensity images.
This allows registration without any centroid or moment-based preprocessing.
}
\label{fig:toyexample}
\end{figure*}

\subsection{{\method} learns dense features from sparse images}
\label{sec:toyexample}

A key strength of {\method} is the ability to learn interpretable dense features from sparse intensity images for accurate and robust image matching.
This is particularly pertinent for medical image registration, where intensity images often exhibit significant heterogeneity in their gradient profiles, making registration difficult.
We design a toy task to isolate and demonstrate this behavior.
In this task, the fixed and moving images are generated by placing a square of size $32{\times}32$ pixels on an empty canvas of $128{\times}128$ pixels. %
The probability of the squares in the fixed and moving images having non-zero overlap is set to 50\%.
The objective is to find an affine transformation to align the two images.
However, classical optimization methods will fail this task 50\% of the time, since there is no gradient of the loss function when the squares do not overlap, illustrated by the flat loss landscape in \cref{fig:toyexample}.
In contrast, deep networks learn features that significantly flatten this loss landscape in the feature space. 
To demonstrate this, we train a network to output multi-scale feature maps that is used to iteratively optimize \cref{eq:reg} to recover an affine transform.
We choose a 2D UNet architecture, and the multi-scale feature maps are recovered from different layers of the decoder path of the UNet. 
Since the features are trained to maximize dice overlap, the loss landscape is much flatter, and the network is able to recover the affine transform with $>99\%$ overlap regardless of whether there is any initial overlap or not (\cref{app:toyexample}).
This allows registration of labelmaps with sparse gradients without any centroid or moment-based preprocessing~\cite{legouhy2023polaffini,greedy}, which is typically done to offset the lack of gradients in the loss landscape.
Moreover, end-to-end learning also enables learning of features that are most conducive to registration, unlike existing work ~\cite{wu2015scalable,ma2021image,wu2013unsupervised,quan2022self} that may not contain discriminative registration-aware features about anatomical labels due to stagewise training. %

\edit{\subsection{Comparison of in-distribution performance}}
\label{sec:cmp-km}

\edit{\paragraph{Datasets}
We evaluate our method on two datasets - the OASIS dataset for inter-subject brain MRI registration, and the NLST dataset for intra-subject lung CT registration.
}

\edit{\textbf{OASIS}: The OASIS dataset~\cite{oasisdataset} contains 414 T1-weighted MRI scans of the brain with label maps containing 35 subcortical structures extracted from automatic segmentation with FreeSurfer and SAMSEG.
We use the preprocessed version and train-val split from the Learn2Reg challenge \cite{hering2022learn2reg} where all the volumes are skull-stripped, intensity-corrected and center-cropped to $160{\times}192{\times}224$.
We evaluate the Dice score and the 95th percentile of the Hausdorff distance (HD95) between the warped and fixed label maps.
}

\edit{
\textbf{NLST}: The National Lung Screening Trial (NLST) dataset~\cite{nlst} consists of intra-subject inspiration-expiration pairs.
The preprocessed version consists of 200 training pairs and 10 validation pairs, with corresponding keypoints obtained using \editv{automatic landmark detection using the Foerstner operator}.
Owing to large variability in lung volume due to inspiration and expiration, the NLST dataset requires large deformation fields to align the two volumes reliably.
We evaluate the 70th percentile of target registration error (TRE30) of the keypoints between the warped and fixed volumes from the validation dataset.
}

\edit{\paragraph{Baselines} We consider a variety of baselines for this comparison.
Our primary contribution is enabling the synergy of task-aware feature learning and powerful black-box solvers end-to-end.
Therefore, the baselines are categorized into three relevant groups -- \textbf{(a)} using intensity images or generic pretrained features combined with iterative optimization-based methods, \textbf{(b)} parametric regression of warp fields using neural network, and \textbf{(c)} learning-based explicit unrolled iterative methods. 
}

\edit{For the OASIS daaset, we consider \textbf{(a)} SyN~\cite{avants_symmetric_2008}, NiftyReg~\cite{niftyreg}, Log Demons~\cite{vercauteren_symmetric_2008}, FireANTs~\cite{jena2024fireants}, ConvexAdam~\cite{convexadam}, DINO-Reg~\cite{dinoreg}, 
\textbf{(b)} SynthMorph~\cite{hoffmann2021synthmorph}, KeyMorph~\cite{wang2023robust}, Cyclical Self-Training \cite{bigalke2023unsupervised}, \textbf{(c)}  multimodal SUITS ~\cite{suits} and GradIRN ~\cite{gradirn}.
For the NLST dataset, we consider (a) ConvexAdam, SyN, FireANTs (b) VoxelMorph~\cite{vxmpp}, unigradICON~\cite{tian2024unigradicon} (with and without instance optimization), Vector-Field Attention~\cite{vfa}, Im2grid~\cite{im2grid}, and (c) RWC-Net~\cite{rwcnet}.
All methods are trained with a combination of intensity and label or keypoint matching losses, wherever applicable.
}

\edit{\paragraph{Results} \cref{fig:oasis} summarizes the results.
On the OASIS dataset, we observe that all iterative optimization methods perform in the same ballpark without supervision.
We run ConvexAdam with the intensity images and do not observe any improvement over the unsupervised baselines.
We swap the intensity images with DINO features (DINO-reg without ensembling) and observe no improvement in performance - bolstering our claim that generic features do not guarantee task-specific performance.
Iterative methods like GradIRN and SUITs are modified and trained on both the intensity images and label maps, but do not show significant improvement either, due to the limited expressivity of unrolled optimizations.
Supervised parametric baselines like LapIRN and LKU-Net show much better performance for in-distribution datasets, but completely breakdown for out-of-distribution datasets (\cref{sec:ood}). 
Our method shows a significant improvement over unsupervised iterative methods, generic features, and explicit unrolling of optimization.
}

\edit{
On the NLST dataset, we see a similar trend where unsupervised optimization methods like ConvexAdam and FireANTs show solid performance, while parametric methods like VoxelMorph, unigradICON, Vector-Field Attention, and Im2grid show relatively poor performance.
unigradICON substantially improves with instance optimization, indicating the necessity of instance optimization for robust registration.
RWC-Net being an iterative method also reports a poorer performance compared to ConvexAdam and FireANTs, showing that powerful optimization solvers with handcrafted features can surpass learned features with limited expressivity of unrolled optimization.
{\method} improves over the unsupervised baselines and parametric warp field estimators, showing robust performance to multiple anatomical structures and large deformations. 
}

\begin{figure*}[h]
\newlength{\imagewidth}     %
\setlength{\imagewidth}{0.115\linewidth}  %
\centering

\rotatebox{90}{\adjustbox{valign=m}{\hspace{1em}KeyMorph}}
\includegraphics[width=\imagewidth]{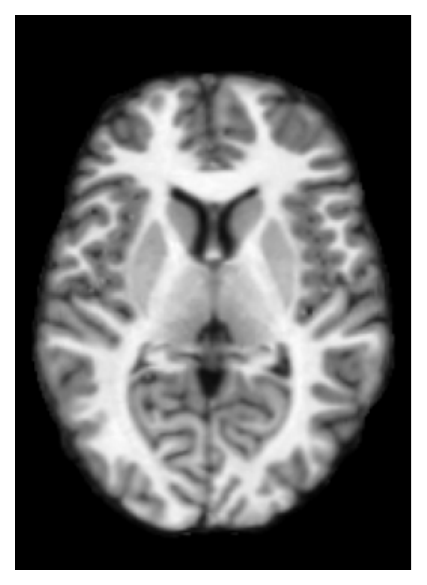}
\includegraphics[width=\imagewidth]{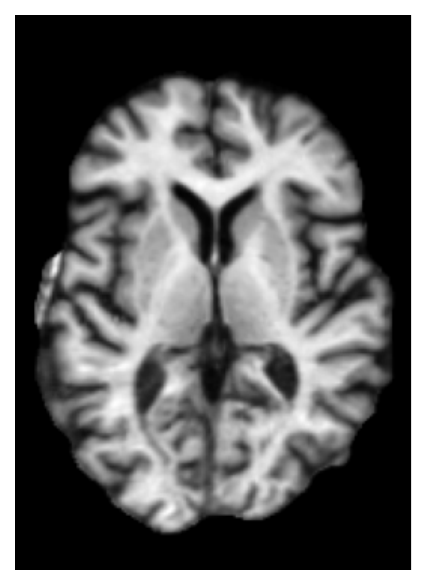}
\includegraphics[width=\imagewidth]{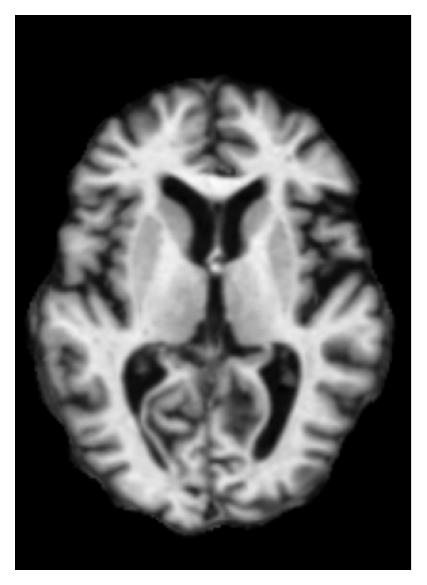}
\includegraphics[width=\imagewidth]{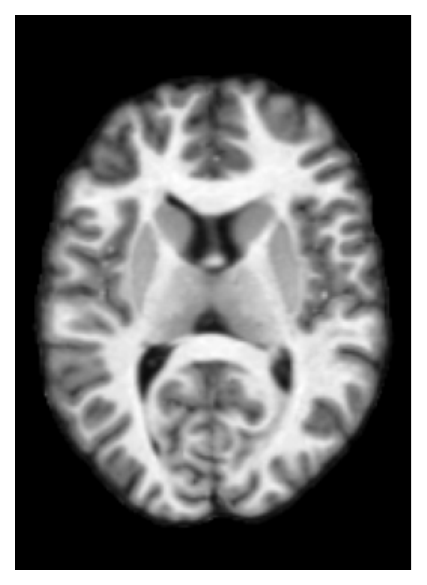}
\includegraphics[width=\imagewidth]{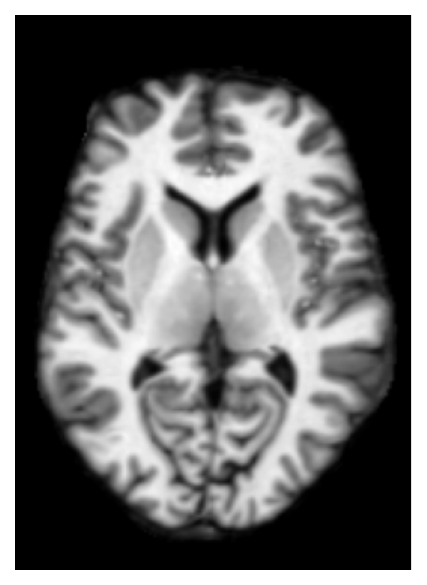}
\includegraphics[width=\imagewidth]{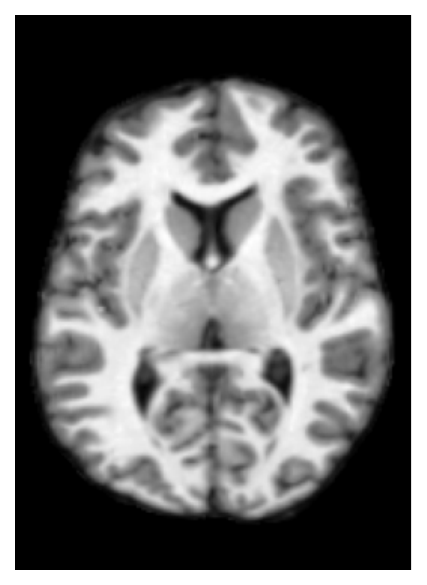}
\includegraphics[width=\imagewidth]{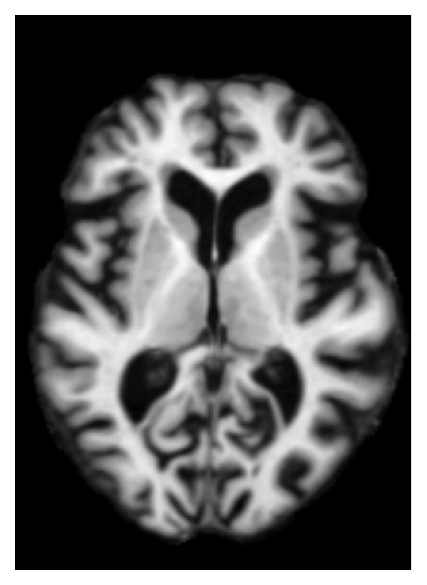}
\includegraphics[width=\imagewidth]{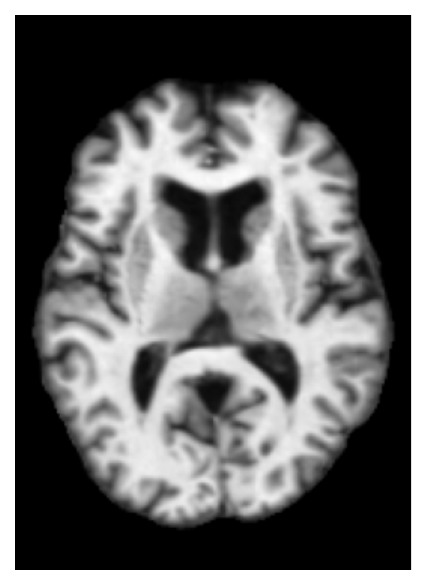}

\hrule

\rotatebox{90}{\adjustbox{valign=m}{\hspace{2.2em}Ours}}
\includegraphics[width=\imagewidth]{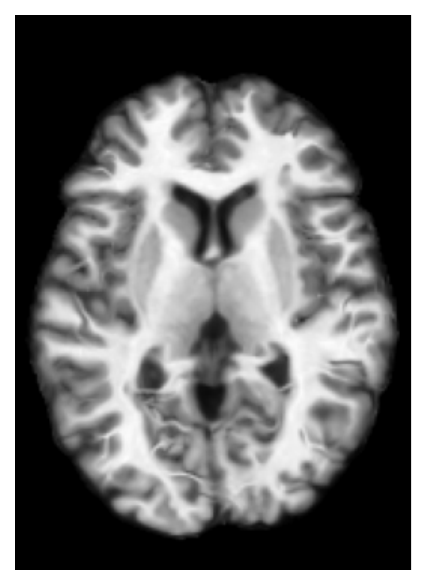}
\includegraphics[width=\imagewidth]{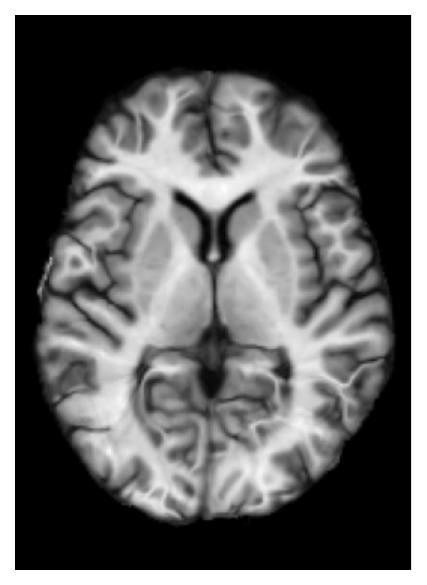}
\includegraphics[width=\imagewidth]{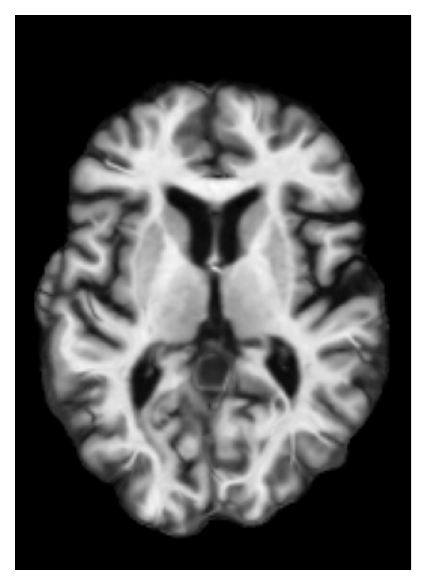}
\includegraphics[width=\imagewidth]{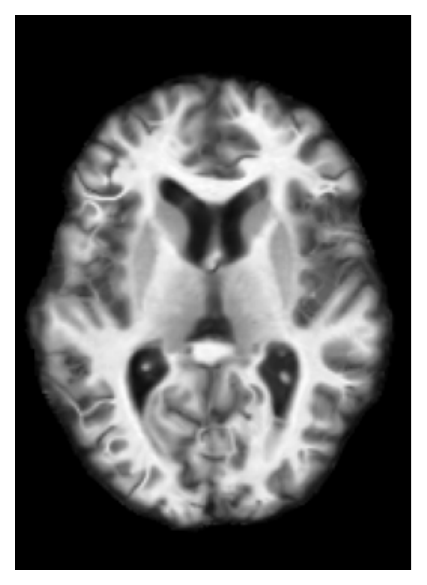}
\includegraphics[width=\imagewidth]{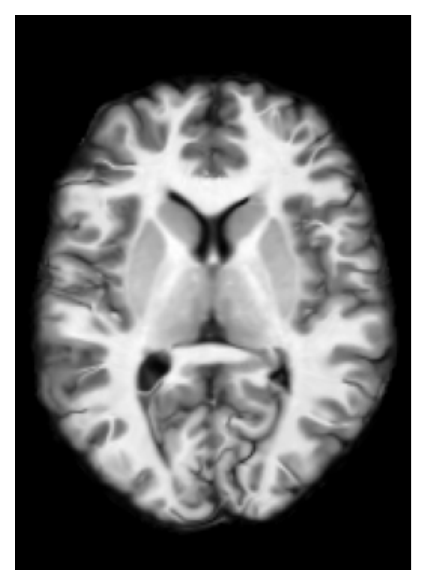}
\includegraphics[width=\imagewidth]{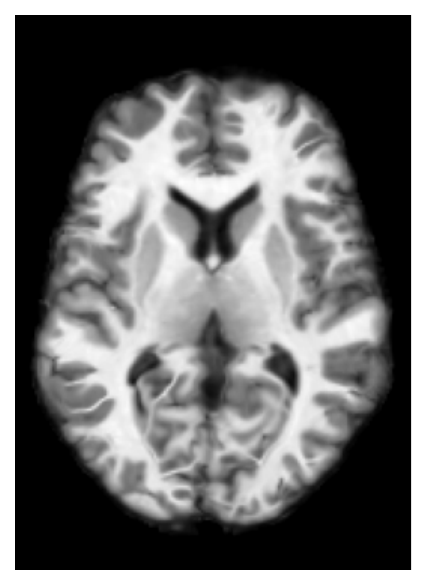}
\includegraphics[width=\imagewidth]{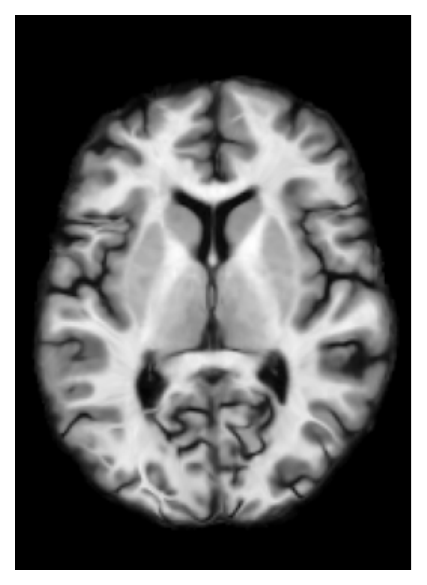}
\includegraphics[width=\imagewidth]{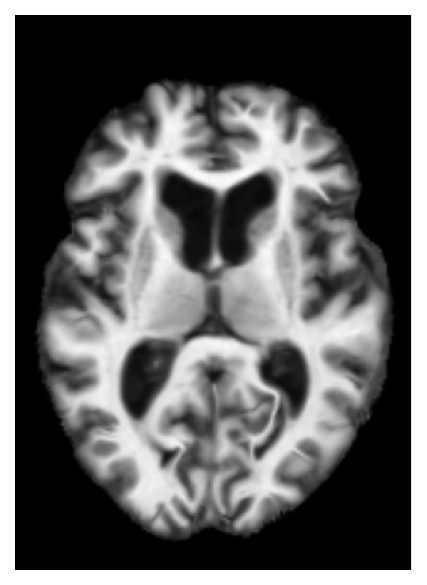}

\hrule

\rotatebox{90}{\adjustbox{valign=m}{\hspace{0.5em}Fixed Image}}
\includegraphics[width=\imagewidth]{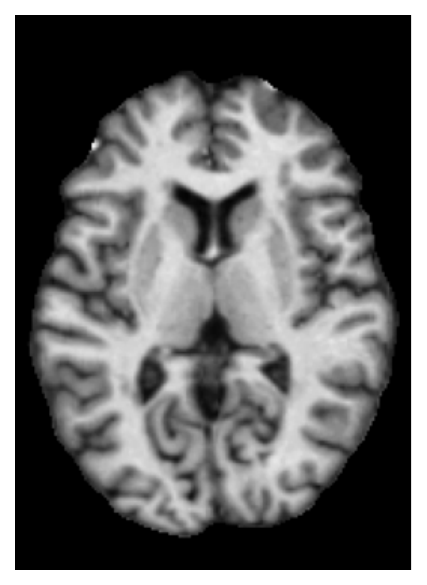}
\includegraphics[width=\imagewidth]{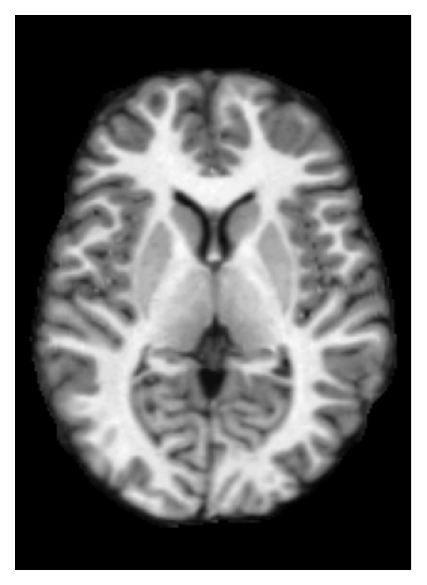}
\includegraphics[width=\imagewidth]{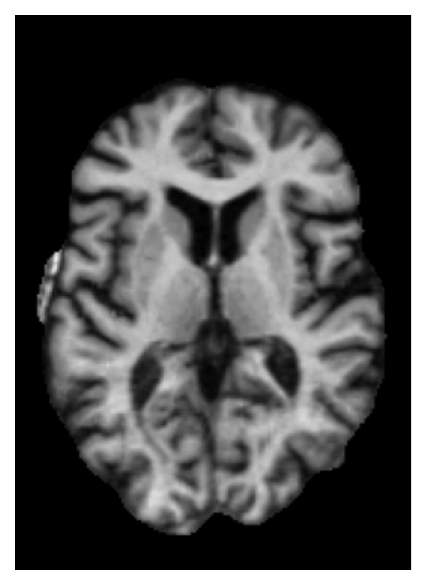}
\includegraphics[width=\imagewidth]{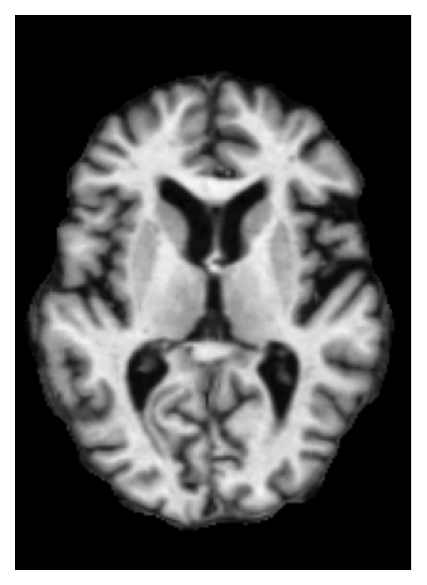}
\includegraphics[width=\imagewidth]{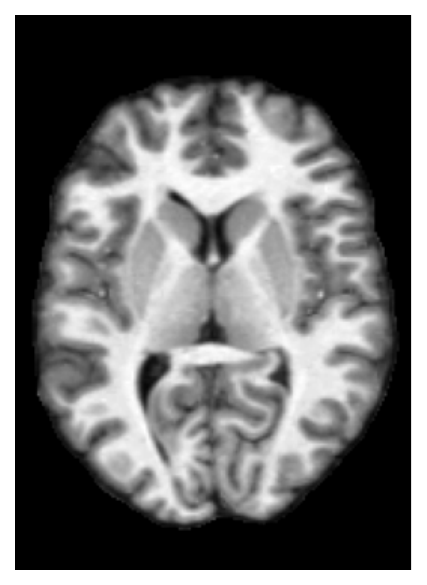}
\includegraphics[width=\imagewidth]{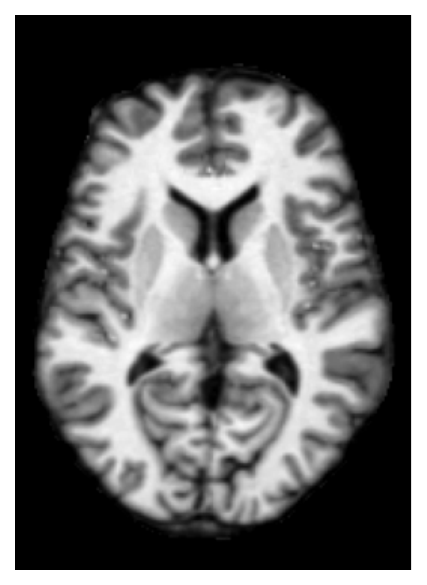}
\includegraphics[width=\imagewidth]{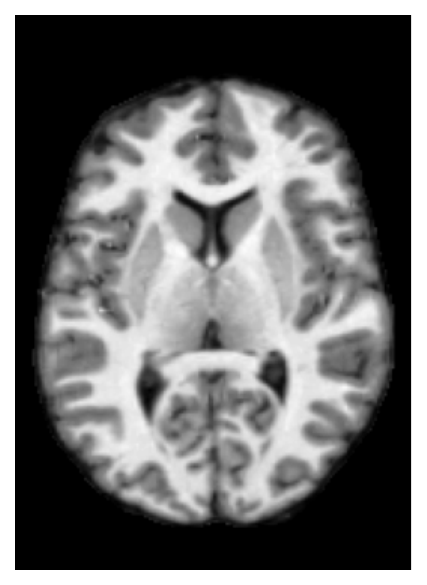}
\includegraphics[width=\imagewidth]{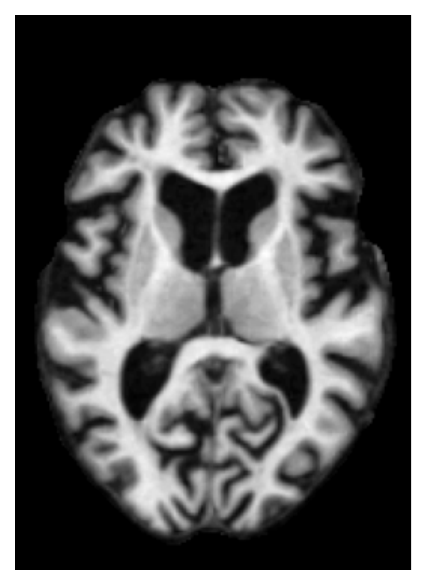}

\rotatebox{90}{\adjustbox{valign=m}{\hspace{0.5em}Moving Image}}
\includegraphics[width=\imagewidth]{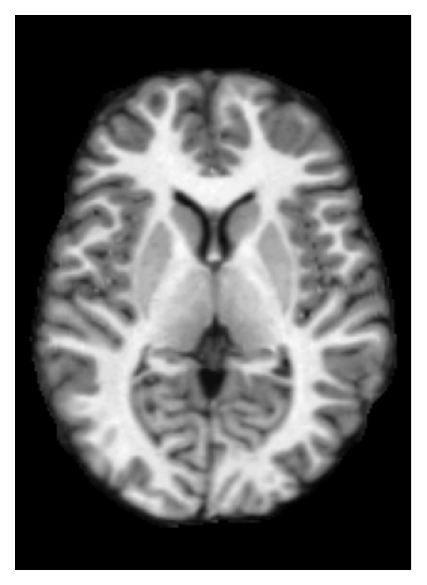}
\includegraphics[width=\imagewidth]{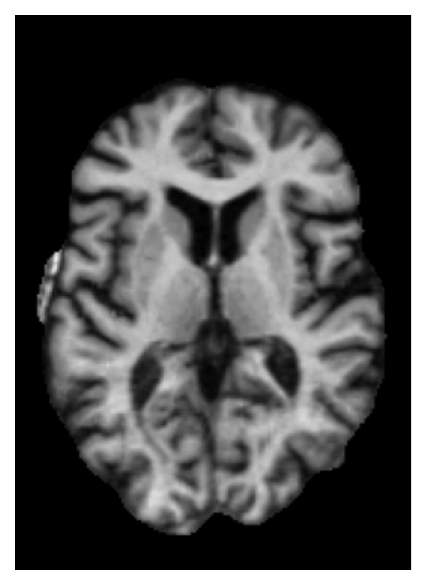}
\includegraphics[width=\imagewidth]{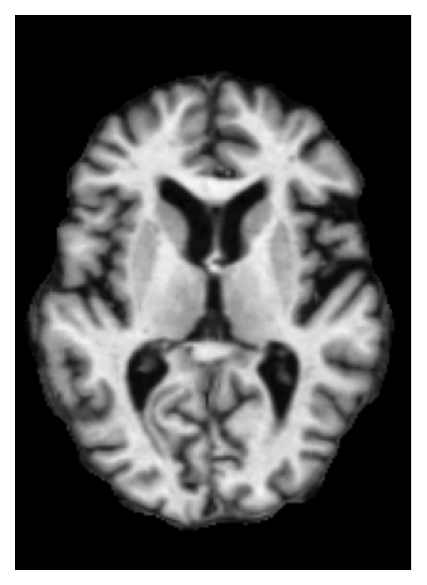}
\includegraphics[width=\imagewidth]{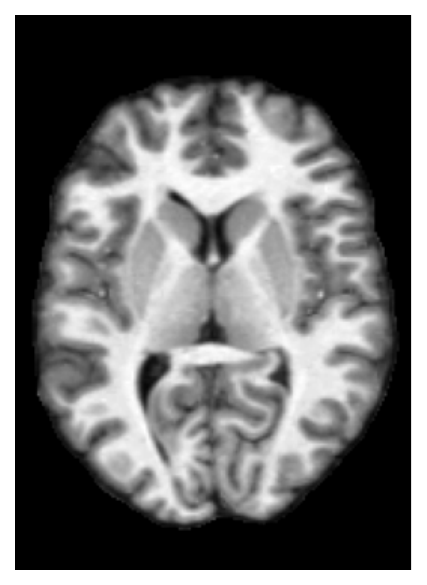}
\includegraphics[width=\imagewidth]{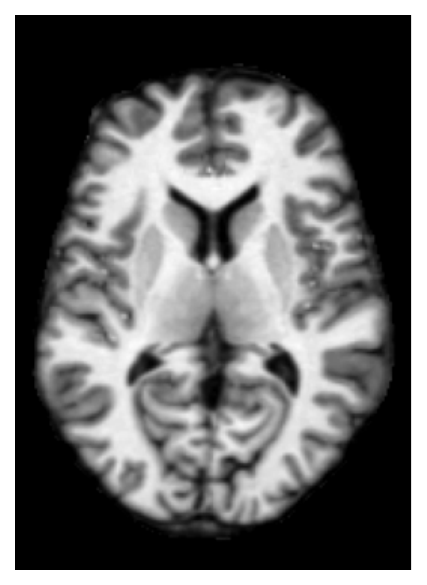}
\includegraphics[width=\imagewidth]{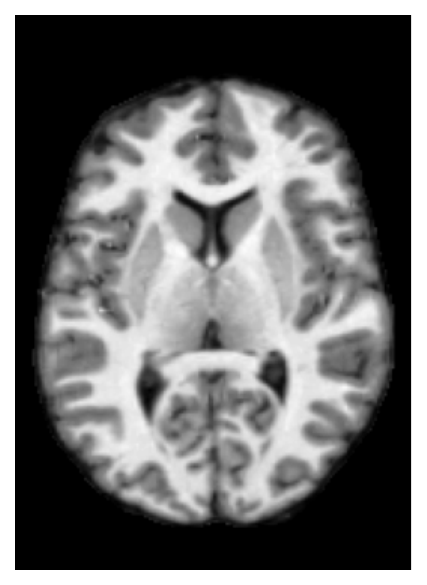}
\includegraphics[width=\imagewidth]{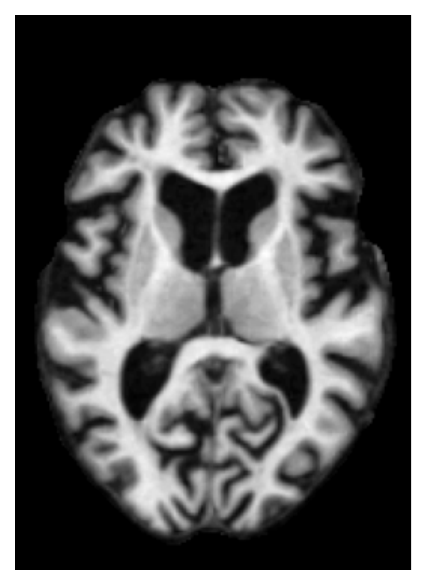}
\includegraphics[width=\imagewidth]{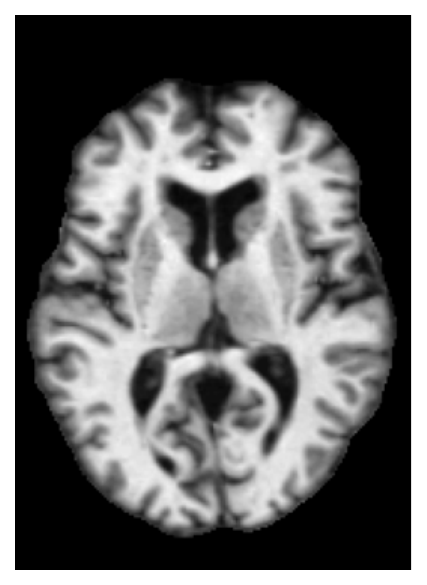}

\caption{\textbf{Qualitative comparison of KeyMorph and our method on OASIS dataset.} The first row shows the warped images using KeyMorph and the second row shows the warped images using our method. The third and fourth rows show the fixed and moving images, respectively.
The OASIS dataset consists of skull-stripped T1-MRI brains that are affinely registered to the Talairach space, consequently we focus on deformable registration.
KeyMorph uses 512 keypoints to parameterize a thin-plate spline transformation, while our method uses an optimizer to predict a dense deformation field. Our method demonstrates high fidelity registration, compared to KeyMorph that only partially warps large differences in ventricles (last two columns). More qualitative comparisons, including segmenation maps, and predicted warp fields are shown in \cref{fig:app-qual-oasis,fig:app-qual-oasis-2,fig:app-qual-warp}.
\label{fig:qual-oasis}
}

\end{figure*}

\edit{\paragraph{Key differences with closely related methods}
The closest works to our method are \textbf{(a)} KeyMorph that emulates feature learning from images for (non-iterative) registration, and \textbf{(b)} GradIRN, RWC-Net, and SUITs that emulate iterative optimization with explicit recurrent modules.
Using a framework like KeyMorph limits the warp representation that can be computed using closed-form solutions like affine, or thin-plate splines (TPS).
TPS represents a very limited class of warps, cannot be guaranteed to be diffeomorphic, and a vast majority of widely used parameterizations (free-form, SVF, geodesic, LDDMM, SyN) do not admit closed form solutions rendering KeyMorph unsuitable for many advanced registration applications.
We compare the qualitative expressivity of the warp field and transformed images generated by KeyMorph with that of our method in \cref{fig:qual-oasis,fig:app-qual-warp}.
Explicit recurrent modules, on the other hand, are stateful and are limited to few iterations due to memory constraints.
This also limits the expressivity of the generated warp fields despite not being limited to closed-form solutions.
Moreover, we note that KeyMorph is highly compute-intensive, quickly running out of memory on an A6000 GPU with 512 keypoints, even with a truncated UNet backbone and float16 mixed precision training.
GradIRN, RWC-Net, and SUITs face memory contraints because of their explicit recurrent modules, and are limited to a few iterations.
On the other hand, {\method} produces dense multi-scale image features, which would equivalently correspond to about $192 * 224 * 160 * (1 + 1/8 + 1/64) * 16 / 3 \sim 41$ million keypoints for a standard MRI image across multiple scales, and can be run for a hundreds of iterations without memory constraints.
This allows us to express maximal expressivity both in the feature representation and the capacity of the optimization solver.
}

\begin{table}[ht]
\centering
\caption{\textbf{Quantitative performance on OASIS and NLST validation sets.}
{\method} learns high-fidelity features incorporating both image and label matching into iterative optimization, showing superior performance compared to a variety of baselines.
}
\label{fig:oasis}
\resizebox{\linewidth}{!}{%
\begin{tabular}{lcc}
\hline
\multicolumn{3}{c}{\textbf{Validation metrics on OASIS}} \\ \hline
\textbf{Method} & \textbf{Dice} & \textbf{HD95} \\ \hline
Affine (Baseline)  & 0.572 $\pm$ 0.051 & 3.831 $\pm$ 0.718 \\
\hline
ANTs~\cite{avants_symmetric_2008} & 0.786 $\pm$ 0.033 & 2.209 $\pm$ 0.534 \\
NiftyReg~\cite{niftyreg} & 0.775 $\pm$ 0.029 & 2.382 $\pm$ 0.723 \\
LogDemons~\cite{vercauteren_symmetric_2008} & 0.804 $\pm$ 0.022 & 2.068 $\pm$ 0.448\\
FireANTs~\cite{jena2024fireants} & 0.791 $\pm$ 0.028 & 2.793 $\pm$ 0.602 \\ 
SynthMorph~\cite{hoffmann2021synthmorph} & 0.785 $\pm$ 0.023 & 2.311 $\pm$ 0.452 \\ \hline
\edit{ConvexAdam + intensity}~\cite{convexadam} & \edit{0.792 $\pm$ 0.030} & \edit{2.710 $\pm$ 0.555} \\ 
\edit{DINO-reg}~\cite{dinoreg} & \edit{0.509 $\pm$ 0.031} & \edit{5.667 $\pm$ 0.638} \\
\edit{Cyclic-Reg}~\cite{bigalke2023unsupervised} & \edit{0.763 $\pm$ 0.033} & \edit{2.539 $\pm$ 0.723} \\
\edit{GradIRN}~\cite{gradirn} & \edit{0.746 $\pm$ 0.016} & \edit{8.232 $\pm$ 0.715} \\
\edit{SUITs}~\cite{suits} & \edit{0.615 $\pm$ 0.047} & \edit{3.923 $\pm$ 0.498} \\
KeyMorph (MSE) & 0.608 $\pm$ 0.039 & 3.886 $\pm$ 0.458 \\
KeyMorph (Dice) & 0.642 $\pm$ 0.021 & 3.560 $\pm$ 0.394  \\
Ours (UNet backbone) & \textbf{0.853 $\pm$ 0.018} & \textbf{1.675 $\pm$ 0.379} \\
Ours (LKU backbone) & \textbf{0.862 $\pm$ 0.017} & \textbf{1.584 $\pm$ 0.351} \\ \hline
\end{tabular} 
}
\vspace{0.5pt}
\resizebox{\linewidth}{!}{
\begin{tabular}{lc}
\hline
\multicolumn{2}{c}{\textbf{Validation metrics on NLST}} \\ \hline
\textbf{Method} & \textbf{TRE30 (in mm)} \\ \hline
Zero displacement (Baseline) & 9.76 \\
VoxelMorph~\cite{balakrishnan2019voxelmorph} & 4.12 \\
Im2Grid~\cite{im2grid} & 3.05 \\
SyN & 3.04 \\ 
Vector-Field Attention~\cite{vfa} & 2.31 \\
RWC-Net~\cite{rwcnet} & 2.11 \\
unigradICON~\cite{tian2024unigradicon} & 2.07 \\
unigradICON + instance optimization & 1.77 \\
FireANTs & 1.28 \\ 
FireANTs + MIND & 1.18 \\
ConvexAdam + MIND & 1.17 \\
Ours + MIND & \textbf{1.02} \\ \hline
\end{tabular}
}
\end{table}

\subsection{{\method} inherits robustness to domain shift from iterative optimization}
\label{sec:ood}

A key requirement of registration algorithms is to be robust to a spectrum of scanner configurations, acquisition, preprocessing and labelling protocols, since there are different standards across institutions.
Existing prediction-based DLIR methods are very sensitive to domain shift~\cite{neuralinstanceopt,jena2024deep,jian2024mamba}, and catastrophically fail on other brain datasets.
On the contrary, {\method} inherits the domain agnosticism of the optimization solver, and is robust under feature distortions introduced by domain shift.

\textbf{Datasets}
We evaluate the robustness of the trained models on three brain datasets: LPBA40, IBSR18, and CUMC12 datasets~\cite{shattuck2008construction,ibsr,klein_evaluation_2009}.
Contrary to the OASIS dataset, these datasets were obtained on different scanners, affinely pre-aligned to different atlases (MNI305, Talairach) with varying algorithms used for skull-stripping, bias correction (BrainSuite, autoseg), and different \textit{manual} labelling protocols for different anatomical regions (as opposed to automatically generated Freesurfer labels in OASIS).
Unlike the OASIS dataset, these datasets also have different voxel sizes for different brain scans, and IBSR18 and CUMC12 datasets have non-uniform anisotropic volumes. %
More details about the datasets are provided in \cref{app:datasets}.
\edit{
We note that increasing label map overlap with automatically generated labels during training is easier for DL-based parametric registration methods. 
Therefore, the performance of DL-based methods on \textit{unseen, manually generated} parcellations is crucial for clinical translation.
}
\edit{The aforementioned aspects of the chosen community-standard datasets make them challenging for DLIR methods, and highlight the crucial shortcoming of these methods, i.e. lack of generalization to domain shift.
}

\textbf{Baselines}
We employ a variety of deep learning baselines for this experiment.
We consider the original VoxelMorph~\cite{balakrishnan2019voxelmorph} pretrained model that is trained using an unsupervised objective function, SynthMorph~\cite{hoffmann2021synthmorph} that is trained on procedurally generated synthetic data using upsampled Perlin noise.
\edit{Cyclical-Reg~\cite{bigalke2023unsupervised} is similar to SynthMorph in that it is trained on a self-supervised objective without any label or image supervision.
The training framework emulates a few consistencies of the predicted warp field like inverse-consistency and matching the results of iterative optimization.
}
Furthermore, two pyramidal architectures that mimic multi-scale prediction - LapIRN~\cite{mok2020large} and its conditional counterpart named Conditional LapIRN~\cite{mok2021conditional} are also suitable prediction-based baselines.
A symmetric normalization network dubbed SymNet~\cite{mok2020fast} that performs symmetric predictions from the fixed and moving images is also used to compare with their non-symmetric counterparts.
The pretrained models in SymNet and LapIRN are trained without dice loss; we also train models that include dice loss for comparison.
We also include a large kernel UNet (LKU)~\cite{lku} which has showed high accuracy in the OASIS dataset, albeit with implausible deformations~\cite{jian2024mamba}.
We also consider three variants of transformer-based TransMorph for registration~\cite{chen_transmorph_2022}.
Specifically, we use the provided pretrained model for \textit{TransMorph-large} and two variants of \textit{TransMorph-regular} trained with and without Dice loss.
\edit{Finally, we consider ConvexAdam, DINO-reg, multimodalSUITs, and GradIRN as baselines employing iterative optimization.
}

This assortment of baselines represent a spectrum of design choices in deep learning for registration, and are representative of the state-of-the-art in DLIR.
\edit{These methods show excellent performance on in-distribution datasets with automatically generated parcellations.}
To evaluate the generalization to out-of-distribution datasets, we train all models on the OASIS training split and evaluate on all pairs of the LPBA40, IBSR18, and CUMC12 datasets.

Owing to the predictive paradigm of most baselines, we also evaluate their performance with and without instance optimization. 
Following VoxelMorph++~\cite{vxmpp}, we finetune the output representation for 100 iterations with the normalized cross-correlation (NCC) loss, and Adam optimizer with a learning rate of $10^{-3}$.
Note that almost none of these baselines come with instance optimization postprocessing, therefore we manually implement, evaluate and validate the performance of the instance optimization solver for each baseline, requiring significant effort.

\textbf{Evaluation}
We evaluate across a variety of configurations -- (i) preserving the anisotropy of the volumes or resampling them to 1mm isotropic (denoted as \textit{anisotropic} or \textit{isotropic} respectively), and (ii) center-cropping the volumes to match the size of the OASIS dataset (denoted as \textit{Crop} and \textit{No Crop}).
The results for all three datasets are shown in \cref{fig:robustness} sorted by mean Dice score; quantitative comparison is also shown in Appendix \cref{tab:robustness}.
\cref{fig:robustness} shows boxplots with each color representing a different method, and a more translucent shade for the baseline without instance optimization.
Note that TransMorph, VoxelMorph, and SynthMorph do not work for sizes that are different than the OASIS dataset due to design decisions and implementation constraints, therefore they only work in the \textit{Crop} setting.
The IBSR18 dataset consists of different volumes with different levels of anisotropy; consequently resampling them to 1mm isotropic leads to different voxel sizes.
These volumes cannot be concatenated along the channel dimension, consequently every DLIR method cannot run under this configuration (\cref{fig:robustness}(a)). %
In contrast, similar to KeyMorph, our method employs a dual-stream-like architecture that processes one volume at a time.
Since our method utilizes a dual-stream-like convolutional architecture processing one volume at a time, the fixed and moving images can have different voxel sizes, i.e. \textbf{feature extraction is not contingent on the voxel sizes of the moving and fixed images being equal}.
The optimization solver can also handle different voxel sizes for the fixed and moving volumes -- which is useful in applications like multimodal registration (in-vivo to ex-vivo, histology to 3D, pre-operative to intra-operative, microscopy to MRI). %
This unprecedented flexibility brings forth a new operational paradigm in deep learning for registration combining feature learning to incorporate label fidelity with optimization-as-a-layer to be robust, widening the scope of applications for registration with deep features.
This experiment provides a few key insights \edit{about existing DLIR methods}. 

\begin{figure*}[htpb!]
    \includegraphics[width=\linewidth]{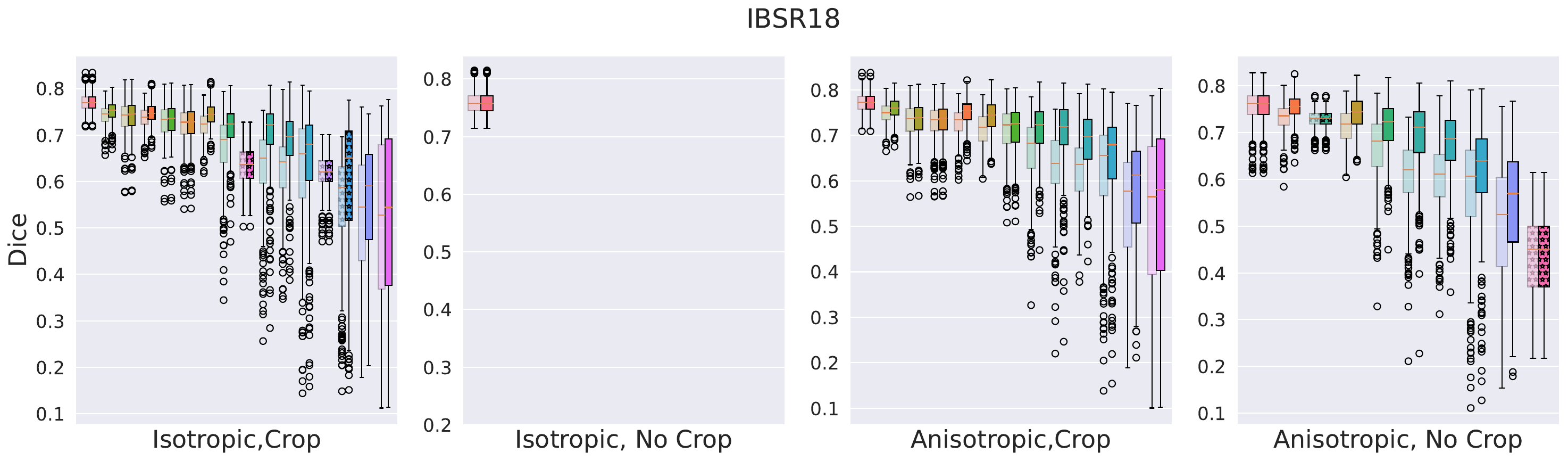}
    \includegraphics[width=\linewidth]{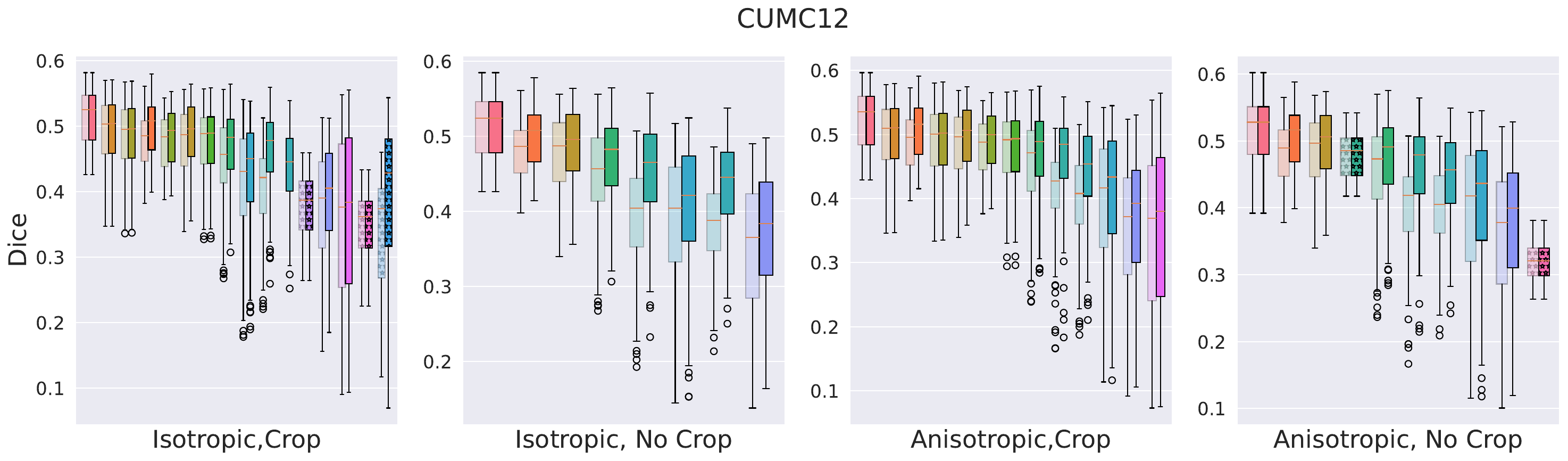}
    \begin{minipage}{0.52\linewidth}
        \includegraphics[width=\linewidth]{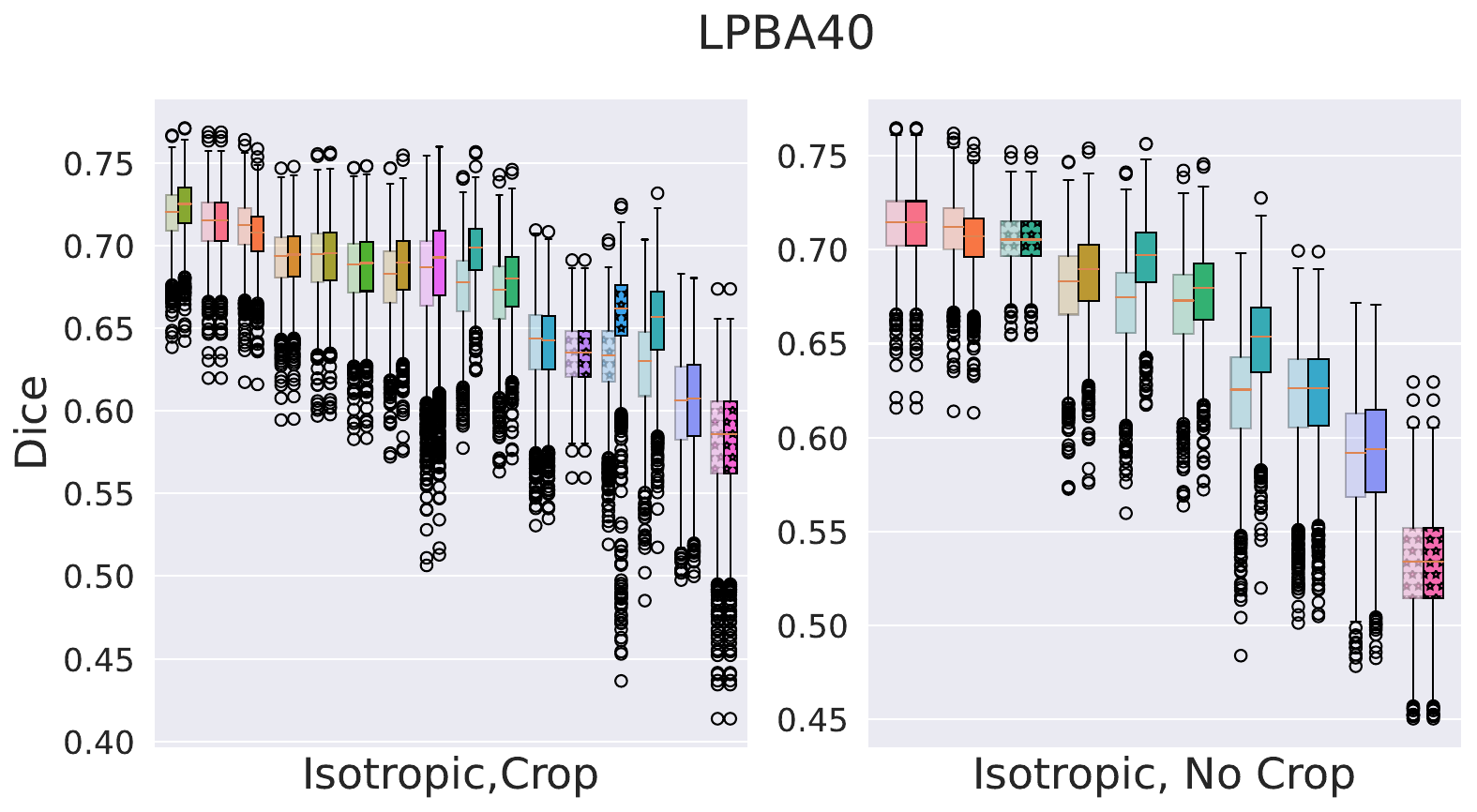}
    \end{minipage}
    \begin{minipage}{0.48\linewidth}
        \centering
        \includegraphics[width=0.7\linewidth]{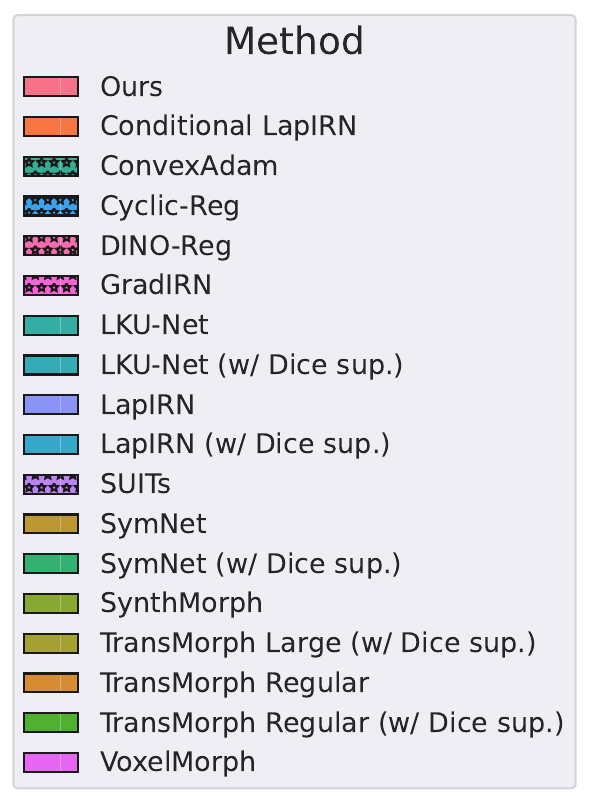} 
    \end{minipage}
    \caption{\footnotesize \textbf{Boxplots of Dice scores for three out-of-distribution datasets.} {\method} performs significantly better across three datasets without additional finetuning. Contrary to other baselines that output warp fields considering 1mm isotropic data, leading to a performance drop with anisotropic volumes, {\method} performs better with anisotropic data due to the optimization's resolution-agnostic nature. Even with image-space instance optimization, almost all baselines underperform compared to {\method}.}
    \label{fig:robustness}
\end{figure*}

\subsubsection{Predictive registration methods do not generalize their performance under domain shift}
Image registration is a highly ill-defined and non-convex problem, which is NP-hard to solve in general.
Learning a \edit{parametric} statistical model to amortize optimization can learn a distribution of warps that are specific to the training dataset.
However, there is no explicit mechanism to ensure that the predicted warp field indeed performs correspondence in \textit{any} space of  feature maps. 
For domain shift in the input images, the warp fields predicted by the model need not be the local minima of \textit{any} optimization function.
This implies that predictive methods for registration would not easily generalize outside the training domain.
Moreover, this lack of generalization is not mitigated by label supervision during the training phase, as evident by baselines with supervised label losses underperforming their unsupervised counterparts.
This behavior is not noticed by us alone; ~\cite{mok2022affine} observe that the supervised models are inferior to their unsupervised models in the LPBA dataset, indicating anatomical knowledge injected to the model with supervision may not generalize well to unseen data beyond the training data.
The need for instance optimization (IO) for improved performance is shown to be necessary for foundational models as well~\cite{tian2024unigradicon}.
The benefit of amortized optimization does not hold anymore since IO becomes a necessity and consequently a bottleneck for generalization to domain shift.
In fact, most of the inference time is now dominated by the (sequential) IO routine. 
However, instance optimization routines have become fast, motivating a shift towards robust feature learning paradigm instead.

\subsubsection{DLIR methods do not provide good initialization for out-of-distribution data} 
Despite the need for instance optimization, one may want to use predictive registration methods for initialization to reduce the number of iterations required for the subsequent instance optimization.
However, predictive methods do not provide good initialization either, as the performance of baselines does not surpass our method even with 100 iterations at the finest scale, compared to only 20 iterations at the finest scale for our method in ~\cref{fig:robustness}.
If the initialization is downsampled to perform multi-scale instance optimization, most of the initialization information is lost during downsampling.
For example, if a multi-scale instance optimization is performed with the coarsest scale at 1/4$^{\text{th}}$ resolution, around \textbf{98.4\%} ($= 63/64$) of the initialization is discarded. 
This kind of instance optimization then closely resembles classical intensity-based optimization instead, rendering the initialization from predictive methods redundant.
Another limitation of instance optimiztion is also observed in ~\cite{neuralinstanceopt} wherein instance optimization typically achieves minimal improvements on solidly trained neural networks.
For out-of-distribution data, our experiments also corrobate the fact that initialization from learned coarser feature maps (ours) is consistently robust compared to initialization from predictive methods.

\subsubsection{{\method} remedies both these issues using high-fidelity multi-scale features}
\begin{figure}[ht!] %
    \centering
    \includegraphics[width=\linewidth]{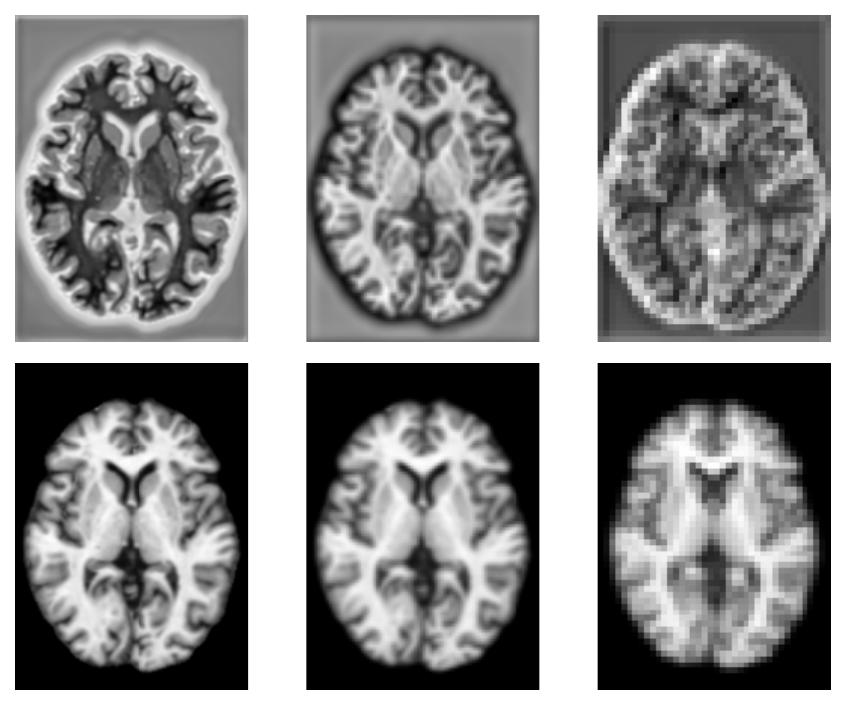}
    \caption{\footnotesize Examples of multi-scale features learned by the feature extractor. Scale-space features (\textit{bottom row}) obtained by downsampling the image downsample all image features indiscriminately. Our features (\textit{top row}) preserve necessary anatomical information at all scales, and introduce inhomogenity in the feature space for better optimization (watershed effect and enhanced contrast near gyri and a halo around the outer surface to delineaate background from gray matter). }
    \label{fig:multiscalefeatures}
\end{figure}

Under our feature learning paradigm, we are able to circumvent the bad initialization problem by not predicting any warps at all, and instead performing a multi-stage instance optimization with learned features.
\cref{fig:multiscaleablation,fig:multiscalefeatures} show that our learned feature maps provide higher-fidelity warps compared to intensity images at all levels, while being interpretable.
Since most of the iterative computation is performed at the coarser scales, this leads to fast runtimes than baselines with instance optimization.
{\method} also provides robust performance and low variance across different datasets, as shown in \cref{fig:robustness}.
Our novel methodology sidesteps initialization using prediction altogether. %

\subsection{Robust feature learning enables zero-shot performance by switching optimizers at test-time}
\label{sec:zeroshot}

Another major advantage of our framework is that we can switch the optimizer \textit{at test time} without any retraining.
This is useful when the registration constraints evolve over time (i.e. initially diffeomorphic transforms were required but now non-diffeomorphic transforms are acceptable), or when the registration is used in a pipeline where different parameterizations (freeform, diffeomorphic, geodesic, B-spline) may be compared.
Since our framework decouples the feature learning from the optimization, we can switch the optimizer arbitrarily at test time, at no additional cost.
A crucial requirement is that learned features should not be too sensitive to the instance optimization routine. 

\begin{table*}[htpb!]
\centering
\resizebox{1.02\linewidth}{!}{%
\begin{tabular}{lccc|ccc}
\hline
\textbf{Optimizer} & \multicolumn{3}{c}{\textbf{SGD}} & \multicolumn{3}{c}{\textbf{FireANTs (diffeomorphic)}} \\
\textbf{Architecture} & \textbf{DSC} & \textbf{HD95} & {$\mathbf{\% (\| J\| < 0)}$} & \textbf{DSC} & \textbf{HD95} & {$\mathbf{\% (\| J\| < 0)}$} \\
\hline
UNet Encoder & 0.845 $\pm$ 0.018 & 1.790 $\pm$ 0.433 & 0.7866 $\pm$ 0.1371 & 0.834 $\pm$ 0.018 & 1.847 $\pm$ 0.410 & 0.0000 $\pm$ 0.0000 \\
LKU Encoder & 0.849 $\pm$ 0.018 & 1.733 $\pm$ 0.401 & 0.8079 $\pm$ 0.1308 & 0.838 $\pm$ 0.018 & 1.806 $\pm$ 0.373 & 0.0000 $\pm$ 0.0000 \\
UNet & 0.853 $\pm$ 0.018 & 1.675 $\pm$ 0.379 & 1.0718 $\pm$ 0.1662 & 0.842 $\pm$ 0.018 & 1.748 $\pm$ 0.397 & 0.0000 $\pm$ 0.0000 \\
LKU & 0.862 $\pm$ 0.017 & 1.584 $\pm$ 0.351 & 0.8646 $\pm$ 0.1429 & 0.849 $\pm$ 0.017 & 1.740 $\pm$ 0.345 & 0.0000 $\pm$ 0.0000 \\ \hline
\end{tabular}
}
\caption{\textbf{Zero shot performance by switching optimizers at test-time}. Our method is trained on the OASIS dataset with the SGD optimizer to obtain the warp field. At inference time, we use an SGD optimizer for no constraint on the warp field, and the FireANTs optimizer to ensure diffeomorphic warps. Across all architectures, the Dice Score remains robust, with only a slight dip attributed to the constraints introduced by diffeomorphic mappings. The SGD optimization introduces $\sim$1\% singularities, while FireANTs shows no singularities.}
\label{tab:diffeo}
\end{table*}

\begin{figure}[ht!] 
    \centering
    \includegraphics[width=\linewidth]{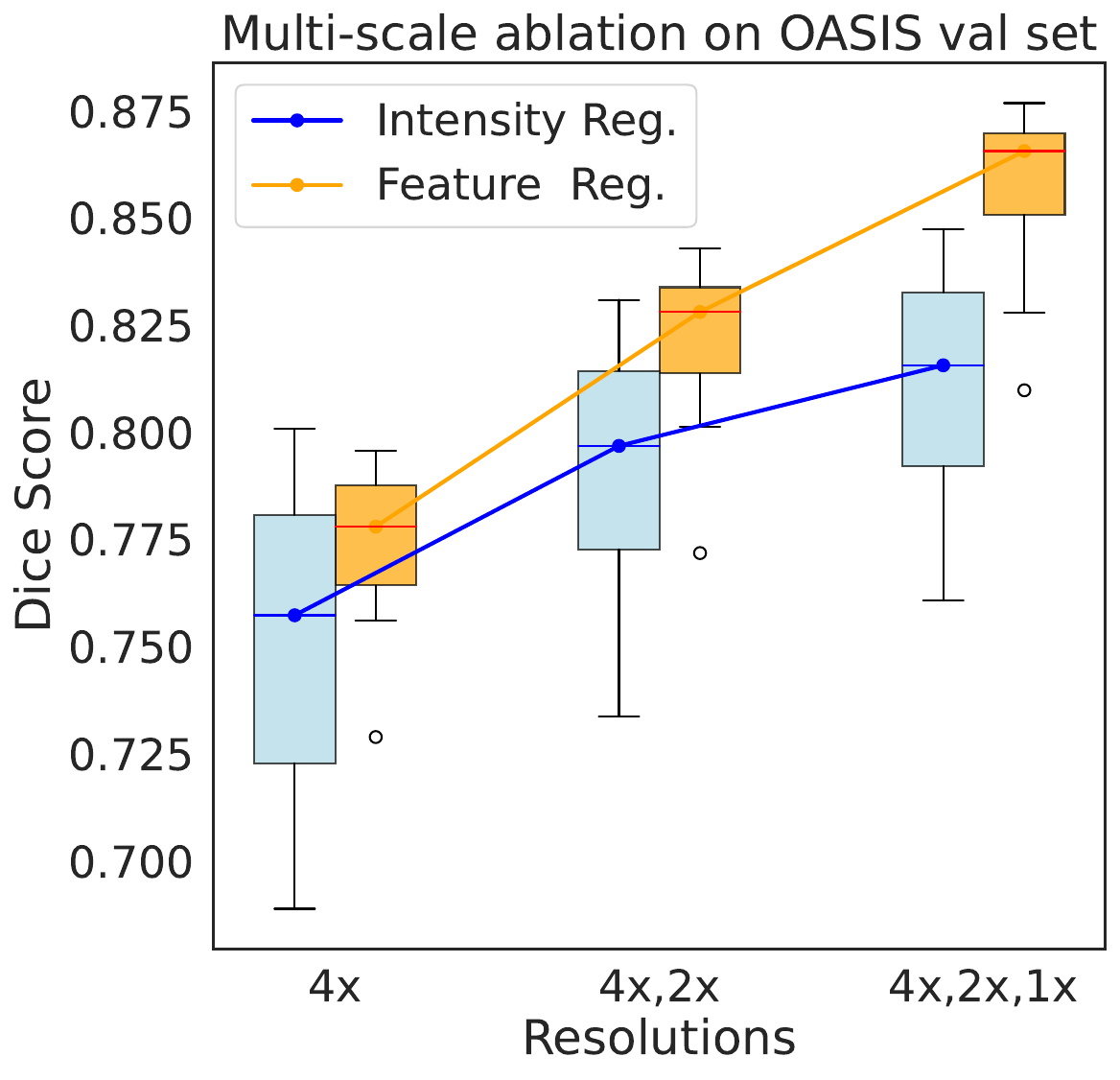}
    \caption{
        \textbf{Ablation on fidelity of multi-scale features compared to multi-scale intensity images}. To show that multi-scale features provide more label-aware information than intensity images alone, we perform registration on the OASIS validation set using multi-scale features and intensity images. 
        For intensity-based multi-scale registration, the intensity images are smoothed and downsampled at each level.
        x-axis shows the resolutions at which optimization is performed, and y-axis shows the distribution of Dice scores.
        For identical multi-scale optimization routines, feature-based registration provides better label alignment than intensity images at all resolutions.
        This demonstrates the efficacy of task-awareness in features learned using our framework.
    }
    \label{fig:multiscaleablation}
\end{figure}

\begin{figure*}[ht!] 
    \centering
    \begin{minipage}{0.48\linewidth}
        \centering
        \includegraphics[width=\linewidth]{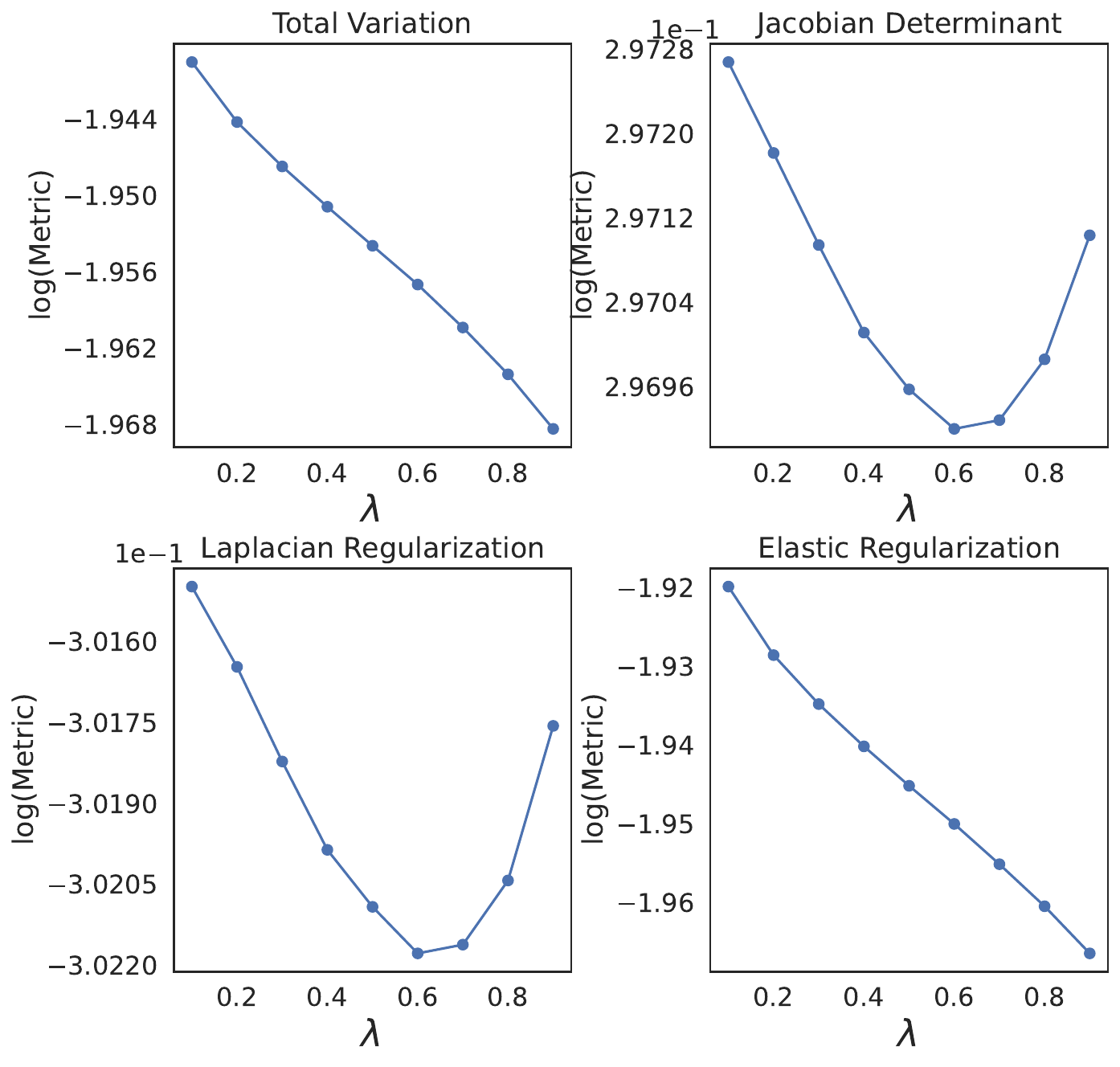}
        \subcaption{\textbf{Effect of $\lambda$ on different $R_u$ with HyperMorph}.
        }
        \label{fig:hyperparam_ablation_hyp}
    \end{minipage}
    \begin{minipage}{0.48\linewidth}
        \centering
        \includegraphics[width=\linewidth]{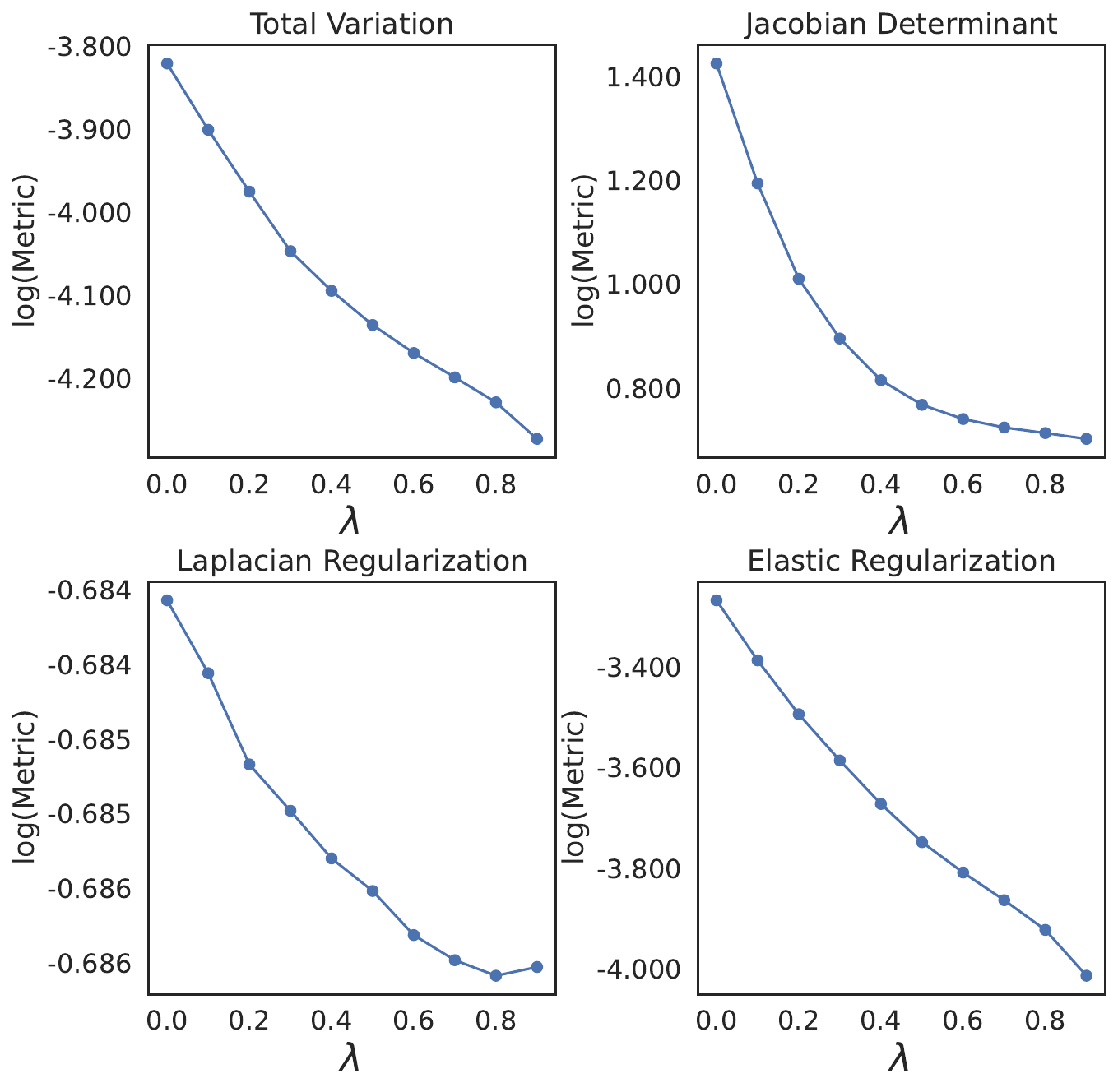}
        \subcaption{\textbf{Effect of $\lambda$ on different $R_u$ with {\method}}.
        }
        \label{fig:hyperparam_ablation_our}
    \end{minipage}
    \caption{
        \textbf{Comparison of regularization at inference time}. 
        With HyperMorph, regularizations like Volume Preservation and Laplacian Registration are not monotonic with the training hyperparameter $\lambda$, and have to be considered during training.
        In contrast, due to the decoupled feature learning and optimization, {\method} can be run with arbitrary regularization families at test time without any retraining, and monotonic trends with $\lambda$ are observed.
    }
    \label{fig:hyperparam_ablation}
\end{figure*}

To demonstrate this functionality, we use the validation set of the OASIS dataset and four network architectures.
We consider the vanilla UNet~\cite{ronneberger2015u} and Large Kernel UNet~\cite{lku} networks, and Encoder-only and Encoder-Decoder architectures for each network.
The difference in architectures are visualized in \cref{fig:architectures}.
These networks were initially trained using the SGD optimizer without any additional constraints on the warp field.
At test time, we switch the optimizer to the FireANTs optimizer ~\cite{jena2024fireants}, that uses a Riemannian Adam optimizer for multi-scale diffeomorphisms.
If the features had overfit to the training dynamics of the SGD optimizer, we would expect a significant drop in performance at test time.
\edit{Unlike explicit iterative unrolling, implicit optimization theoretically ensures that the gradient of the inputs to the solver is \textit{independent} of the optimization path, and is only dependent on the final result of the solver.}

Results in \cref{tab:diffeo} compare the Dice score, 95th percentile of the Haussdorf distance (denoted as \textit{HD95}) and percentage of volume with negative Jacobians (denoted as $\% (\| J\| < 0)$) for the two optimizers.
The SGD optimizer introduces anywhere from $0.79\%$ to $1.1\%$ of singularities in the registration, while the FireANTs optimizer does not introduce any singularities.
A slight drop in performance can be attributed to the additional implicit constraints imposed by diffeomorphic transforms.
However, the high-fidelity features lead to a much better label overlap than FireANTs run with image features (\cref{fig:oasis,fig:multiscaleablation}).
Our framework introduces an unprecedented amount of flexibility at test time that is an indispensible feature in deep learning for registration, and can be useful in a variety of applications where the registration requirements change over time, without expensive retraining.

\subsection{Interpretability of features}

Decoupling of feature learning and optimization allows us to examine the feature images obtained at each scale to understand what feature help in the registration task. 
Classical methods use scale-space images (smoothened and downsampled versions of the original image) to avoid local minima, but lose discriminative image features at lower resolutions.
Moreover, intensity images may not provide sufficient details to perform label-aware registration. 
Since our method learns dense features to minimize label matching losses, we can observe which features are necessary to enable label-aware registration.
\cref{fig:multiscalefeatures} highlights differences between scale-space images and features learned by our network.
At all scales, the features introduces heterogeneity using a watershed effect and enhanced contrast to improve label matching performance. 

\subsection{Inference time}
DLIR methods have been very popular due to their fast inference time by performing amortized optimization~\cite{balakrishnan2019voxelmorph}.
Classical methods generally focus on robustness and reproducubility, and do have GPU implementations for fast inference.
However, modern optimization toolkits ~\cite{mang_claire_2019,jena2024fireants} utilize massively parallel GPU computing to register images in seconds, and scale very well to ultrahigh resolution imaging.
A concern with optimization-in-the-loop methods is the inference time.
\cref{tab:inferencetime} shows the inference time for our method for all four architectures.
These inference times are fast for a lot of applications, and the plug-and-play nature of our framework makes {\method} amenable to rapid experimentation and hyperparameter tuning.

\begin{table}[ht] %
    \centering
    \resizebox{\linewidth}{!}{%
    \begin{tabular}{lcc}
        \toprule
        \textbf{Architecture} & \textbf{Neural net} (sec) & \textbf{Optimization} (sec) \\ \hline
        UNet & 0.444 & 1.693 \\
        UNet-E & 0.433 & 1.555 \\
        LKU & 0.795 & 1.463 \\
        LKU-E & 2.281 & 1.457 \\
        \bottomrule
    \end{tabular}
    }
    \vspace{0.5pt}
    \resizebox{\linewidth}{!}{%
        \begin{tabular}{lccc}
            \toprule
            \textbf{Method} & \textbf{Iterations} & \textbf{Time (sec)} & \textbf{Avg. throughput (it/s)} \\ \hline
            SUITS & 15 & 255.211 & 0.058 \\
            GradIRN & 9 & 0.351 & 25.641 \\
            Ours & 350 & \textbf{1.463} & \textbf{239.23} \\
            \bottomrule
        \end{tabular}
    }
    \caption{{\textbf{Inference time for various architectures}. A multi-scale optimization takes only $\sim1.5$ seconds to run all iterations (no early stopping) making it suitable for most applications. This is compared to the time for neural network's feature extraction which is architecture dependent.}}
    \label{tab:inferencetime}
\end{table}

\subsection{{\method} provides flexible Regularization Tuning}

DLIR methods are typically trained with a \textit{fixed} loss function and regularization, leading to inflexible regularization for novel image contrasts, resolutions, or anatomy.
HyperMorph~\cite{hoopes2021hypermorph} introduced a method to amortize optimization over different hyperparameters in a deep network by providing the regularization parameter $\lambda$ as an input to the network. 
The HyperMorph network is trained with the following loss function conditioned on $\lambda$:
\begin{equation}
    \mathcal{C}_\theta(\varphi, \lambda) = (1 - \lambda){L}(I_f, I_m \circ \varphi_\theta) + \lambda R_v(\varphi_\theta)
\end{equation}
where $R_v(\varphi)$ is the total variation on the velocity field of the diffeomorphic transform.
\begin{equation}
    R_v(\varphi_\theta) = \| \nabla v_\theta \|_2^2 \quad , \quad \varphi_\theta = \exp_{\text{Id}}(v_\theta)
\end{equation}
However, the regularization is fixed during training, and a model trained to minimize the total variation may not have similar regularization effects on other unseen regularization families, like Jacobian regularization, curvature, or Laplacian regularizations.
Incorporating $n$ different regularization families would require a combinatorial amount of conditional inputs to capture the full hyperparameter space.
This will require significant training time, and will still be inflexible for other unseen hyperparameter families.
In contrast, our method can work with \textit{arbitrary} unseen regularization families and hyperparameters at test time without any retraining.

To demonstrate this, we consider the pretrained HyperMorph model.
For our method, we perform feature training on \cref{eq:reg} without any regularization, and at inference time, we add a regularization term to the optimization loss as follows:
\begin{equation}
    \mathcal{C}_\theta(\varphi, \lambda) = (1 - \lambda){L}(F_f, F_m \circ \varphi) + \lambda R_u(\varphi)
\end{equation}
We consider four families of $R_u$:
\begin{itemize}
    \item \textbf{Total variation of the warp field}:
    $R_u(\varphi) = \int_{\Omega} \| \nabla \varphi \|_2^2 d\Omega$. We hypothesize that this term will be directly affected by the total variation of the velocity field in HyperMorph, as the exponential map of a smooth velocity field is likely to be smooth, due to the smoothness of the exponential map itself.
    \item \textbf{Elastic reg}:
    $R_u(\varphi) = \int_{\Omega} \left( \alpha \|\nabla \varphi\| + \beta \|\nabla^2 \varphi\| \right)d\Omega$. This term is performed implicitly in the popular SyN algorithm, and is likely to be affected by the total variation energy in HyperMorph as well. We set $\alpha = \beta = 1$ for this experiment.
    \item \textbf{Jacobian det}:
    $R_u(\varphi) = \int_{\Omega} \left( \left| \text{det}(\nabla \varphi) - 1\right|_2^2 \right) d\Omega$. This term is used in diffeomorphic registration to ensure volume preservation, and this term is less likely to have a monotonic relationship with the total variation of the velocity field.
    \item \textbf{Laplacian reg}:
    $R_u(\varphi) = \int_{\Omega} \|\Delta \varphi\|_2^2 d\Omega$. The effects on this regularization are not monotonic with the total variation of the velocity field.
\end{itemize}
For the HyperMorph model, we evaluate the regularization losses for each $\lambda$ to see the effect of $R_v$ on other regularization families $R_u$.
Results in ~\cref{fig:hyperparam_ablation_hyp} show that total variation and elastic regularization follow monotonic trends with $\lambda$ since reducing $\| \nabla v \|_2^2$ will induce smoothness to the velocity field, and consequently smoothness to the warp field due to the smoothness of the exponential map.
However, the Laplacian and Jacobian regularization do not follow monotonic trends with $\lambda$, indicating that additional training would be required to incorporate these regularizations.
In contrast, ~\cref{fig:hyperparam_ablation_our} shows that {\method} can work with arbitrary regularization families at test time without any retraining, providing immense flexibility to arbitrary registration constraints at test time.

\edit{\subsection{Ablation on choice of implicit gradient}
\label{sec:ablationimplicitgradient}
}

\begin{figure*}[ht]
    \centering
    \begin{minipage}{0.3\linewidth}
        \includegraphics[width=\linewidth]{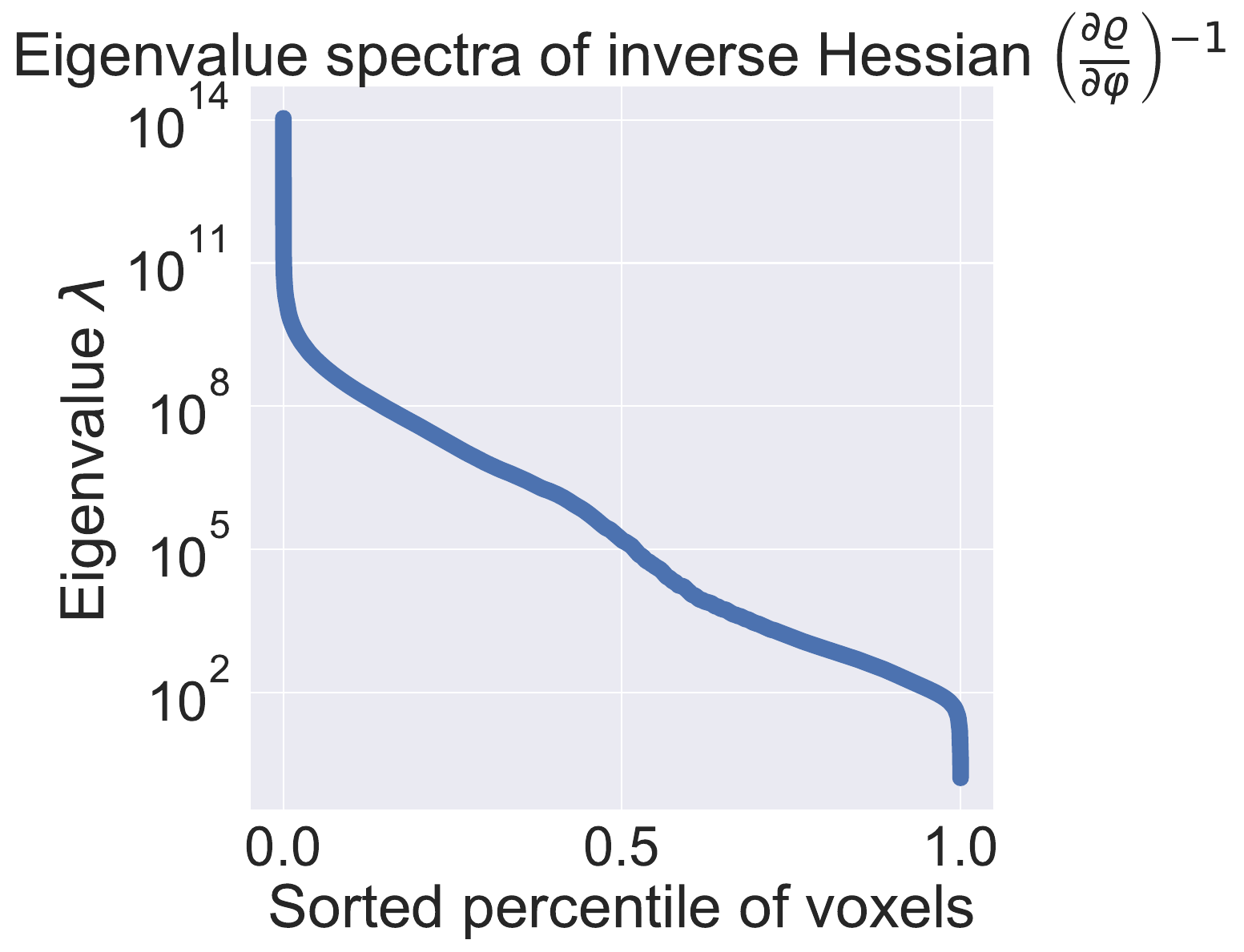}
    \end{minipage}
    \begin{minipage}{0.68\linewidth}
        \includegraphics[width=\linewidth]{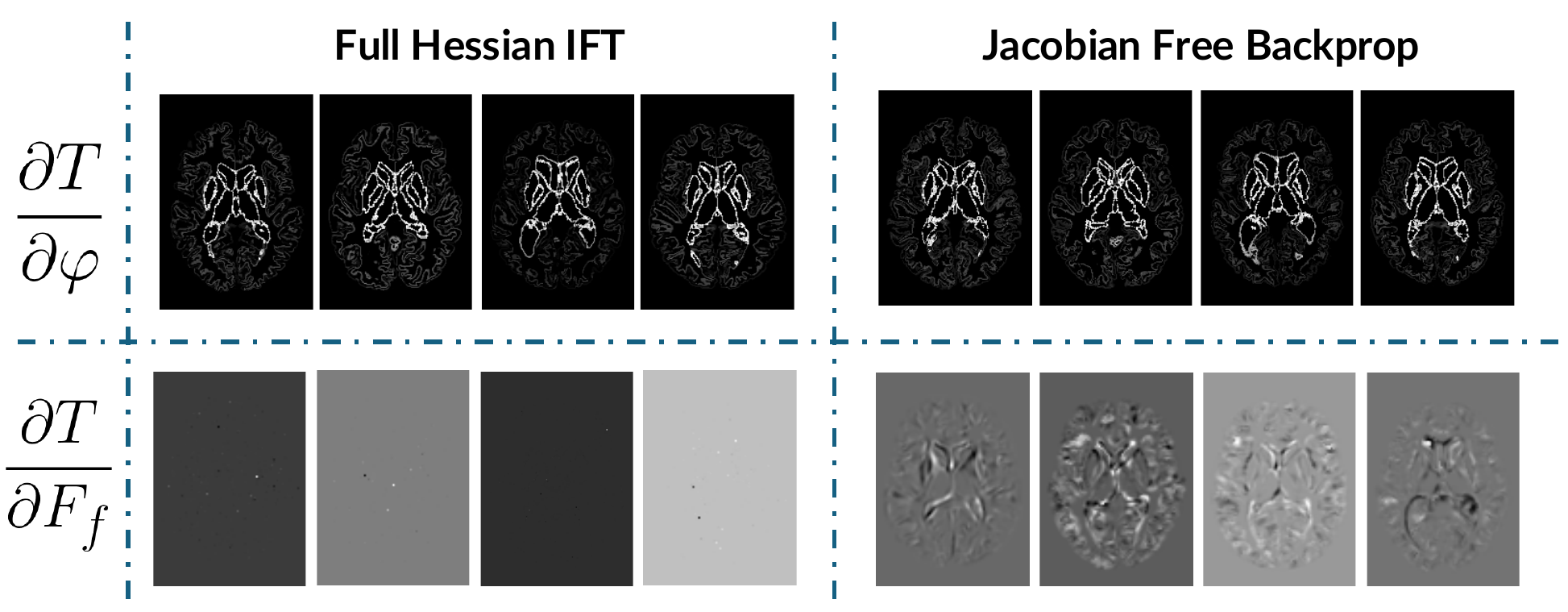}
    \end{minipage}
    \caption{\textbf{Qualitative comparison of different backward passes for DIO}. \textbf{(Left)} Top eigenvalues of the inverse Hessian skew the feature gradient due to their large magnitude compared to the rest of the eigenspectra, \textbf{(Right)} qualitatively demonstrates the effect of the Hessian on the gradient of the training loss with respect to the transformation field $\varphi$ and the fixed feature $F_f$ using different instantiations of the backward pass at the beginning of training. Gradients w.r.t. feature images from Hessian-based IFT are very sparse and do not facilitate network learning. On the contrary, gradients obtained using JFB are dense and the network quickly convergences to low training loss.}
    \label{fig:qual-ift}
\end{figure*}

\edit{In all our experiments, we use the Jacobian-free Backprop (~\cite{fung_jfb_2021}) approximation for approximating the gradient of the feature image.
We ablate on the following choices of implicit gradient approximations: (a) full Hessian, (b) unrolled phantom gradients~\cite{geng_torchdeq_2023,geng2021training} (UPG), and (c) Jacobian-free Backprop, on the OASIS dataset.
Note that phantom gradients simply correspond to BPTT-like unrolling over $k$ steps.
We train the network with the same architecture and hyperparameters for 100 epochs, and evaluate the performance on the validation set.
\begin{table}
    \centering
    \resizebox{\linewidth}{!}{ %
    \begin{tabular}{lc}
        \toprule
        Method & Dice \\
        \midrule
        Full Hessian IFT & 0.688 \\
        Unrolled Phantom Gradients (UPG) ($k=10^*$) & 0.782 \\
        Unrolled Phantom Gradients (UPG) ($k=5$) & 0.841 \\
        Unrolled Phantom Gradients (UPG) ($k=3$) & 0.842 \\
        Jacobian-free Backprop & \textbf{0.862} \\
        \bottomrule
    \end{tabular}
    }
    \caption{\textbf{Ablation on choice of implicit gradient approximation.} On the OASIS dataset, Jacobian-free Backprop achieves highest validation score while being computationally efficient.
    The full Hessian IFT suffers from the ill-conditioned Hessian of the registration problem, leading to poor convergence.
    We also observe monotonic decrease in validation performance with increasing $k$ for UPG. $^*$ indicates that the model runs out of memory at finest resolution.
    }
    \label{tab:implicitgradient}
\end{table}
\cref{tab:implicitgradient} shows that Jacobian-free Backprop provides the best performance, followed by unrolled phantom gradients with $k=3$.
For the UPG variants, we run out of memory with $k=10$ at the finest resolution due to the computational demands of explicit unrolling.
The results for UPG also show that explicit unrolling is both computationally demanding and unstable compared to cheaper variants like JFB.
For the full Hessian IFT, we observe poor training performance due to the ill-conditioning of the Hessian $\nabla_{\varphi}^2 C(\varphi, F_f, F_m) = \frac{\partial \varrho}{\partial \varphi}$.
Since this ill conditioned Hessian's inverse is multiplied with the incoming warp gradient $\frac{\partial T}{\partial \varphi}$, the feature gradient $\frac{\partial T}{\partial F}$ is sparse and noisy.
This severe ill-conditioning of the inverse Hessian is also observed in ~\cite{jena2024fireants}.
}

\edit{
We also examine the eigenspectra of the inverse Hessian matrix and its effect on the feature gradient computation in \cref{fig:qual-ift}.
We observe in \cref{fig:qual-ift} that although both full Hessian IFT and JFB receive the same gradients $\frac{\partial T}{\partial \varphi}$, they both provide significantly different feature gradients. 
The top eigenvalues of the Hessian matrix skew the gradient due to their large magnitude compared to the rest of the eigenspectra.
This leads to sparse gradients with respect to the feature images (visualized as bright and dark speckles), consequently leading to poor training performance.
This ablation provides motivation for future work to precondition the Hessian while addressing its ill-conditioned nature.
}

\edit{\section{Conclusion and Future Work}}

\edit{ 
DLIR methods provide several benefits such as amortized optimization, ability to leverage weak supervision and learn from large (labeled) datasets.
However, coupling of the feature learning and optimization steps in DLIR methods limits the flexibility and robustness of the deep networks.
Existing attempts to synergize optimization and feature learning in deep networks have been limited due to two reasons.
First, storing the entire computational graph of iterative optimization of 3D images will require an excessive (and often infeasible) memory footprint.
Second, existing methods have limited mathematical formulation to enable the ability to backpropagate features from a generic iterative optimization-based solver to learnable, \textit{task-aware} features of images.
We highlight the shortcoming of the existing classes of methods that aim to mitigate this issue, and propose a novel paradigm that incorporates optimization-as-a-layer for learning-based frameworks.
This paradigm allows the use of advanced black-box optimization toolkits in the forward pass, and a mathematically sound formulation to backpropagate features from the optimization solver to the feature learning network, without any additional compute or memory overhead.  \\
}

\edit{
Our comprehensive experimental setup on multiple datasets measuring both in-distribution and out-of-distribution performance demonstrates two solid empirical conclusions. 
First, task-aware learning is required for task-aware performance.
Using generic feature extractors or emulating iterative solvers only for a few steps cannot achieve asymptotically optimal in-distribution performance.
Second, iterative optimization is \textit{necessary} for robustness to out-of-distribution data.
Regardless of the performance of parameteric deep learning methods on in-distribution data, most methods fail to generalize on out-of-distribution data.
Multi-scale iterative optimization is therefore necessary for robustness to unseen image characteristics typically encountered in real-world clinical scenarios.
Since {\method} combines task-specific feature learning and black-box iterative optimization end-to-end, our method achieves state-of-the-art performance on the in-distribution setting, and is robust to out-of-distribution data. 
Densification of features from our method also leads to better optimization landscapes, and our method is robust to unseen anisotropy and domain shift.
To our knowledge, our method is the first to switch between {transformation representations} (free-form to diffeomorphic) at \textit{test time} without any retraining.
This comes with fast inference runtimes compared to baselines that utilize recurrent architectures for explicit unrolling, and interpretability of the features used for optimization. We aim to stabilize the training dynamics of the Hessian-based IFT solver, and explore multimodal registration for future work.
}

\section*{Acknowledgements}

This work was supported by the National Institutes of Health (NIH) under grants RF1-MH124605, R01-HL133889, R01-EB031722, U24-NS135568,  National Science Foundation (IIS-2145164, CCF-2212519), the Office of Naval Research (N00014-22-1-2255).

\bibliographystyle{elsarticle-harv}
\bibliography{ref}
\clearpage

\appendix
\section{Appendix}

\subsection{Implicit bias of optimization for registration}
\label{app:implicitbias}
Model based systems, such as deep networks are not immune to inductive biases due to architecture, loss functions, and optimization algorithms used to train them. 
Functional forms of the deep network induce constraints on the solution space, but optimization algorithms are not excluded from such biases either.
The implicit bias for Gradient Descent is a well-studied phenomena for overparameterized linear and shallow networks. 
Gradient Descent for linear systems leads to an optimum that is in the span of the input data starting from the initialization~\cite{zhang2021understanding,soudry2018implicit,ji2018gradient,pesme2021implicit,wu2020direction}.
This bias is also dependent on the chosen representation, since that defines the functional relationship of the gradients with the parameters and inputs.
This limits the reachable set of solutions by the optimization algorithm when multiple local minima exist.

In the case of image registration, the optimization limits the space of solutions (warps) that can be obtained by the SGD algorithm.
To show this, we consider the transformation $\varphi$ as a set of particles in a Langrangian frame that are displaced by the optimization algorithm to align the moving image to the fixed image.
Consider a regular grid of particles, whose locations specify the warp field.
Let the location of $i$-th particle at iteration $t$ be $\varphi^{(t)}(\mathbf{x}_i)$.
For a fixed feature image $F_f$, moving image $F_m$ and current iterate $\varphi^{(t)}$, the gradient of the registration loss with respect to particle $i$ at iteration $t$ is given by
\begin{equation}
    \frac{\partial C(F_f, F_m\circ \varphi^{(t)})}{\partial \varphi^{(t)}(\mathbf{x}_i)} = C'_{i}(F_f, F_m\circ \varphi^{(t)}) \nabla F_m(\varphi^{(t)}(\mathbf{x}_i))
\end{equation}
where $$C'_i(F_f, F_m\circ \varphi^{(t)}) = \frac{\partial C(F_f, F_m\circ \varphi^{(t)})}{\partial M(\varphi^{(t)}(\mathbf{x}_i))}$$ is the (scalar) derivative of scalar loss $C$ with respect to the intensity of $i$-th particle computed at the current iterate, and $\nabla F_m(\varphi^{(t)}(\mathbf{x}_i))$ is the spatial gradient of the moving image at the location of the particle.
Note that the \textbf{direction} of the gradient of particle $i$ is \textit{independent} of the fixed image, loss function, and location of other particles -- it only depends on the spatial gradient of the moving image at the location of the particle. %
This restricts the movement of a particle located at any given location along a 1D line whose direction is the spatial gradient of the moving image at that location. %
Since $F_f$ and $F_m$ are computed independently of each other  (and therefore no information of $F_f$ and $F_m$ is contained in each other), the space of solutions of $\varphi$ is restricted by this implicit bias.
This is restrictive because the similarity function and fixed image do not influence the direction of the gradient, and the optimization algorithm is biased towards solutions that are in the direction of the gradient of the moving image.

\begin{figure*}[h!]
    \includegraphics[width=\linewidth]{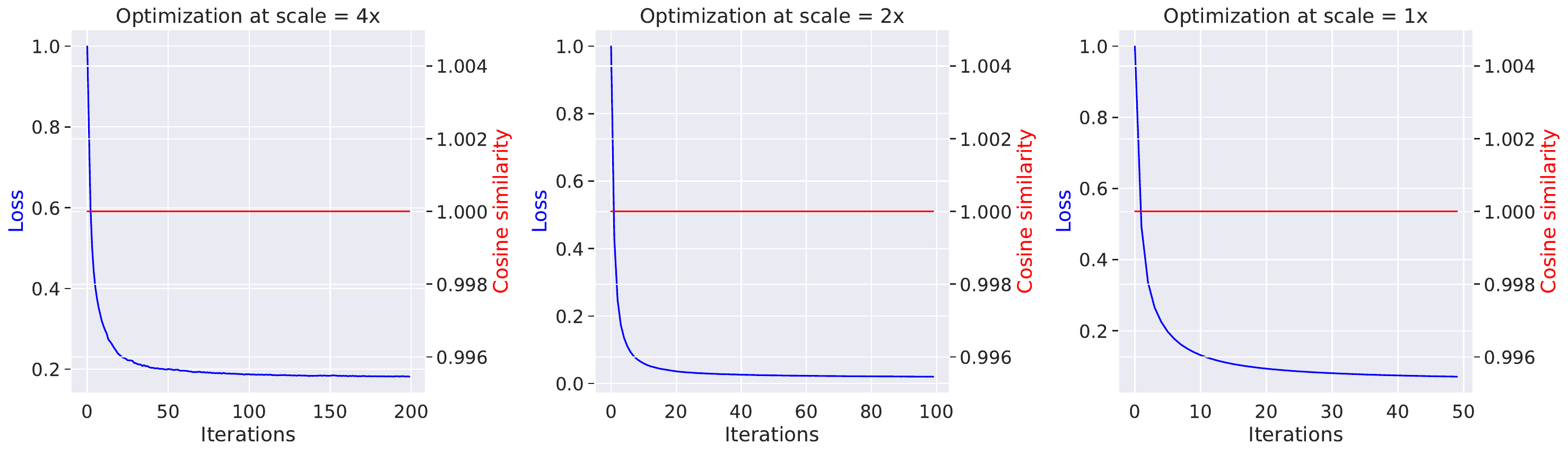}
    \caption{\textbf{Implicit bias in SGD for image registration.} The plot shows the loss curves for a multi-scale optimization of two feature images. Each plot also shows the absolute cosine similarity of per-pixel gradients obtained by $C$ and $C_{\text{surrogate}}$ at each iteration. Note that over the course of optimization, the cosine similarity is always 1 -- demonstrating the implicit bias of the optimization for registration.}
    \label{fig:implicitbiasplot}
\end{figure*}

We show this bias empirically -- we perform multi-scale optimization algorithm using feature maps obtained from the network.
We keep track of two gradients, one obtained by the loss function, and another obtained by the gradient of a surrogate loss $C_{\text{surrogate}}(F_m, \varphi^{(t)}) = \sum_i F_m(\varphi^{(t)}(\mathbf{x}_i))$.
Note that $C_{\text{surrogate}}$ does not depend on the fixed image or the loss function.
The gradient of $C_{\text{surrogate}}$ with respect to the $i$-th particle is given by $\nabla F_m(\varphi^{(t)}(\mathbf{x}_i))$.
At each iteration, we compute the magnitude of cosine similarly between the gradients of $C$ and $C_{\text{surrogate}}$.
\cref{fig:implicitbiasplot} shows that the loss converges, and the  per-pixel gradients can be predicted by $C_{\text{surrogate}}$ alone, as depicted by the magnitude and standard deviation of cosine similarity between $C$ and $C_{\text{surrogate}}$. 
This limits the movement of each particle along a 1D line in an $N$-D space, and limits the degrees of freedom of the optimization by $N$-fold for $N$-D images.
Future work will aim at alleviating this implicit bias to allow for more flexible solutions.

\subsection{Algorithm details}
\begin{figure*}
    \centering
    \includegraphics[width=\linewidth]{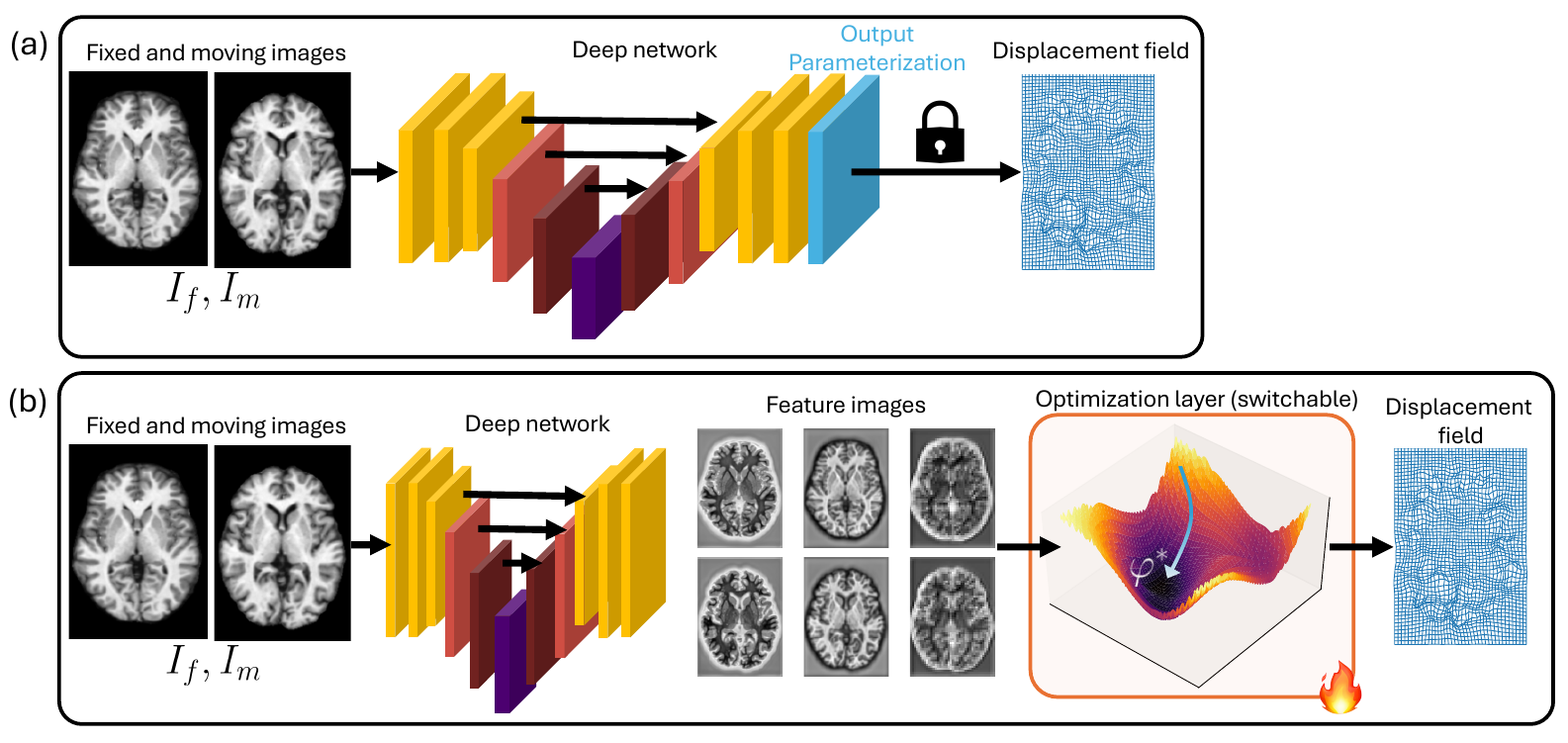}
    \caption{\textbf{Comparison of typical DLIR method and our method.} \textbf{(a)} shows the pipeline of a typical deep network. The neural network architecture takes the channelwise concatenation of the fixed and moving images as input, and outputs a warp field, which has a \textit{fixed} transformation representation (SVF, free-form, B-splines, affine, etc. denoted as the blue locked layer). This representation is fixed throughout training and cannot be switched at test-time, without additional finetuning of the network. \textbf{(b)} shows our framework wherein the fixed and moving images are input \textit{separately} into a feature extraction network that outputs multi-scale features.
    These features are then passed onto an {iterative black-box} solver than can be \textit{implicitly differentiated} to backpropagate the gradients from the optimized warp field back to the feature network. This allows for a more flexible transformation representation, and the optimization solver can be switched at test-time with zero finetuning.}
    \label{fig:dlir-vs-ours}
\end{figure*}

{\method} is a learnable framework that leverages \textit{implicit differentiation} of an arbitrary black-box optimization solver to learn features such that registration in this feature space corresponds to good registration of the images and additional label maps.
This additional indirection leads to learnable features that are registration-aware, interpretable, and the framework inherits the optimization solver's versatility to variability in the data like difference in contrast, anisotropy, and difference in sizes of the fixed and moving images.
We contrast our approach with a typical classical optimization-based registration algorithm in \cref{alg:pseudocode}. 
A classical multi-scale optimization routine \textit{indiscriminately} downsamples the intensity images, and does not retain discriminative information that is useful for registration.
Since our method is trained to maximize label alignment from all scales, multi-scale features obtained from our method are more discriminative and registration-aware.
We also compare DIO with a typical DLIR method in \cref{fig:dlir-vs-ours}.
Note that the fixed end-to-end architecture and functional form of a deep network subsumes the representation choice into the architecture as well, limiting its ability to switch to arbitrary transformation representations at inference time without additional retraining.
Our framework therefore combines the benefits of both classical (robustness to out-of-distribution datasets, and zero-shot transfer to other optimization routines) and learning-based methods (high-fidelity, label-aware, and  registration-aware).

\begin{figure*}[h!]
\begin{minipage}{0.99\linewidth}
    \begin{algorithm}[H]
        \caption{Classical registration pipeline}
        \label{alg:classical-pcode}
        \begin{algorithmic}[1]
            \State \textbf{Input:} Fixed image $I_f$, Moving image $I_m$ \\ Scales $[s_1, s_2, \ldots, s_n]$, Iterations $[T_1, T_2, \ldots T_n]$, $n$ levels.
            \State Initialize $\varphi = \mathbf{Id}_{s_1}$.  \Comment{Initialize warp to identity at first scale}
            \State Initialize  ${l} = 1$.   \Comment{Initialize current scale}
            \While{${l} \le n$}
                \State Initialize $i = 0$
                \State Initialize $I_f^{l}, I_m^{l} = \text{downsample}(I_f, s_{l}),  \text{downsample}(I_m, s_{l})$
                \While{$i < T_{{l}}$}
                    \State $L_i = C(I_f^{l}, I_m^{l} \circ \varphi^i)$
                    \State Compute $\nabla_{\varphi} L$
                    \State Update $\varphi^{(i+1)} = \text{Optimize}(\varphi^i, \nabla_{\varphi} L_i)$  \Comment{Optimization algorithm}
                    \State $i = i + 1$
                \EndWhile
                \If{$l < n$}
                    \State $\varphi = \text{Upsample}(\varphi, s_{(l+1)})$ \Comment{Upsample warp to next level}
                \EndIf
                \State $l = l + 1$
            \EndWhile
        \end{algorithmic}
    \end{algorithm}
\end{minipage}
\begin{minipage}{0.99\linewidth}
    \begin{algorithm}[H]
        \caption{Differentiable Implicit Optimization for Registration (Our algorithm)}
        \label{alg:dio-pcode}
        \begin{algorithmic}[1]
            \State \textbf{Input:} \textcolor{red}{Fixed features $\mathcal{F}_f = [F_f^1, F_f^2 \ldots F_f^n]$, Moving features $\mathcal{F}_f = [F_f^1, F_f^2 \ldots F_f^n]$} \\ Scales $[s_1, s_2, \ldots, s_n]$, Iterations $[T_1, T_2, \ldots T_n]$, $n$ levels.
            \State Initialize $\varphi = \mathbf{Id}_{s_1}$.  \Comment{Initialize warp to identity at first scale}
            \State Initialize  ${l} = 1$.   \Comment{Initialize current scale}
            \State Outputs $= []$. \Comment{Save intermediate outputs for backpropagation}
            \While{${l} \le n$}
                \State Initialize $i = 0$
                \State Initialize $I_f^{l}, I_m^{l} = \textcolor{red}{F_f^l, F_m^l}$
                \While{$i < T_{{l}}$}
                    \State $L_i = C(I_f^{l}, I_m^{l} \circ \varphi^i)$
                    \State Compute $\nabla_{\varphi} L$
                    \State Update $\varphi^{(i+1)} = \text{Optimize}(\varphi^i, \nabla_{\varphi} L_i)$  \Comment{Optimization algorithm}
                    \State $i = i + 1$
                \EndWhile
                \State $\textcolor{red}{\text{Outputs.append}\left(\varphi^{(T_l)}\right)}$  \Comment{Save final warp at this level for backpropagation}
                \If{$l < n$}
                    \State $\varphi = \text{Upsample}(\varphi, s_{(l+1)})$ \Comment{Upsample warp for next level}
                \EndIf
                \State $l = l + 1$
            \EndWhile
        \end{algorithmic}
    \end{algorithm}
\end{minipage}
\caption{\textbf{Comparison of a typical classical registration algorithm and DIO:} \cref{alg:classical-pcode} shows a typical classical registration algorithm that uses a multi-scale optimization routine to register the fixed and moving images. 
At each level $l$, the fixed and moving images are downsampled by a factor of $s_l$, therefore trading off between discriminative information and vulnerability to local minima. 
\cref{alg:dio-pcode} shows our algorithm (red text highlights differences compared to \cref{alg:classical-pcode}) that uses a separate scale-space feature at each level.
Unlike classical methods, the scale-space feature can capture different discriminative features at each level to maximize label alignment and the multi-scale nature helps avoid local minima.
}
\label{alg:pseudocode}
\end{figure*}

\begin{figure*}[ht]
    \begin{minipage}{0.99\linewidth}
    \begin{algorithm}[H]
        \caption{Backward pass for DIO}
        \label{alg:dio-pcode-bwd}
        \begin{algorithmic}[1]
            \State \textbf{Input:} \textcolor{red}{Fixed features $\mathcal{F}_f = [F_f^1, F_f^2 \ldots F_f^n]$, Moving features $\mathcal{F}_f = [F_f^1, F_f^2 \ldots F_f^n]$}, Backend $backend$ \\
            Stored outputs $[\varphi^1, \varphi^2 \ldots \varphi^n]$, gradients $\left[\frac{\partial T}{\partial \varphi^1}, \frac{\partial T}{\partial \varphi^2}, \ldots \frac{\partial T}{\partial \varphi^n}\right]$ \\
            Backend $backend$,  $n$ levels. 
            \State Initialize ${l} = 1$.  \Comment{Initialize current scale}
            \While{${l} \le n$}
                \If{$backend == $Hessian}
                    \State Compute $H = \frac{\partial \varrho}{\partial \varphi^l}$
                    \State Update $v = \texttt{linalg.lstsq}\left(H, \frac{\partial T}{\partial \varphi^l} \right)$   \Comment{Full Hessian IFT}
                \Else       
                    \State Update $v = \frac{\partial T}{\partial \varphi^l}$  \Comment{Jacobian-Free Backprop}
                \EndIf
                \State Compute $h = v^T \cdot \varrho(\varphi^l, F_f^l, F_m^l)$
                \State Set $\texttt{grad}(F_f^l) = \texttt{autograd.grad}(h, F_f^l)$
                \State Set $\texttt{grad}(F_m^l) = \texttt{autograd.grad}(h, F_m^l)$
            \EndWhile
        \end{algorithmic}
    \end{algorithm}
    \end{minipage}
\caption{\textbf{Pseudocode for backward pass with DIO:} Given the stored features and outputs from the forward pass, and the gradients w.r.t. final warp from the backward pass, we compute the gradients of the loss function with respect to the fixed and moving features at each level. The gradients are analytically computed depending on the specified backend.}
\label{alg:pseudocode-bwd}
\end{figure*}

\subsection{Toy example}
\label{app:toyexample}

\begin{figure*}[h!]
   \centering 
   \includegraphics[width=\linewidth]{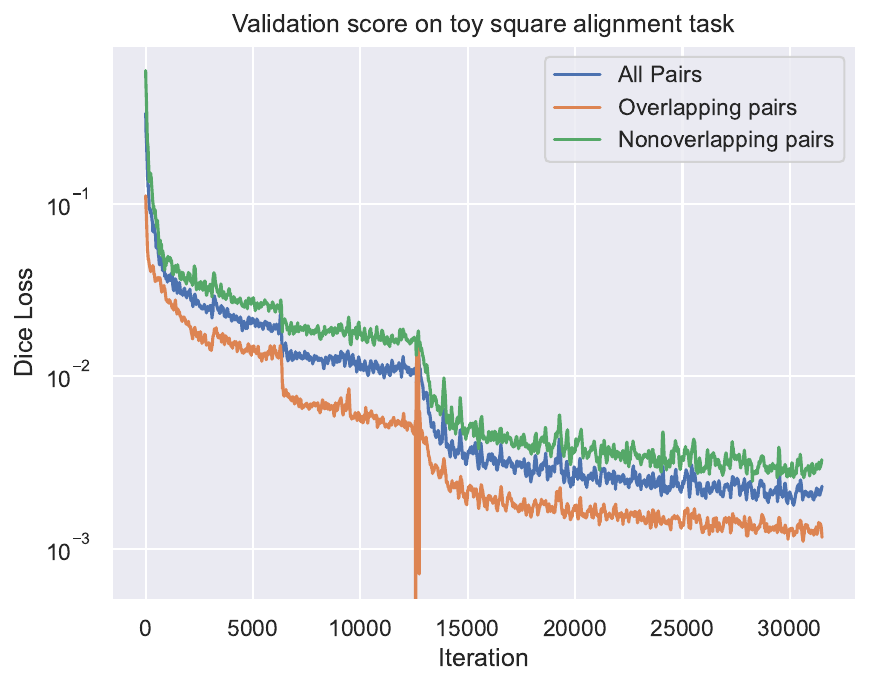}
   \caption{\textbf{Loss curves for toy dataset}. Plot shows three curves - the Dice score for (a) all validation image pairs, (b) image pairs that have non-zero overlap in the image space (therefore a gradient-based affine solver will recover a transform from intensity images), and (c) image pairs that have zero overlap in the image space (therefore any gradient-based solver using intensity images will fail). Our feature network recovers dense multi-scale features (see \cref{fig:toyexample}) which allows all subsets to be registered with >0.99 Dice score.}
   \label{fig:toyexampleloss}
\end{figure*}

\cref{fig:toyexampleloss} shows the loss curves for the toy dataset described in \cref{sec:toyexample}.
An image-based optimization algorithm would correspond to the green curve being a flat line at $1$ due to the flat landscape of the intensity-based loss function. 

\subsection{Quantitative Results}

\cref{tab:robustness} shows the quantitative results of our method for out-of-distribution performance on the IBSR18, CUMC12, and LPBA40 datasets.
In 9 out of 10 cases, DIO demonstrates the best accuracy with fairly lower standard deviations, highlighting the robustness of the model.
DIO therefore serves as a strong candidate for out-of-distribution performance, and can be used in a variety of settings where the training and test distributions differ.

\begin{table*}[htpb!]
\centering
\resizebox{\linewidth}{!}{%
\begin{tabular}{lccccc}
\hline
\textbf{Method} & \textbf{Dice} & \multicolumn{2}{c}{\textbf{Isotropic}} & \multicolumn{2}{c}{\textbf{Anisotropic}} \\
& \textbf{supervision}& \textbf{Crop} & \textbf{No Crop} & \textbf{Crop} & \textbf{No Crop}  \\
\hline
Conditional LapIRN & \xmark & 0.7367 $\pm$ 0.0237 & \ding{55} & 0.7269 $\pm$ 0.0328 & \cellcolor{better_color}0.7317 $\pm$ 0.0303 \\
LapIRN & \xmark & 0.5257 $\pm$ 0.1316 & \ding{55} & 0.5435 $\pm$ 0.1266 & 0.5001 $\pm$ 0.1271 \\
LapIRN  & \cmark & 0.6259 $\pm$ 0.1238 & \ding{55} & 0.6209 $\pm$ 0.1163 & 0.5759 $\pm$ 0.1207 \\
LKU-Net & \xmark & 0.6309 $\pm$ 0.0839 & \ding{55} & 0.6276 $\pm$ 0.0838 & 0.6072 $\pm$ 0.0787 \\
LKU-Net  & \cmark & 0.6267 $\pm$ 0.0776 & \ding{55} & 0.6231 $\pm$ 0.0730 & 0.5992 $\pm$ 0.0757 \\
SymNet & \xmark & 0.7213 $\pm$ 0.0273 & \ding{55} & 0.7116 $\pm$ 0.0398 & \cellcolor{good_color}0.7117 $\pm$ 0.0398 \\
SymNet  & \cmark & 0.6731 $\pm$ 0.0688 & \ding{55} & 0.6672 $\pm$ 0.0731 & 0.6674 $\pm$ 0.0728 \\
TransMorph Large  & \cmark & \cellcolor{good_color}0.7383 $\pm$ 0.0353 & \ding{55} & \cellcolor{good_color}0.7312 $\pm$ 0.0405 & \ding{55} \\
TransMorph Regular & \xmark & 0.7221 $\pm$ 0.0400 & \ding{55} & 0.7289 $\pm$ 0.0417 & \ding{55} \\
TransMorph Regular  & \cmark & 0.7293 $\pm$ 0.0370 & \ding{55} & 0.7113 $\pm$ 0.0520 & \ding{55} \\
VoxelMorph & \xmark & 0.5118 $\pm$ 0.1774 & \ding{55} & 0.5233 $\pm$ 0.1693 & \ding{55} \\
SynthMorph & \cmark & \cellcolor{better_color}0.7423 $\pm$ 0.0225 & \ding{55} & \cellcolor{better_color}0.7476 $\pm$ 0.0238 & \ding{55} \\
Ours (LKU) & \cmark & \cellcolor{best_color}0.7698 $\pm$ 0.0193 & \cellcolor{best_color}0.7587 $\pm$ 0.0208 & \cellcolor{best_color}0.7728 $\pm$ 0.0219 & \cellcolor{best_color}0.7572 $\pm$ 0.0369 \\
\hline
\hline
Conditional LapIRN & \xmark & 0.4793 $\pm$ 0.0373 & \cellcolor{better_color} 0.4804 $\pm$ 0.0368 & \cellcolor{good_color}0.4880 $\pm$ 0.0416 & \cellcolor{better_color}0.4827 $\pm$ 0.0408 \\
LapIRN & \xmark & 0.3719 $\pm$ 0.0897 & 0.3491 $\pm$ 0.0895 & 0.3524 $\pm$ 0.1001 & 0.3556 $\pm$ 0.0989 \\
LapIRN  & \cmark & 0.4121 $\pm$ 0.0907 & 0.3838 $\pm$ 0.0929 & 0.3911 $\pm$ 0.1060 & 0.3896 $\pm$ 0.1063 \\
LKU-Net & \xmark & 0.4054 $\pm$ 0.0641 & 0.3922 $\pm$ 0.0679 & 0.4086 $\pm$ 0.0732 & 0.3999 $\pm$ 0.0697 \\
LKU-Net  & \cmark & 0.3904 $\pm$ 0.0547 & 0.3827 $\pm$ 0.0574 & 0.3967 $\pm$ 0.0745 & 0.3960 $\pm$ 0.0678 \\
SymNet & \xmark & 0.4761 $\pm$ 0.0524 & \cellcolor{good_color}0.4761 $\pm$ 0.0524 & 0.4822 $\pm$ 0.0565 & \cellcolor{good_color}0.4820 $\pm$ 0.0565 \\
SymNet  & \cmark & 0.4457 $\pm$ 0.0675 & 0.4457 $\pm$ 0.0675 & 0.4518 $\pm$ 0.0787 & 0.4521 $\pm$ 0.0786 \\
TransMorph Large  & \cmark & \cellcolor{good_color}0.4827 $\pm$ 0.0531 & \ding{55} & 0.4858 $\pm$ 0.0587 & \ding{55} \\
TransMorph Regular & \xmark & \cellcolor{better_color}0.4929 $\pm$ 0.0502 & \ding{55} & \cellcolor{better_color}0.4967 $\pm$ 0.0540 & \ding{55} \\
TransMorph Regular  & \cmark & 0.4737 $\pm$ 0.0549 & \ding{55} & 0.4741 $\pm$ 0.0628 & \ding{55} \\
VoxelMorph & \xmark & 0.3519 $\pm$ 0.1271 & \ding{55} & 0.3469 $\pm$ 0.1308 & \ding{55} \\
SynthMorph & \cmark & 0.4761 $\pm$ 0.0397 & \ding{55} & 0.4797 $\pm$ 0.0426 & \ding{55} \\
Ours (LKU) & \cmark & \cellcolor{best_color}0.5137 $\pm$ 0.0410 & \cellcolor{best_color}0.5126 $\pm$ 0.0412 & \cellcolor{best_color}0.5237 $\pm$ 0.0433 & \cellcolor{best_color}0.5162 $\pm$ 0.0448 \\
\hline \hline
Conditional LapIRN & \xmark & \cellcolor{good_color}0.7113 $\pm$ 0.0178 & \cellcolor{better_color} 0.7109 $\pm$ 0.0178 & - & - \\
LapIRN & \xmark & 0.6026 $\pm$ 0.0317 & 0.5878 $\pm$ 0.0325 & - & - \\
LapIRN  & \cmark & 0.6395 $\pm$ 0.0269 & 0.6211 $\pm$ 0.0294 &  - & - \\
LKU-Net & \xmark & 0.6746 $\pm$ 0.0230 & 0.6708 $\pm$ 0.0249 &  - & - \\
LKU-Net  & \cmark & 0.6266 $\pm$ 0.0299 & 0.6220 $\pm$ 0.0296 &  - & - \\
SymNet & \xmark & 0.6797 $\pm$ 0.0239 & \cellcolor{good_color}0.6797 $\pm$ 0.0238 &  - & - \\
SymNet  & \cmark & 0.6700 $\pm$ 0.0248 & 0.6698 $\pm$ 0.0248 &  - & - \\
TransMorph Large  & \cmark & 0.6918 $\pm$ 0.0219 & \ding{55} &  - & - \\
TransMorph Regular & \xmark & 0.6919 $\pm$ 0.0191 & \ding{55} &  - & - \\
TransMorph Regular  & \cmark & 0.6855 $\pm$ 0.0225 & \ding{55} &  - & - \\
VoxelMorph & \xmark & 0.6776 $\pm$ 0.0365 & \ding{55} &  - & - \\
SynthMorph & \cmark & \cellcolor{best_color}0.7189 $\pm$ 0.0172 & \ding{55} &  - & - \\
Ours (LKU) & \cmark & \cellcolor{better_color}0.7139 $\pm$ 0.0181 &\cellcolor{best_color}0.7131 $\pm$ 0.0181 &  - & - \\
\hline
\end{tabular}
}
\caption{
\textbf{Quantitative evaluation on out-of-distribution performance on IBSR18, CUMC12, and LPBA40 datasets.} We compare {\method} with other state-of-the-art DLIR methods. The `\textbf{Dice supervision}' column shows if the method is trained with label matching on the OASIS dataset.
We evaluate the performance of the methods with and without isotropic and anisotropic data resampling. 
The results are reported as mean $\pm$ standard deviation.
\colorbox{best_color}{\hspace{0.15cm} \vphantom{X}} = First, \colorbox{better_color}{\hspace{0.15cm} \vphantom{X}} = Second, \colorbox{good_color}{\hspace{0.15cm} \vphantom{X}} = Third best result.
}
\label{tab:robustness}
\end{table*}

\begin{figure*}[htpb!]
    \setlength{\imagewidth}{0.15\linewidth}  %
    \centering

    \rotatebox{90}{\adjustbox{valign=m}{\hspace{1em}KeyMorph}}
    \includegraphics[width=\imagewidth]{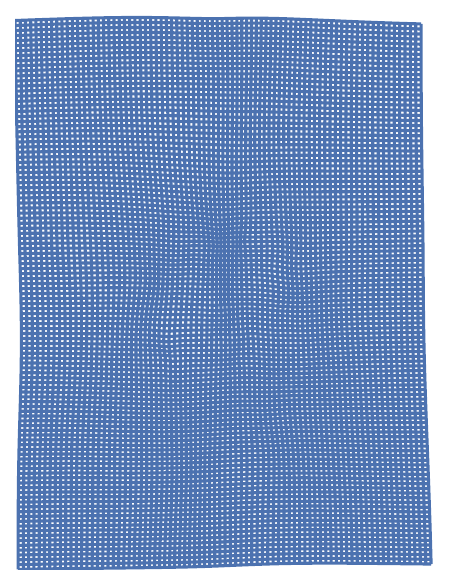}
    \includegraphics[width=\imagewidth]{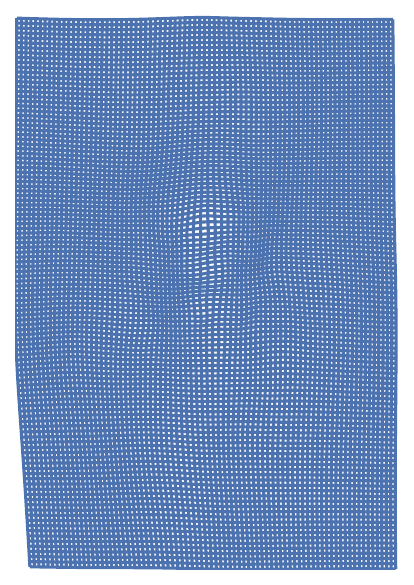}
    \includegraphics[width=\imagewidth]{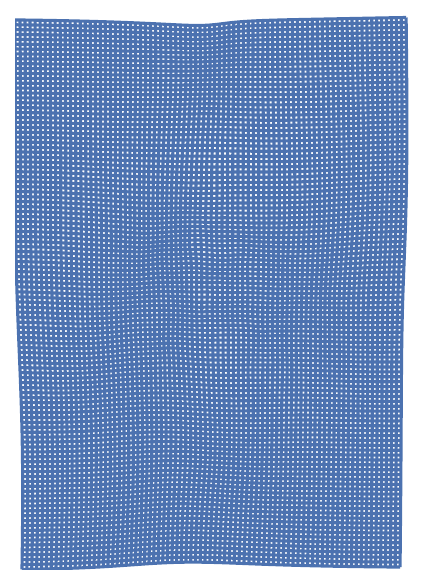}
    \includegraphics[width=\imagewidth]{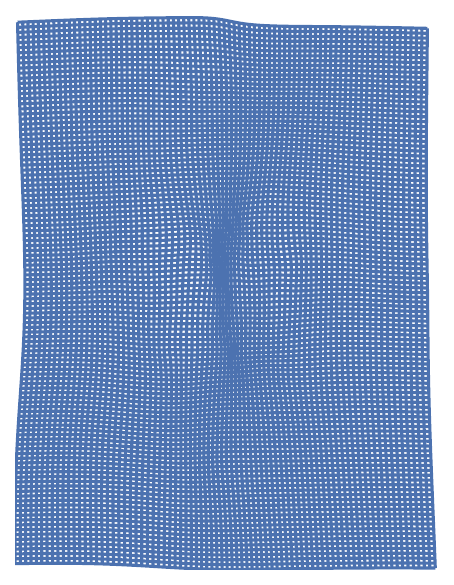}
    \includegraphics[width=\imagewidth]{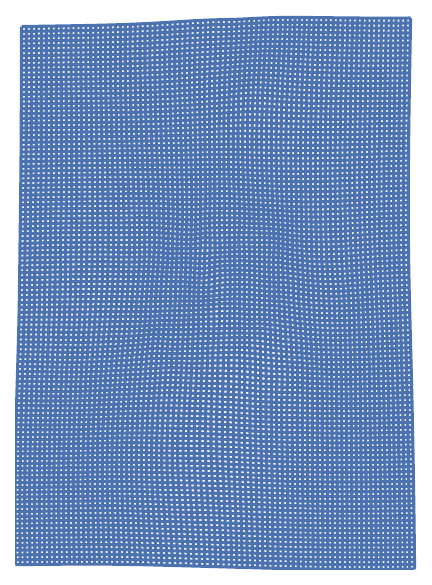}
    \includegraphics[width=\imagewidth]{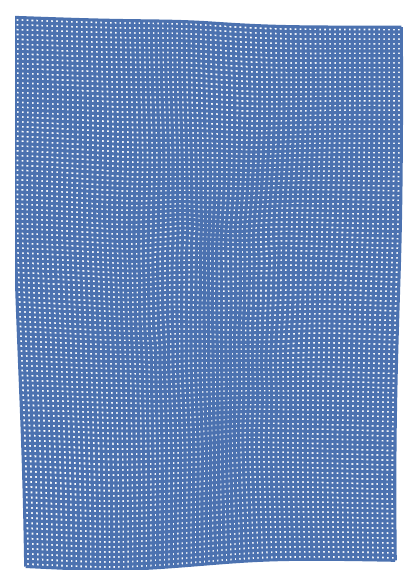}
    
    \rotatebox{90}{\adjustbox{valign=m}{\hspace{1em}KeyMorph}}
    \includegraphics[width=\imagewidth]{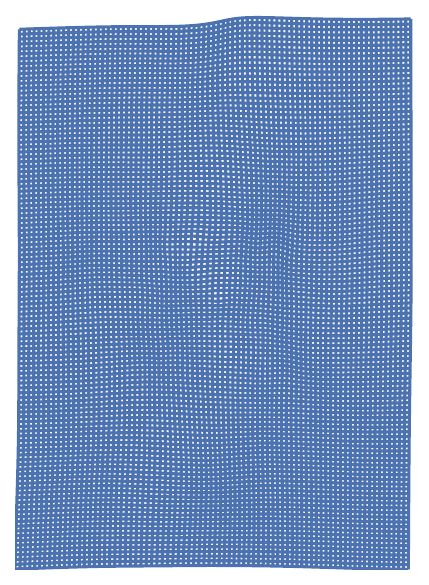}
    \includegraphics[width=\imagewidth]{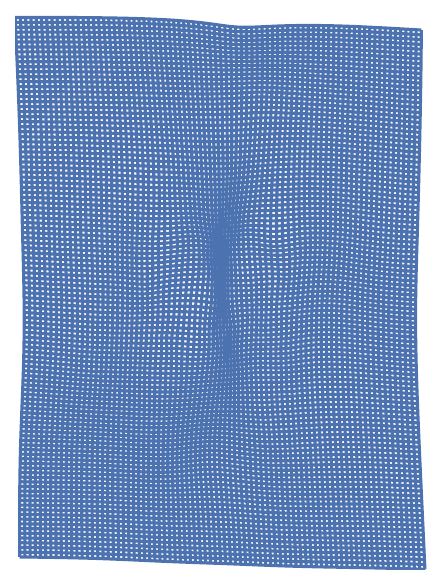}
    \includegraphics[width=\imagewidth]{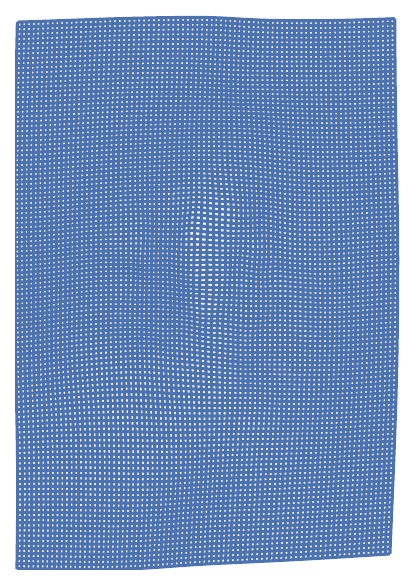}
    \includegraphics[width=\imagewidth]{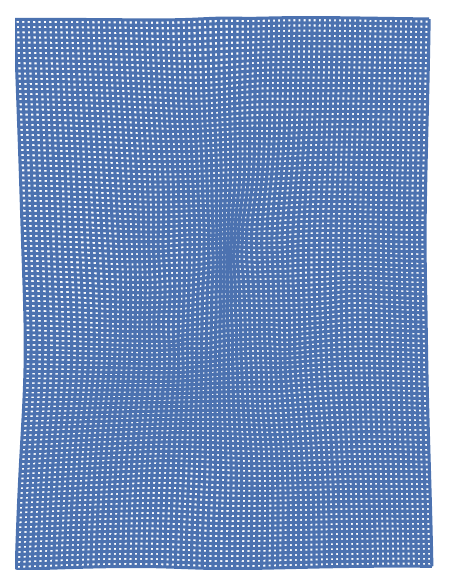}
    \includegraphics[width=\imagewidth]{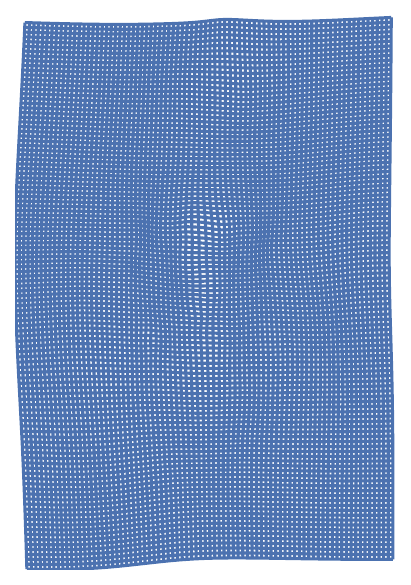}
    \includegraphics[width=\imagewidth]{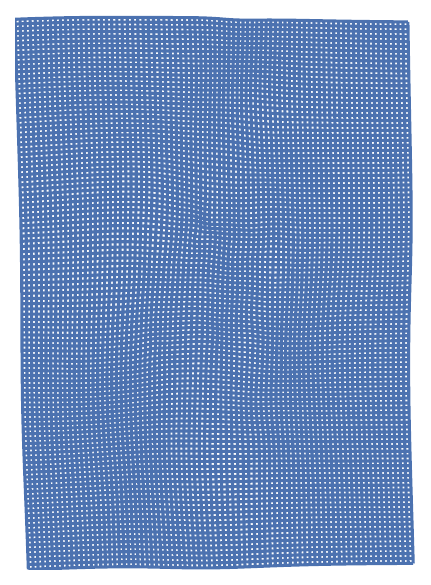}

    \hrule

    \rotatebox{90}{\adjustbox{valign=m}{\hspace{2.2em}Ours}}
    \includegraphics[width=\imagewidth]{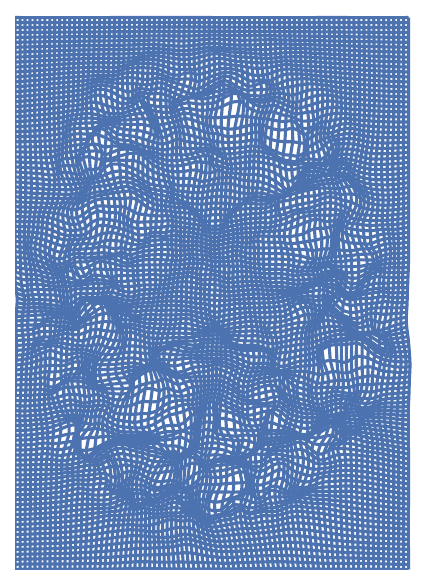}
    \includegraphics[width=\imagewidth]{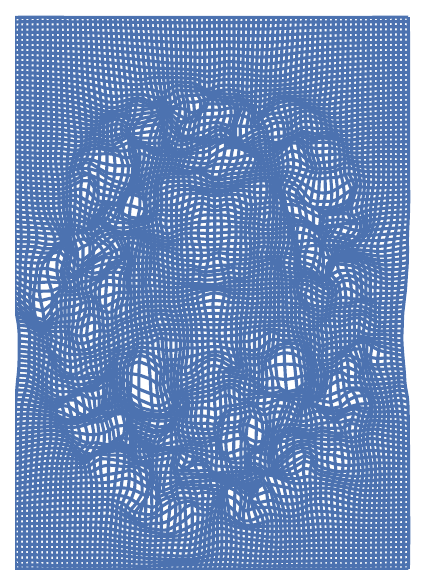}
    \includegraphics[width=\imagewidth]{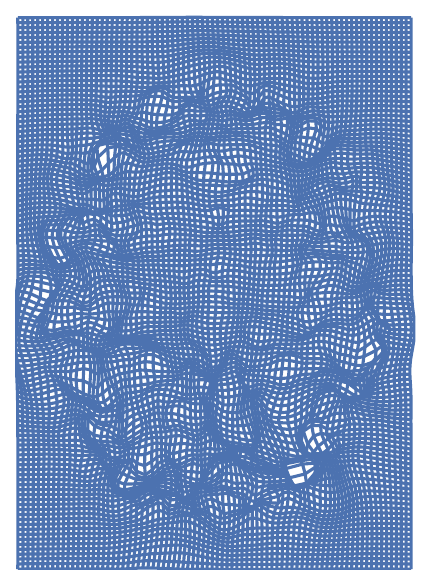}
    \includegraphics[width=\imagewidth]{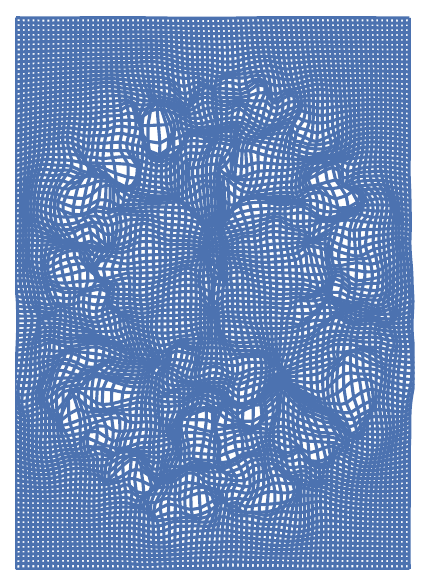}
    \includegraphics[width=\imagewidth]{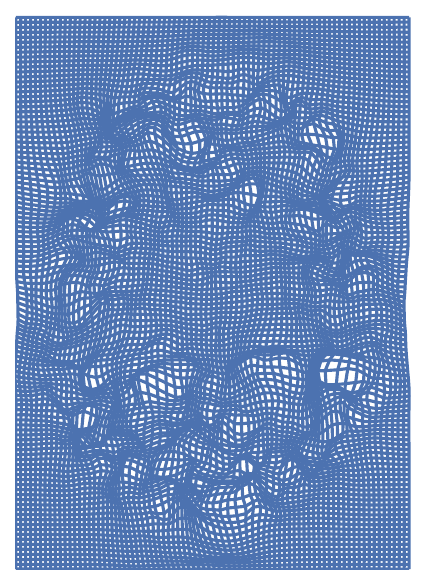}
    \includegraphics[width=\imagewidth]{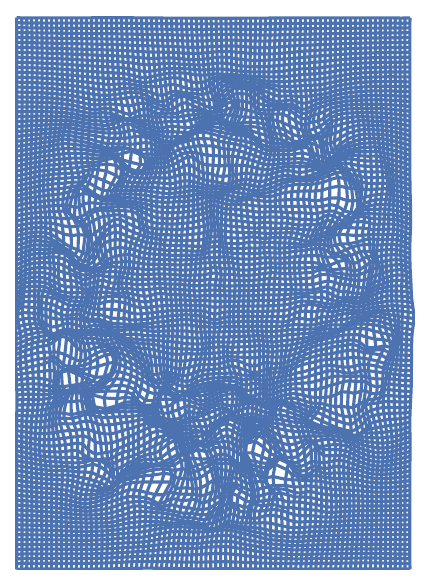}

    \rotatebox{90}{\adjustbox{valign=m}{\hspace{2.2em}Ours}}
    \includegraphics[width=\imagewidth]{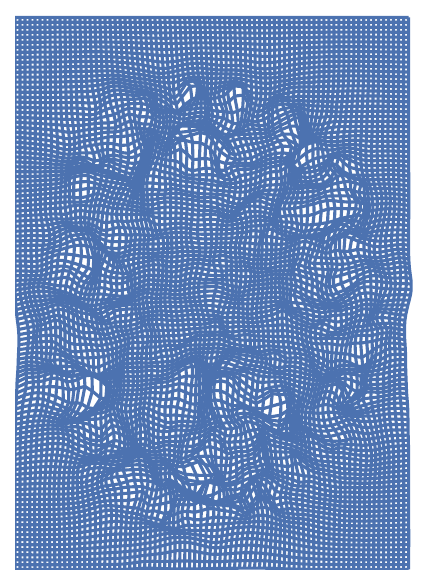}
    \includegraphics[width=\imagewidth]{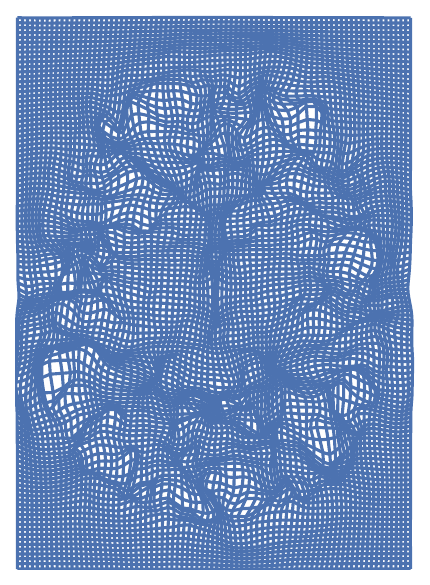}
    \includegraphics[width=\imagewidth]{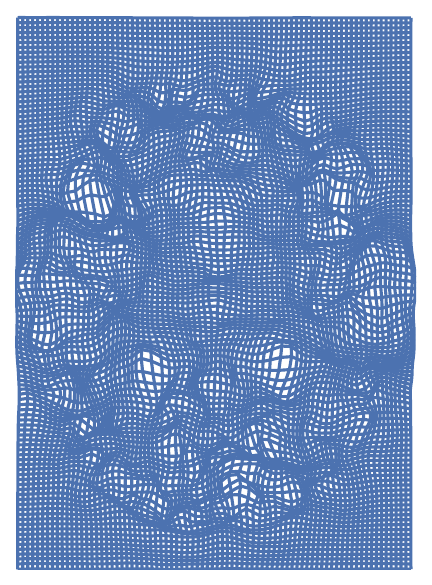}
    \includegraphics[width=\imagewidth]{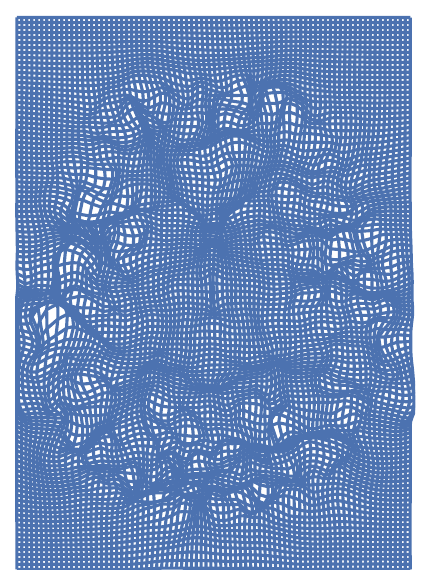}
    \includegraphics[width=\imagewidth]{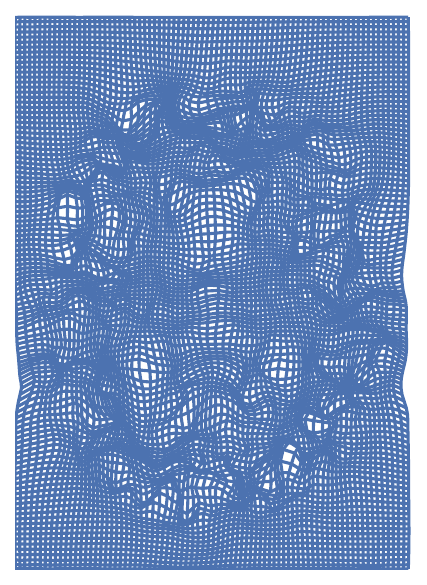}
    \includegraphics[width=\imagewidth]{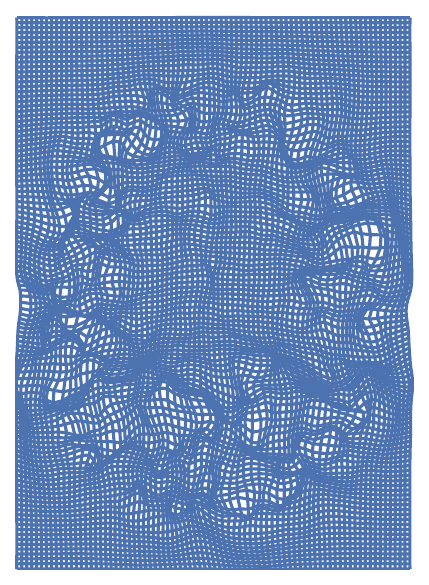}

    \caption{\textbf{Qualitative comparison of warp fields}. 
    Top two rows show the warp fields produced by thin plate spline using keypoints predicted by KeyMorph, bottom two rows show the warp fields produced by a diffeomorphic optimization routine from dense feature maps predicted by our method. 
    Compared to the thin plate spline representation, our method is able to produce complex deformation fields to accurately capture subtle anatomical differences in inter-subject MRI registration.
    }
    \label{fig:app-qual-warp}
\end{figure*}

\begin{figure*}[htpb!]
    \setlength{\imagewidth}{0.115\linewidth}  %
    \centering

    \rotatebox{90}{\adjustbox{valign=m}{\hspace{1em}KeyMorph}}
    \includegraphics[width=\imagewidth]{figures/oasis_results/oasis_00/moved_img_km.pdf}
    \includegraphics[width=\imagewidth]{figures/oasis_results/oasis_01/moved_img_km.pdf}
    \includegraphics[width=\imagewidth]{figures/oasis_results/oasis_02/moved_img_km.pdf}
    \includegraphics[width=\imagewidth]{figures/oasis_results/oasis_03/moved_img_km.pdf}
    \includegraphics[width=\imagewidth]{figures/oasis_results/oasis_04/moved_img_km.pdf}
    \includegraphics[width=\imagewidth]{figures/oasis_results/oasis_05/moved_img_km.pdf}
    \includegraphics[width=\imagewidth]{figures/oasis_results/oasis_06/moved_img_km.pdf}
    \includegraphics[width=\imagewidth]{figures/oasis_results/oasis_07/moved_img_km.pdf}

    \rotatebox{90}{\adjustbox{valign=m}{\hspace{1em}KeyMorph}}
    \includegraphics[width=\imagewidth]{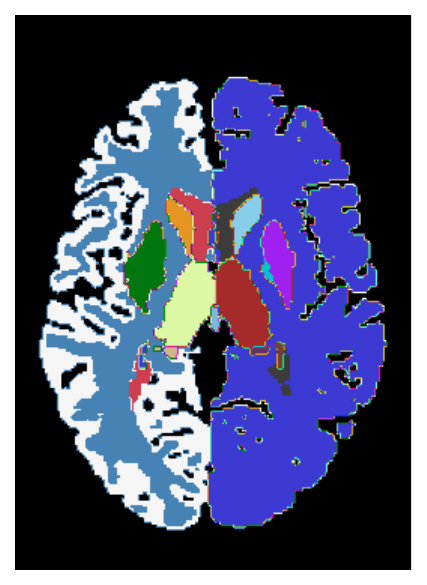}
    \includegraphics[width=\imagewidth]{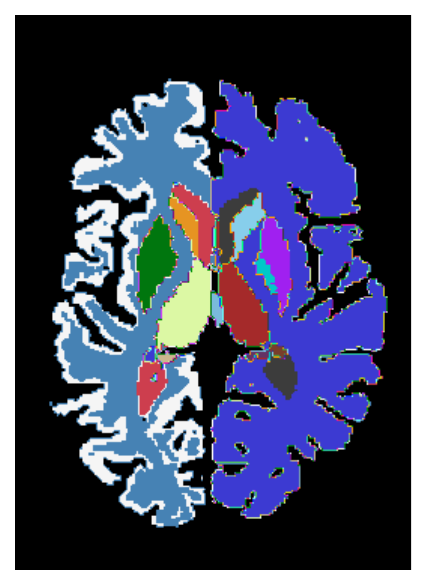}
    \includegraphics[width=\imagewidth]{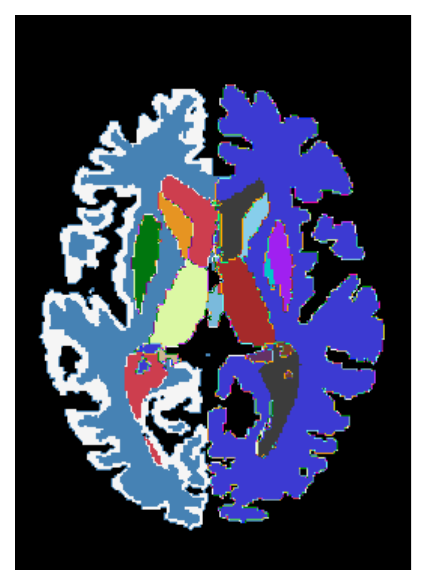}
    \includegraphics[width=\imagewidth]{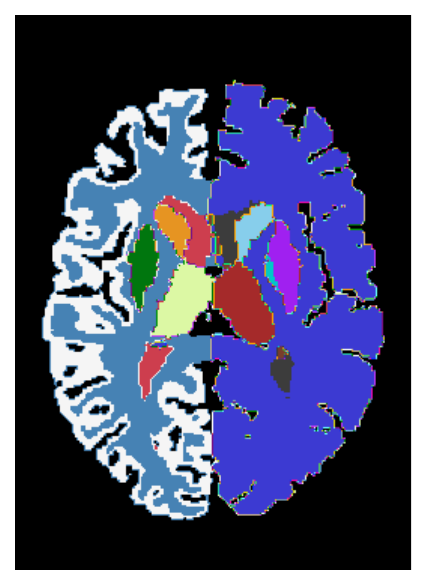}
    \includegraphics[width=\imagewidth]{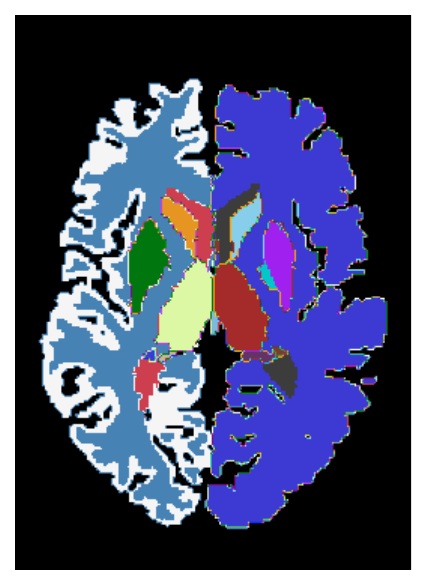}
    \includegraphics[width=\imagewidth]{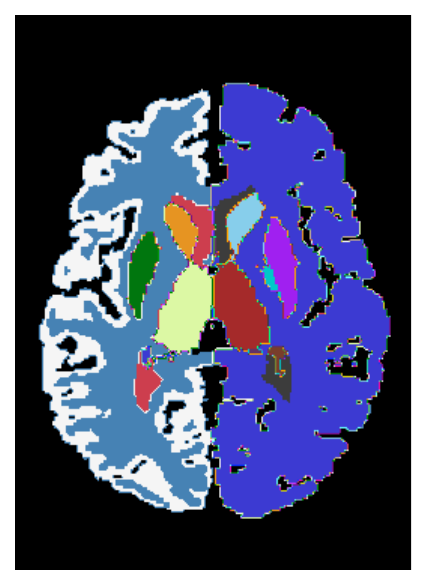}
    \includegraphics[width=\imagewidth]{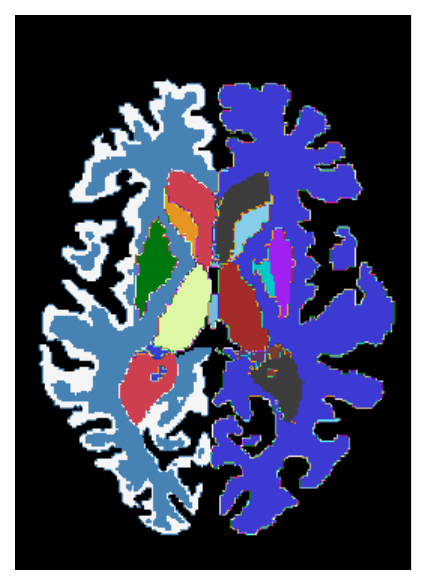}
    \includegraphics[width=\imagewidth]{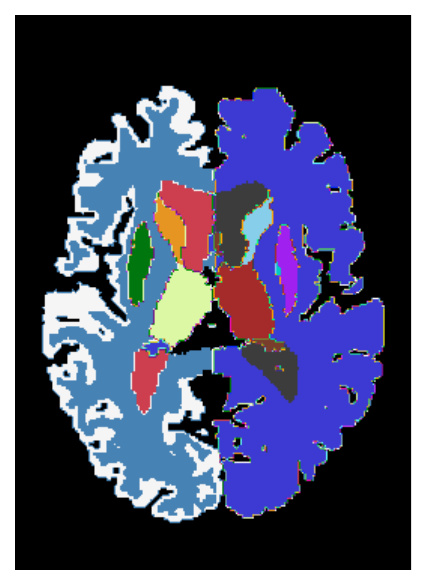}

    \hrule

    \rotatebox{90}{\adjustbox{valign=m}{\hspace{2.2em}Ours}}
    \includegraphics[width=\imagewidth]{figures/oasis_results/oasis_00/moved_img_ours.pdf}
    \includegraphics[width=\imagewidth]{figures/oasis_results/oasis_01/moved_img_ours.pdf}
    \includegraphics[width=\imagewidth]{figures/oasis_results/oasis_02/moved_img_ours.pdf}
    \includegraphics[width=\imagewidth]{figures/oasis_results/oasis_03/moved_img_ours.pdf}
    \includegraphics[width=\imagewidth]{figures/oasis_results/oasis_04/moved_img_ours.pdf}
    \includegraphics[width=\imagewidth]{figures/oasis_results/oasis_05/moved_img_ours.pdf}
    \includegraphics[width=\imagewidth]{figures/oasis_results/oasis_06/moved_img_ours.pdf}
    \includegraphics[width=\imagewidth]{figures/oasis_results/oasis_07/moved_img_ours.pdf}

    \rotatebox{90}{\adjustbox{valign=m}{\hspace{2.2em}Ours}}
    \includegraphics[width=\imagewidth]{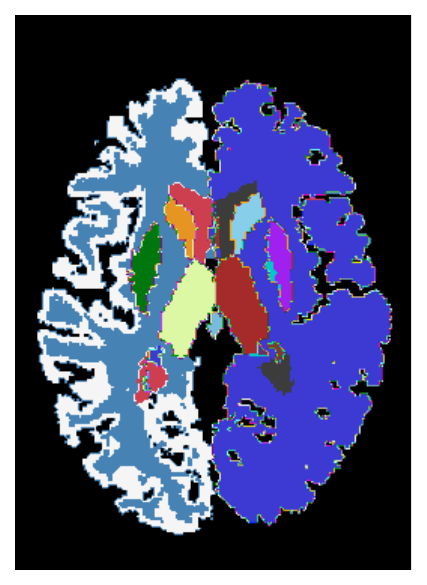}
    \includegraphics[width=\imagewidth]{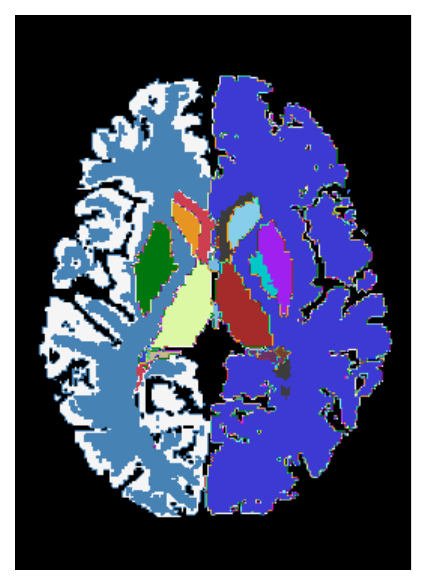}
    \includegraphics[width=\imagewidth]{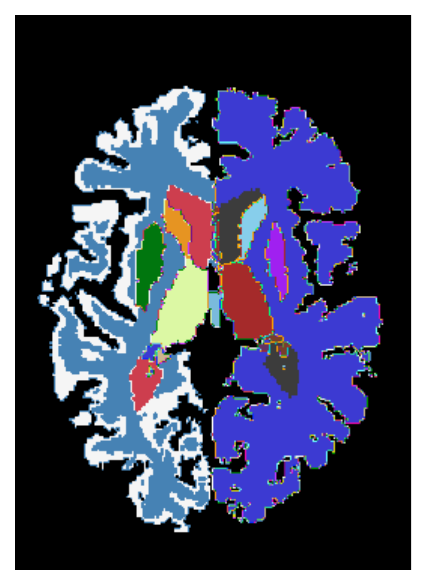}
    \includegraphics[width=\imagewidth]{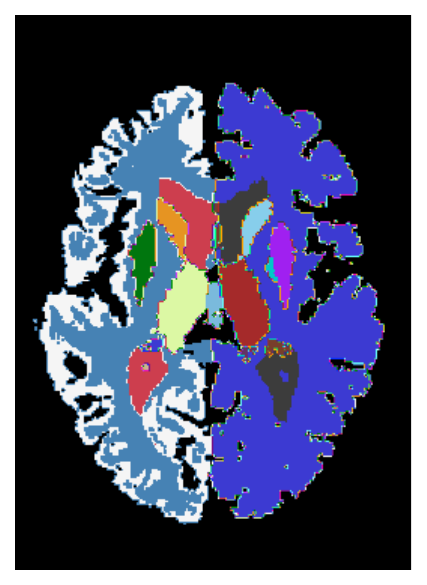}
    \includegraphics[width=\imagewidth]{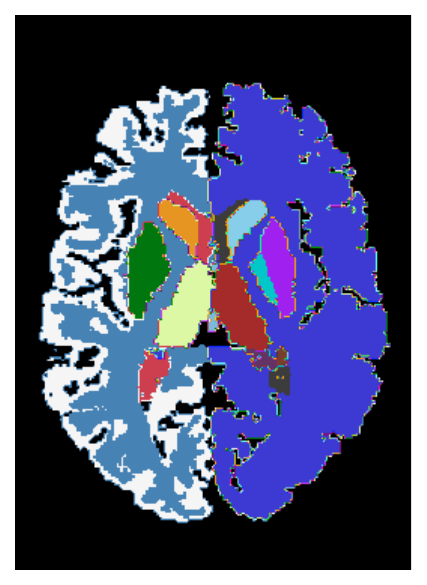}
    \includegraphics[width=\imagewidth]{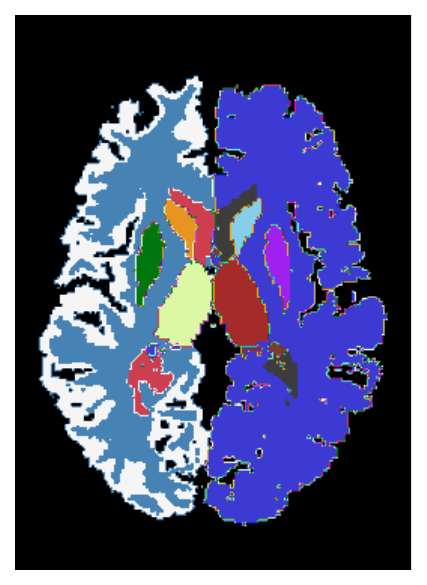}
    \includegraphics[width=\imagewidth]{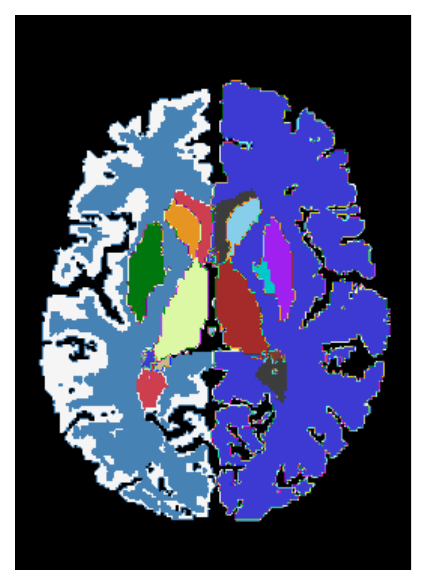}
    \includegraphics[width=\imagewidth]{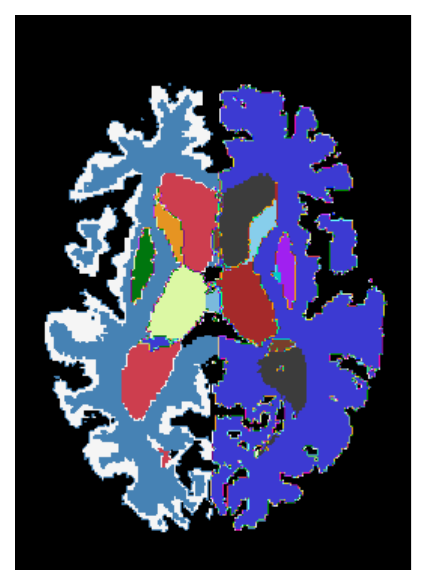}

    \hrule

    \rotatebox{90}{\adjustbox{valign=m}{\hspace{0.5em}Fixed Image}}
    \includegraphics[width=\imagewidth]{figures/oasis_results/oasis_00/fixed_img.pdf}
    \includegraphics[width=\imagewidth]{figures/oasis_results/oasis_01/fixed_img.pdf}
    \includegraphics[width=\imagewidth]{figures/oasis_results/oasis_02/fixed_img.pdf}
    \includegraphics[width=\imagewidth]{figures/oasis_results/oasis_03/fixed_img.pdf}
    \includegraphics[width=\imagewidth]{figures/oasis_results/oasis_04/fixed_img.pdf}
    \includegraphics[width=\imagewidth]{figures/oasis_results/oasis_05/fixed_img.pdf}
    \includegraphics[width=\imagewidth]{figures/oasis_results/oasis_06/fixed_img.pdf}
    \includegraphics[width=\imagewidth]{figures/oasis_results/oasis_07/fixed_img.pdf}
    
    \rotatebox{90}{\adjustbox{valign=m}{\hspace{0.5em}Fixed Seg}}
    \includegraphics[width=\imagewidth]{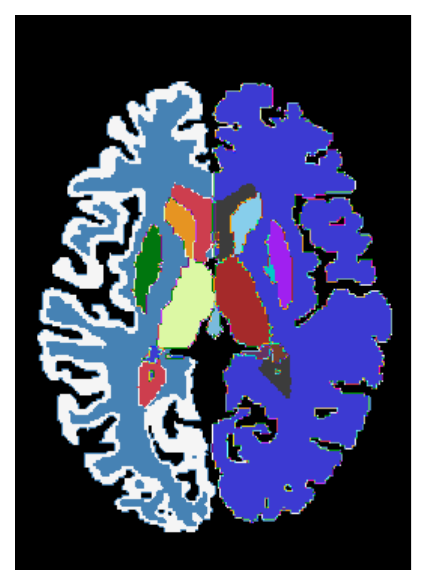}
    \includegraphics[width=\imagewidth]{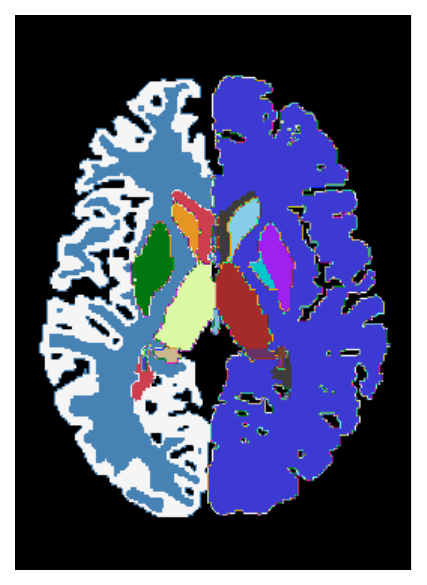}
    \includegraphics[width=\imagewidth]{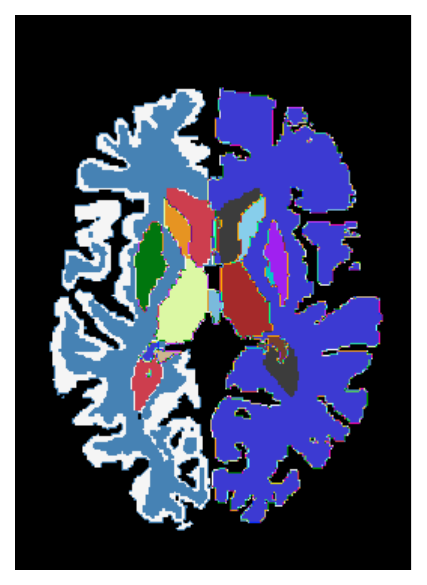}
    \includegraphics[width=\imagewidth]{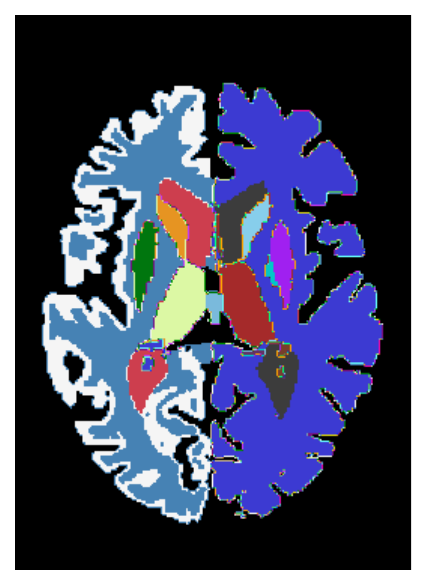}
    \includegraphics[width=\imagewidth]{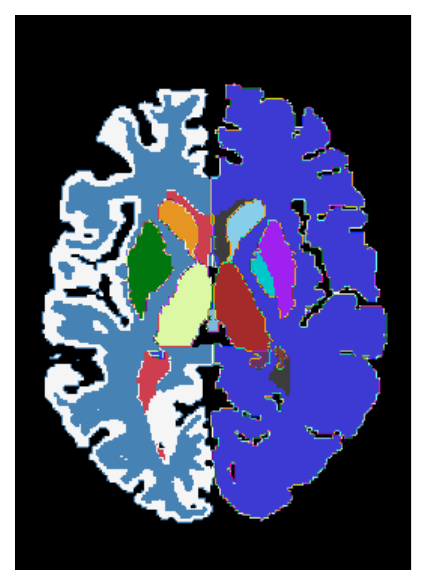}
    \includegraphics[width=\imagewidth]{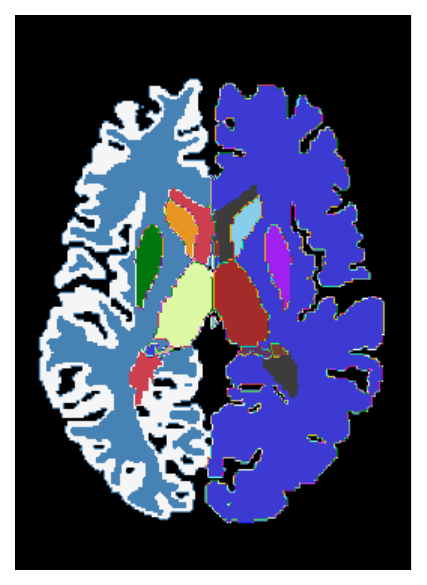}
    \includegraphics[width=\imagewidth]{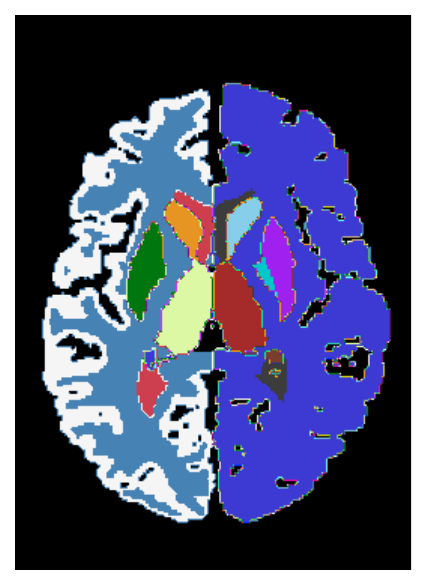}
    \includegraphics[width=\imagewidth]{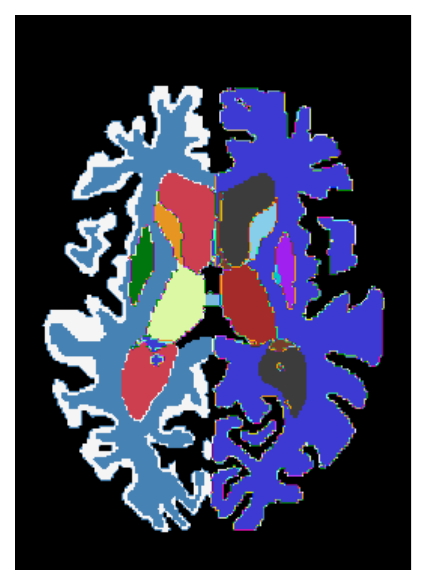}
    
    \rotatebox{90}{\adjustbox{valign=m}{\hspace{0.5em}Moving Image}}
    \includegraphics[width=\imagewidth]{figures/oasis_results/oasis_00/moving_img.pdf}
    \includegraphics[width=\imagewidth]{figures/oasis_results/oasis_01/moving_img.pdf}
    \includegraphics[width=\imagewidth]{figures/oasis_results/oasis_02/moving_img.pdf}
    \includegraphics[width=\imagewidth]{figures/oasis_results/oasis_03/moving_img.pdf}
    \includegraphics[width=\imagewidth]{figures/oasis_results/oasis_04/moving_img.pdf}
    \includegraphics[width=\imagewidth]{figures/oasis_results/oasis_05/moving_img.pdf}
    \includegraphics[width=\imagewidth]{figures/oasis_results/oasis_06/moving_img.pdf}
    \includegraphics[width=\imagewidth]{figures/oasis_results/oasis_07/moving_img.pdf}

    \rotatebox{90}{\adjustbox{valign=m}{\hspace{0.5em}Moving Seg}}
    \includegraphics[width=\imagewidth]{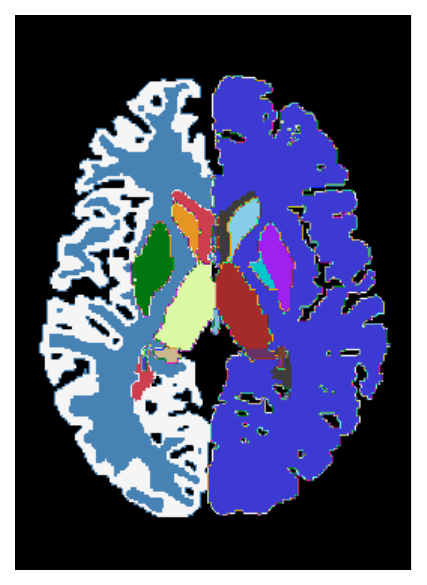}
    \includegraphics[width=\imagewidth]{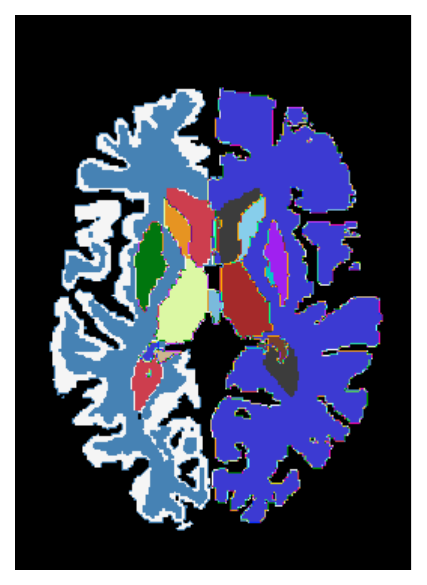}
    \includegraphics[width=\imagewidth]{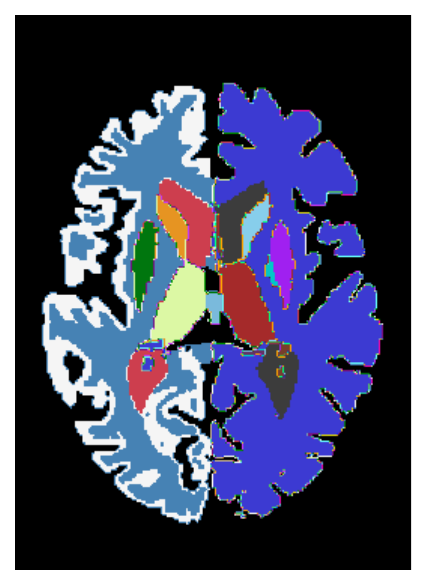}
    \includegraphics[width=\imagewidth]{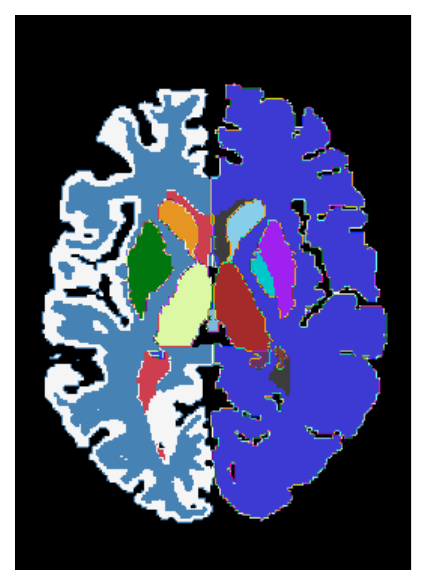}
    \includegraphics[width=\imagewidth]{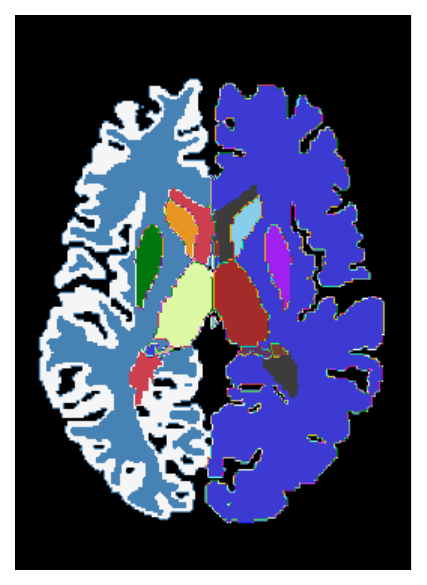}
    \includegraphics[width=\imagewidth]{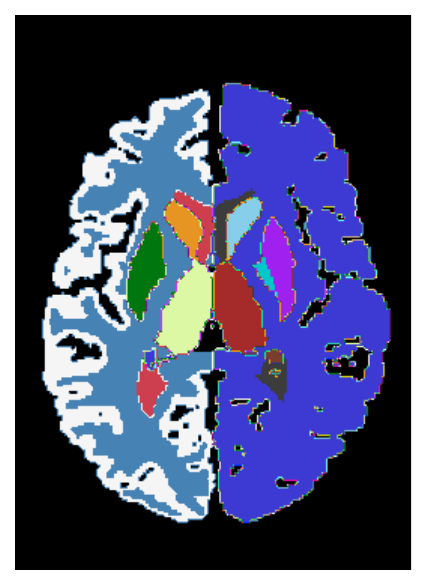}
    \includegraphics[width=\imagewidth]{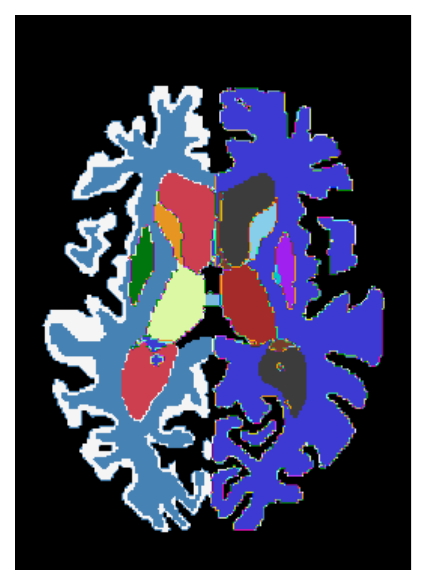}
    \includegraphics[width=\imagewidth]{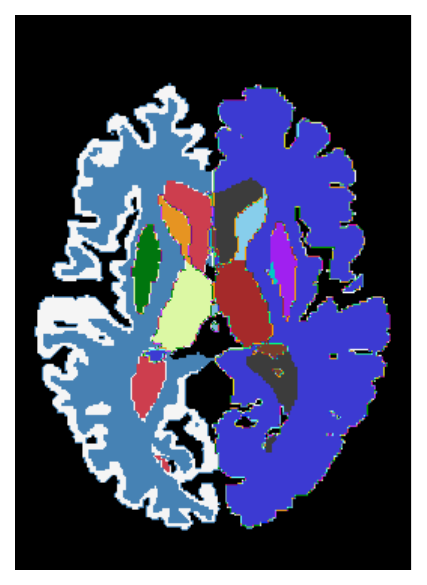}
    
    \caption{\textbf{Qualitative comparison of KeyMorph and our method on OASIS dataset.} 
    Qualitative evaluation of both labelmaps and intensity images shows that dense features from our method are instrumental in being robust and accurately registering complex deformable structures compared to sparse keypoints.
    }
    \label{fig:app-qual-oasis}
\end{figure*}

\begin{figure*}[htpb!]
    \setlength{\imagewidth}{0.115\linewidth}  %
    \centering

    \rotatebox{90}{\adjustbox{valign=m}{\hspace{1em}KeyMorph}}
    \includegraphics[width=\imagewidth]{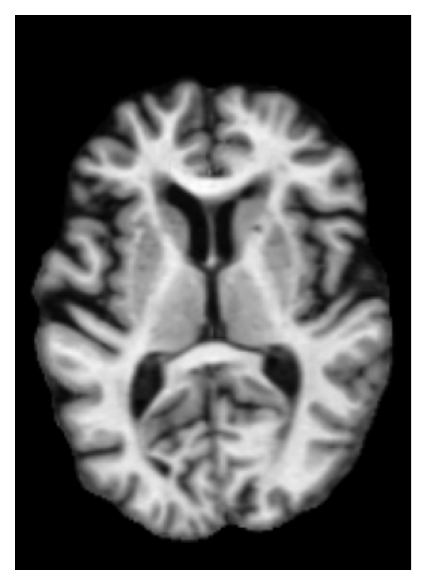}
    \includegraphics[width=\imagewidth]{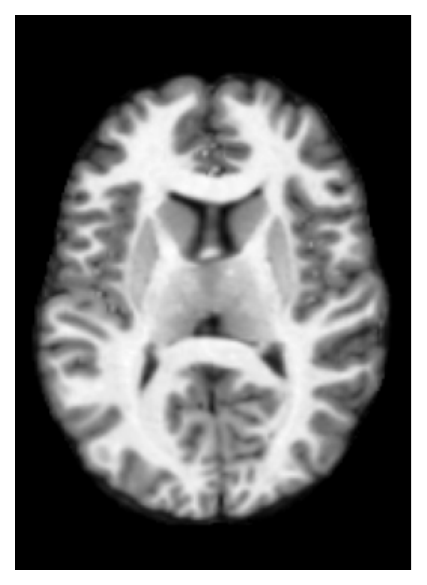}
    \includegraphics[width=\imagewidth]{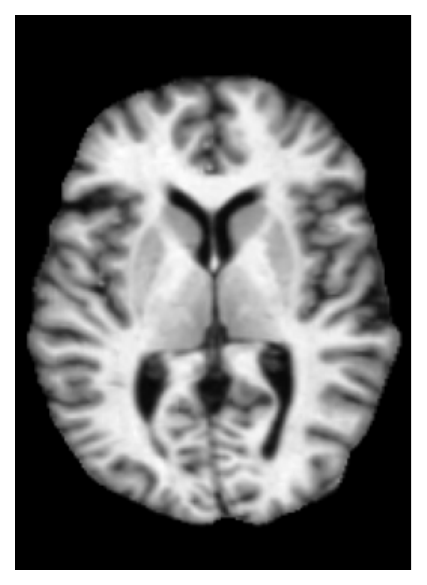}
    \includegraphics[width=\imagewidth]{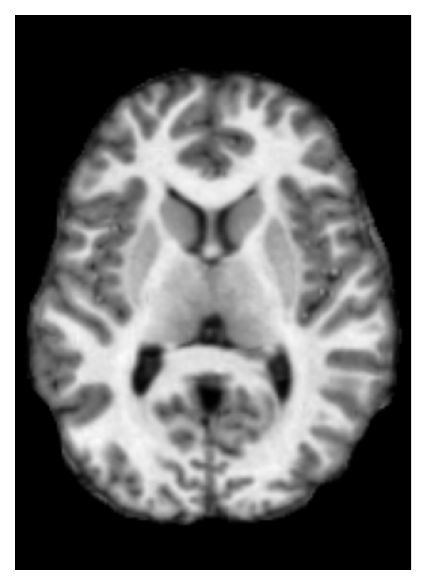}
    \includegraphics[width=\imagewidth]{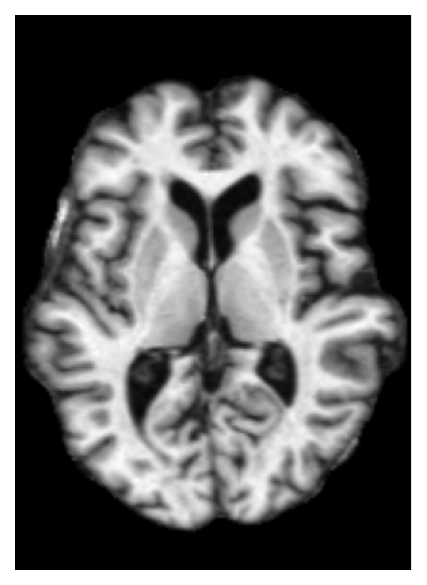}
    \includegraphics[width=\imagewidth]{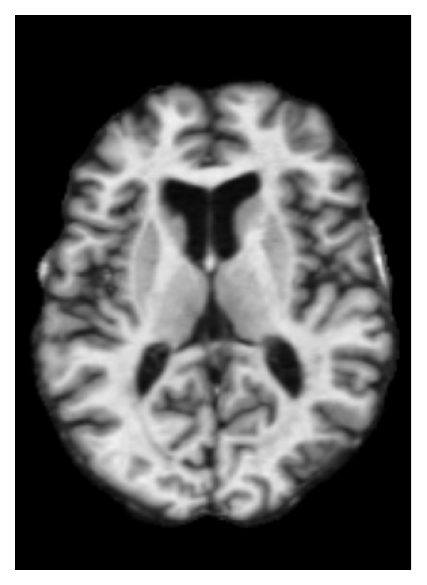}
    \includegraphics[width=\imagewidth]{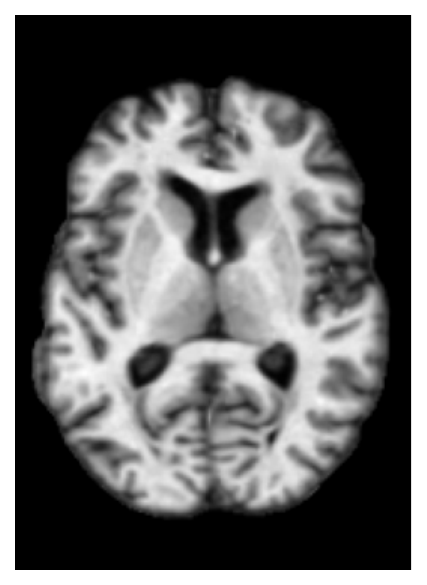}
    \includegraphics[width=\imagewidth]{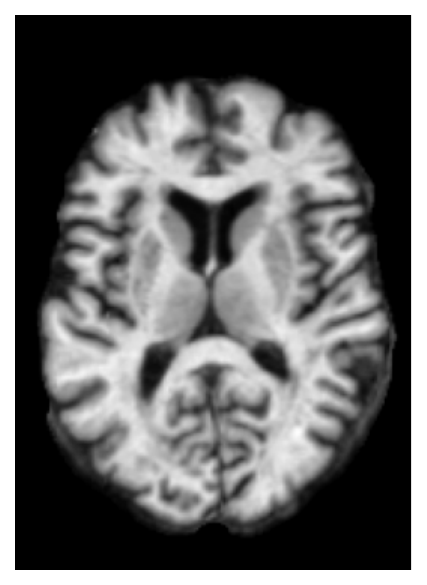}

    \rotatebox{90}{\adjustbox{valign=m}{\hspace{1em}KeyMorph}}
    \includegraphics[width=\imagewidth]{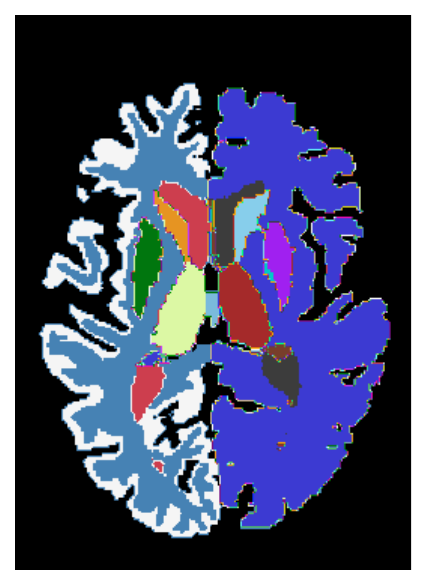}
    \includegraphics[width=\imagewidth]{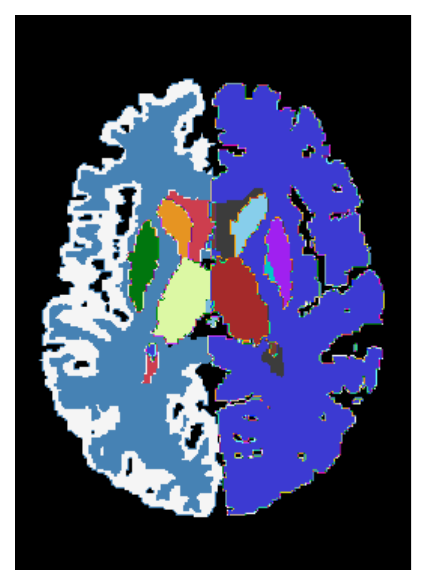}
    \includegraphics[width=\imagewidth]{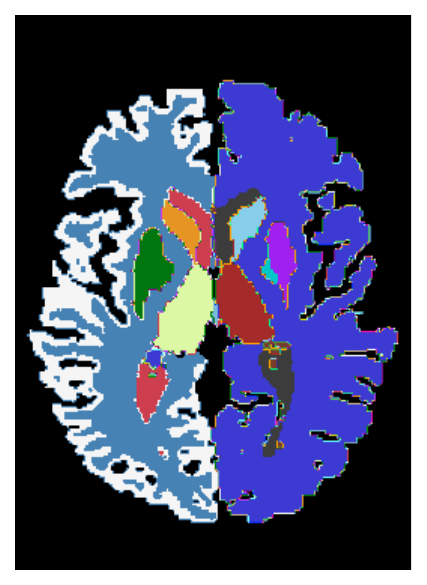}
    \includegraphics[width=\imagewidth]{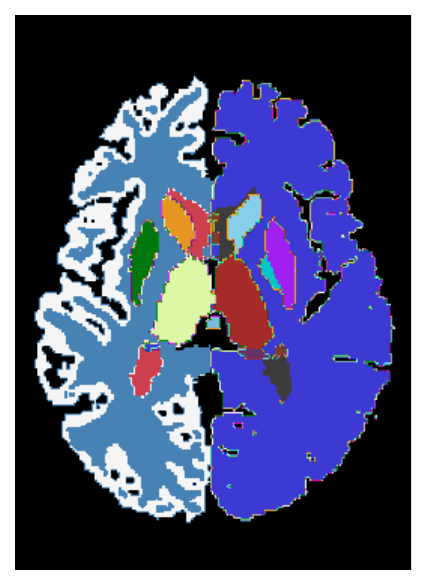}
    \includegraphics[width=\imagewidth]{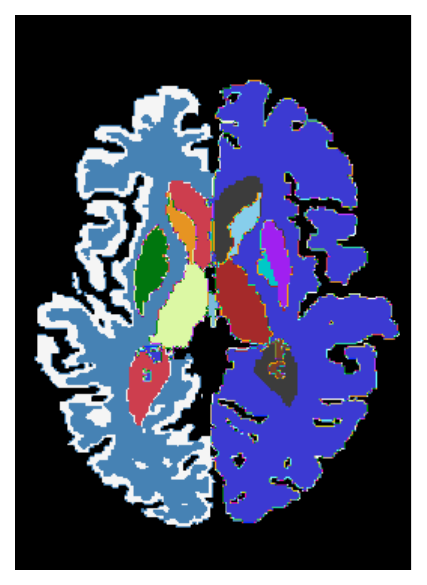}
    \includegraphics[width=\imagewidth]{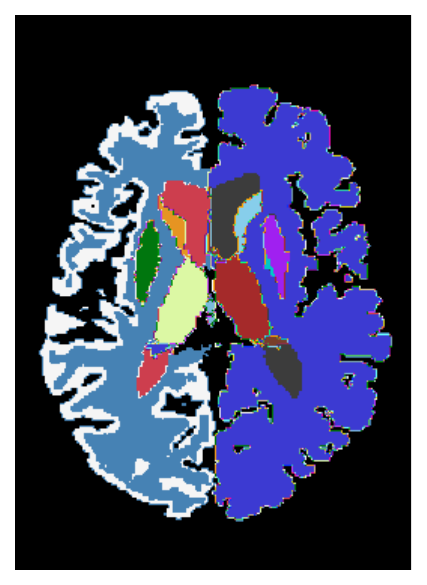}
    \includegraphics[width=\imagewidth]{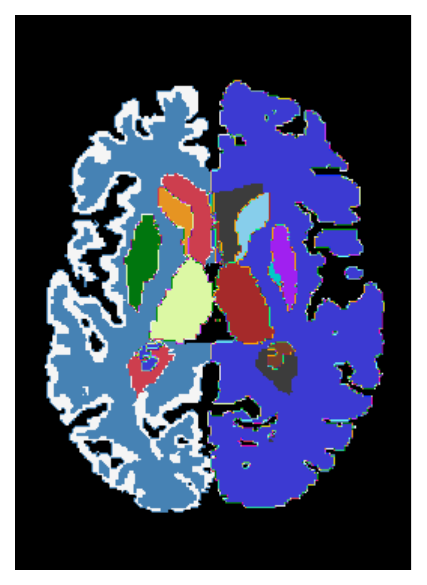}
    \includegraphics[width=\imagewidth]{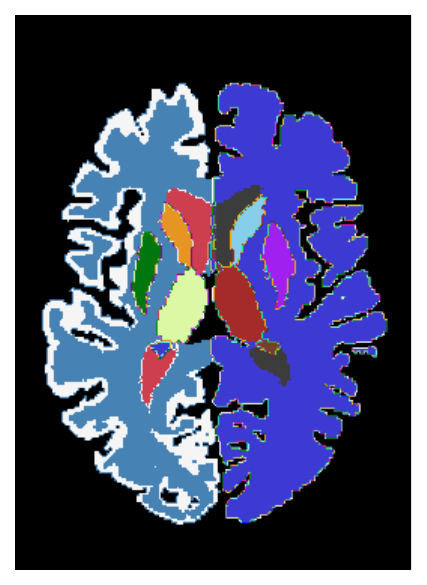}

    \hrule

    \rotatebox{90}{\adjustbox{valign=m}{\hspace{2.2em}Ours}}
    \includegraphics[width=\imagewidth]{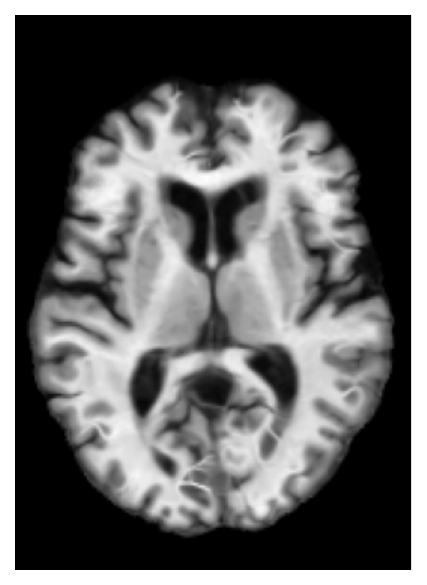}
    \includegraphics[width=\imagewidth]{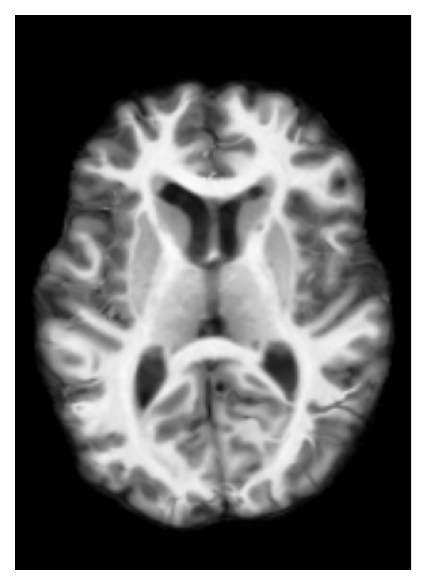}
    \includegraphics[width=\imagewidth]{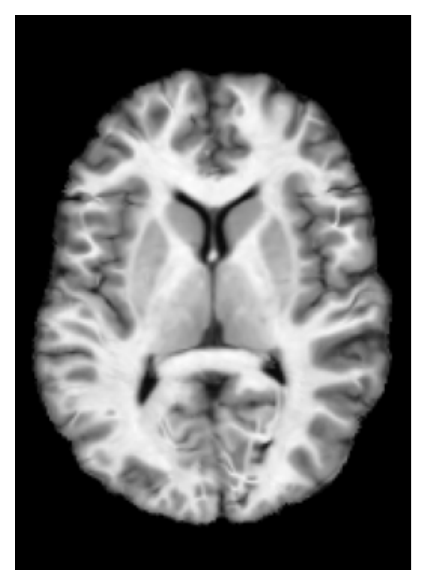}
    \includegraphics[width=\imagewidth]{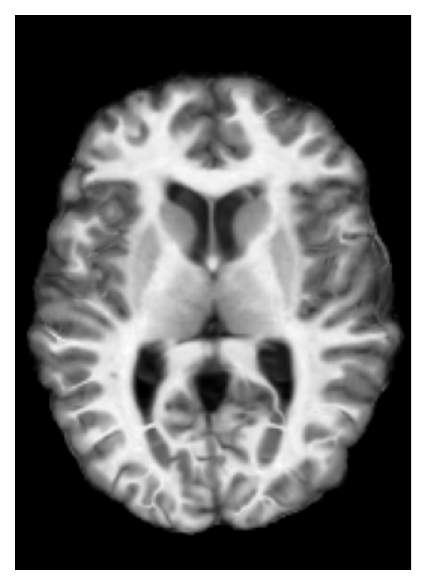}
    \includegraphics[width=\imagewidth]{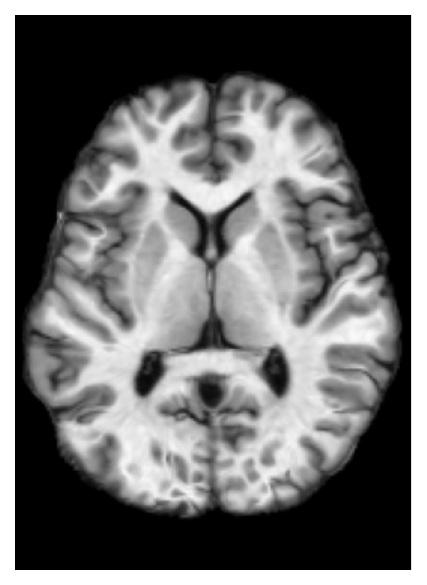}
    \includegraphics[width=\imagewidth]{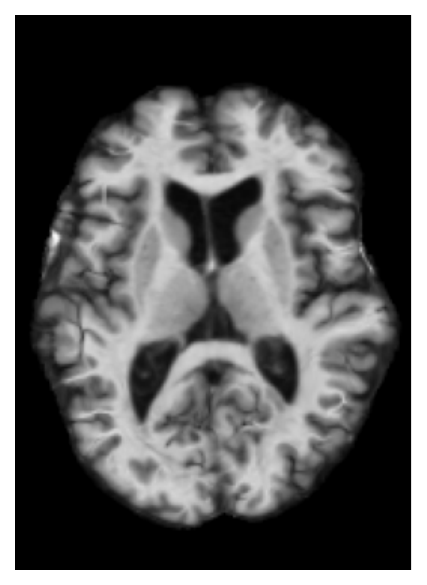}
    \includegraphics[width=\imagewidth]{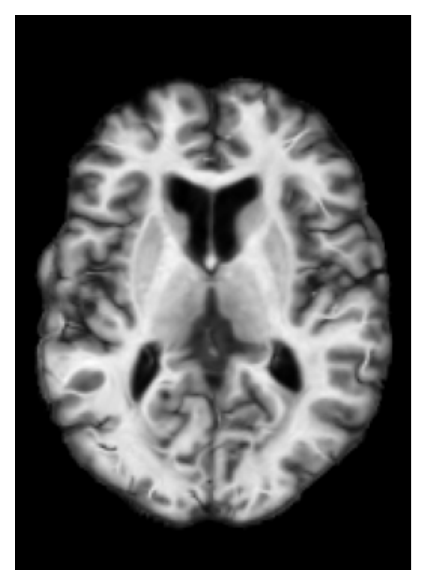}
    \includegraphics[width=\imagewidth]{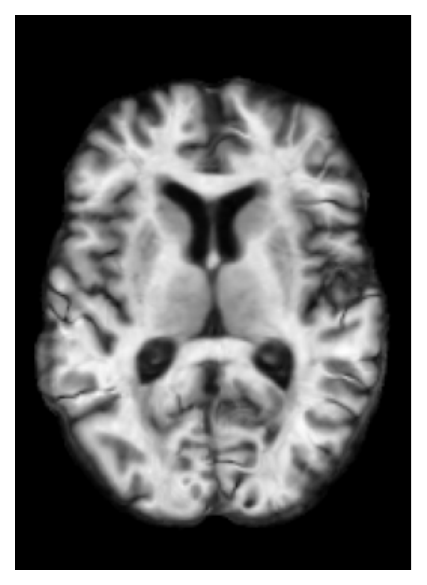}

    \rotatebox{90}{\adjustbox{valign=m}{\hspace{2.2em}Ours}}
    \includegraphics[width=\imagewidth]{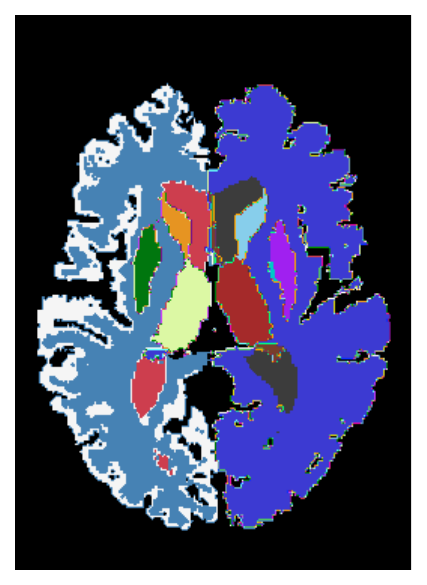}
    \includegraphics[width=\imagewidth]{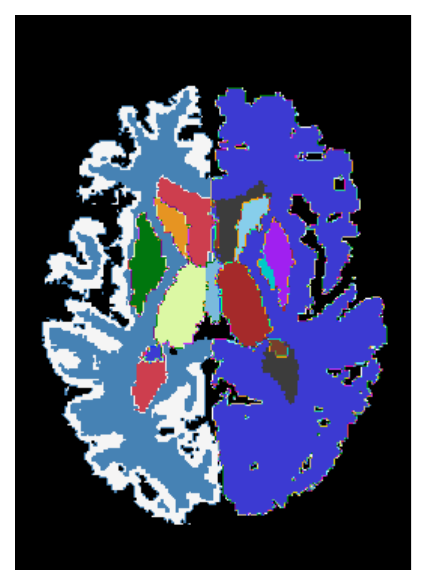}
    \includegraphics[width=\imagewidth]{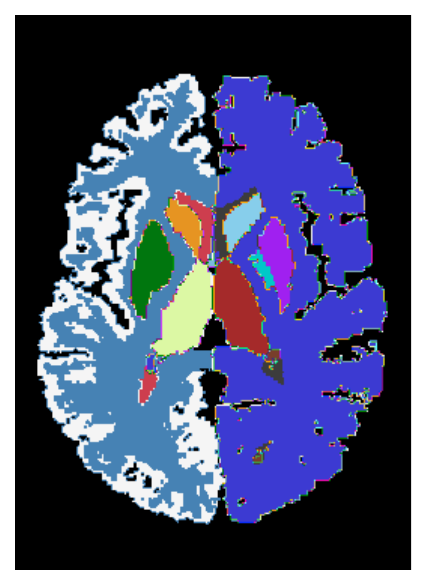}
    \includegraphics[width=\imagewidth]{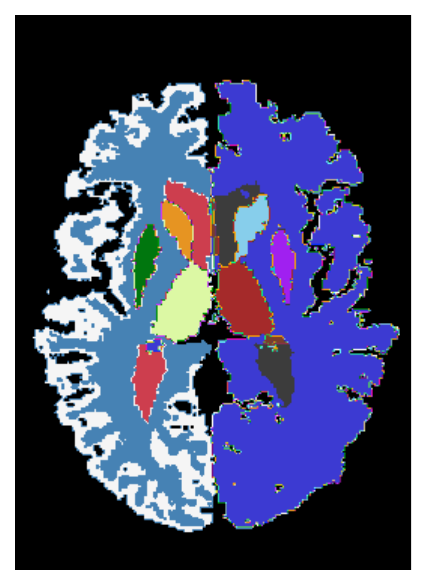}
    \includegraphics[width=\imagewidth]{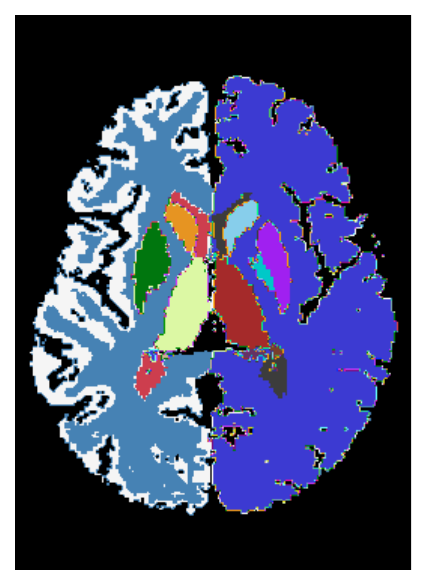}
    \includegraphics[width=\imagewidth]{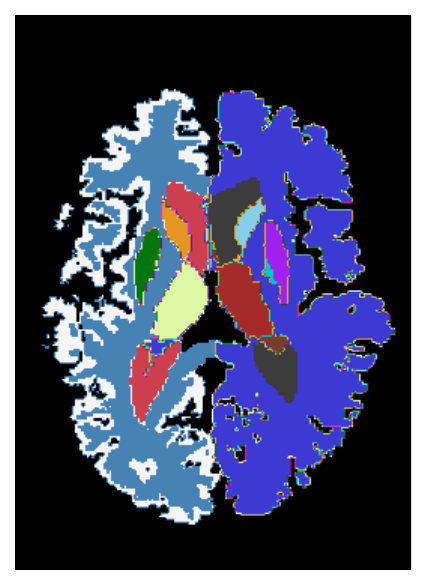}
    \includegraphics[width=\imagewidth]{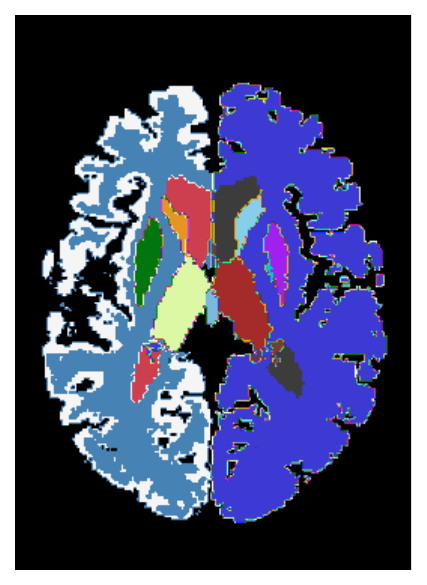}
    \includegraphics[width=\imagewidth]{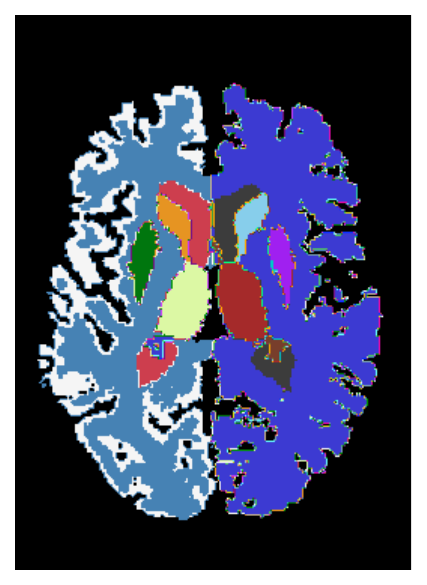}

    \hrule

    \rotatebox{90}{\adjustbox{valign=m}{\hspace{0.5em}Fixed Image}}
    \includegraphics[width=\imagewidth]{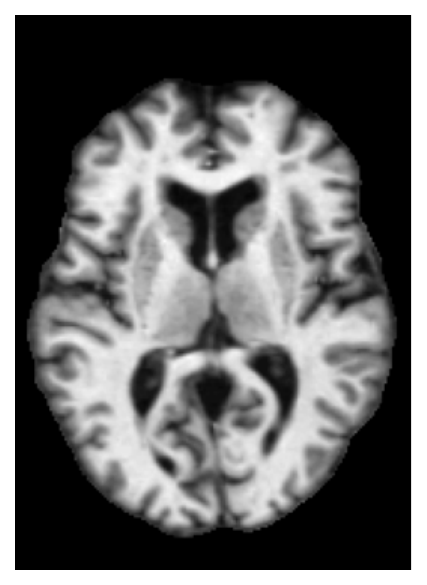}
    \includegraphics[width=\imagewidth]{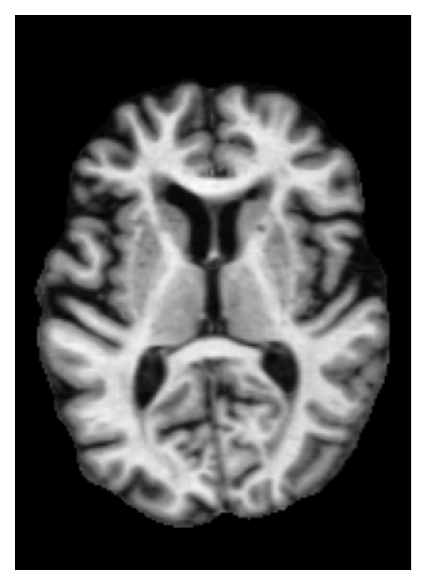}
    \includegraphics[width=\imagewidth]{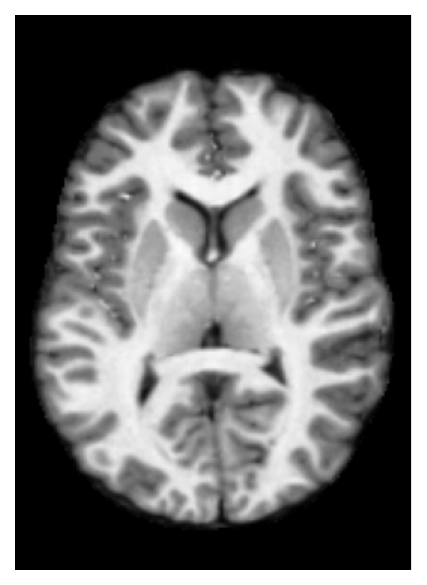}
    \includegraphics[width=\imagewidth]{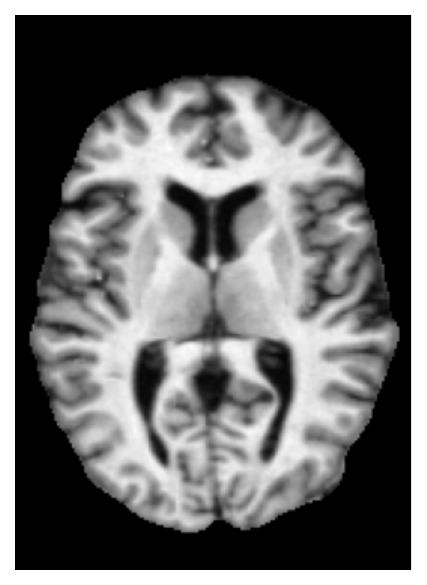}
    \includegraphics[width=\imagewidth]{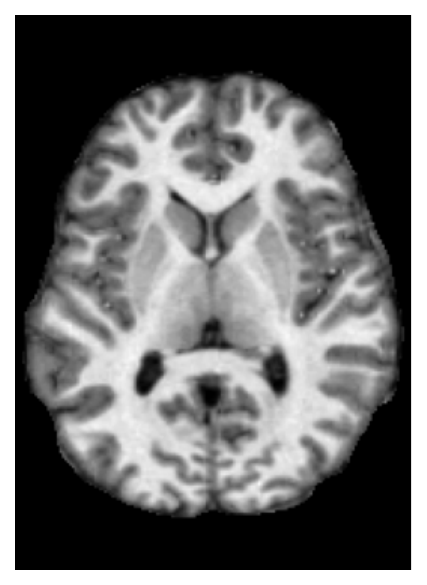}
    \includegraphics[width=\imagewidth]{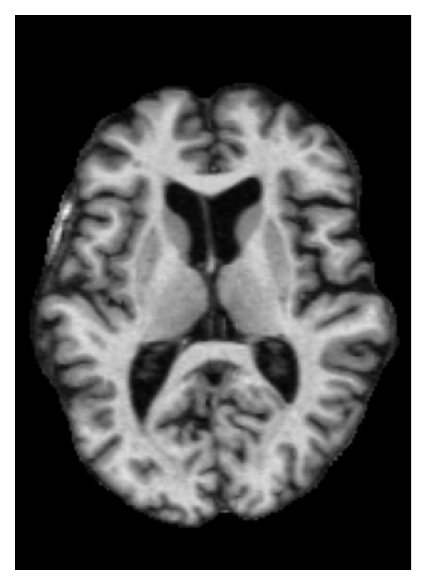}
    \includegraphics[width=\imagewidth]{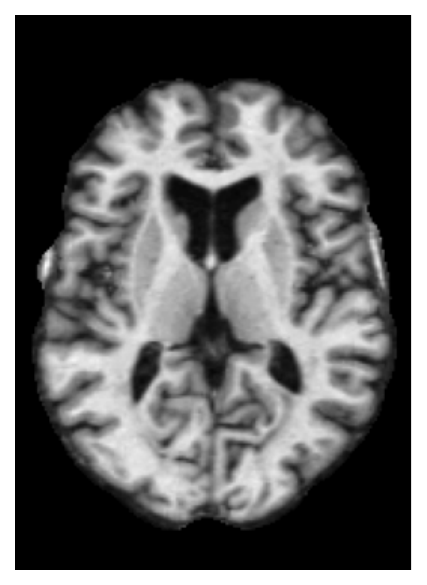}
    \includegraphics[width=\imagewidth]{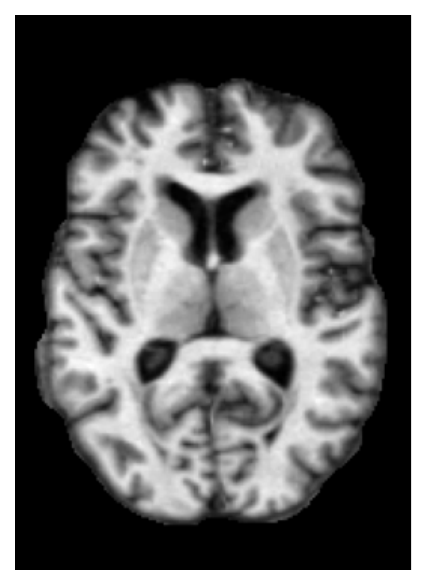}
    
    \rotatebox{90}{\adjustbox{valign=m}{\hspace{0.5em}Fixed Seg}}
    \includegraphics[width=\imagewidth]{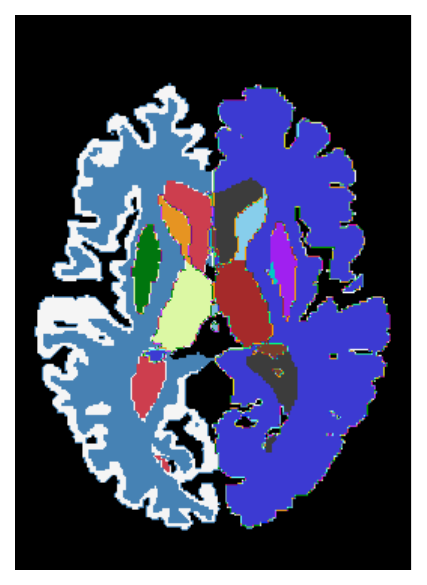}
    \includegraphics[width=\imagewidth]{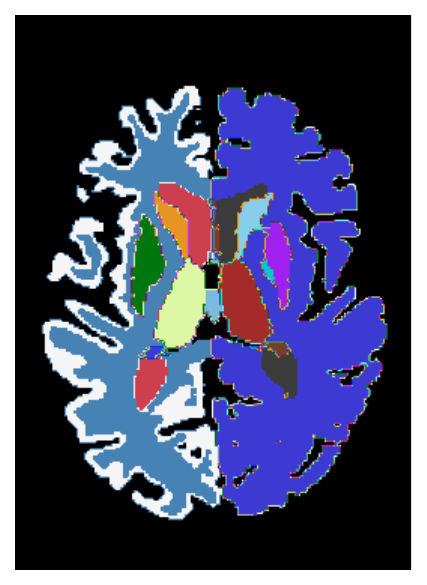}
    \includegraphics[width=\imagewidth]{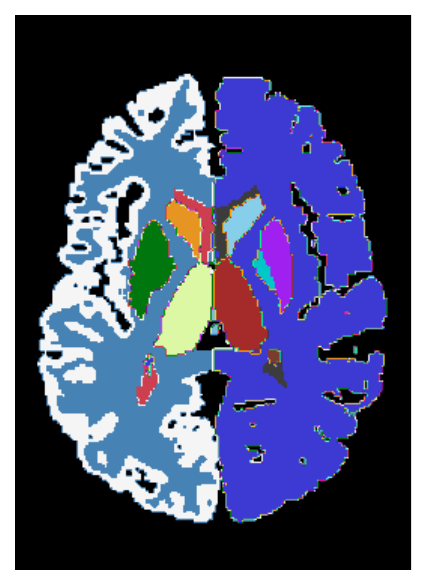}
    \includegraphics[width=\imagewidth]{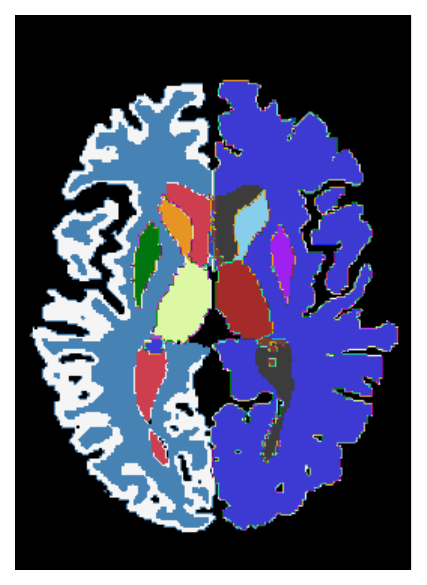}
    \includegraphics[width=\imagewidth]{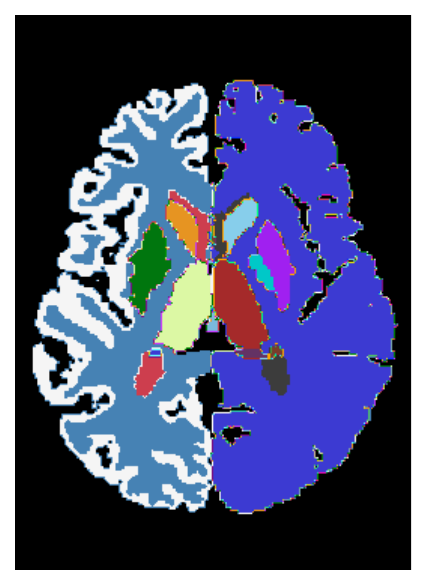}
    \includegraphics[width=\imagewidth]{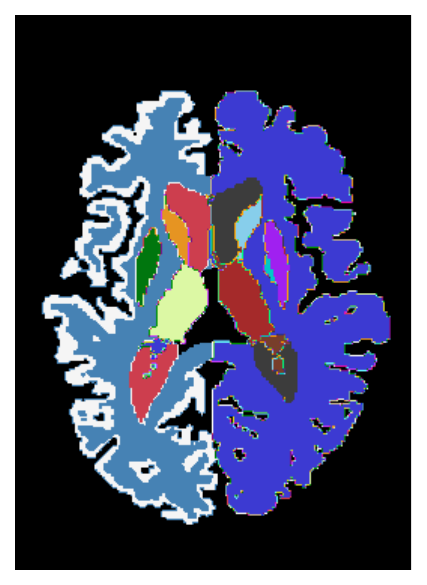}
    \includegraphics[width=\imagewidth]{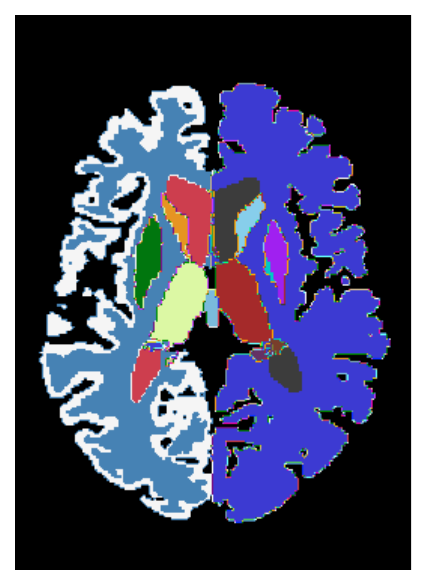}
    \includegraphics[width=\imagewidth]{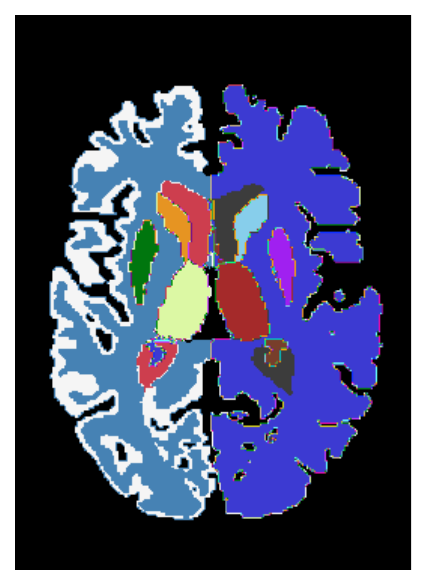}
    
    \rotatebox{90}{\adjustbox{valign=m}{\hspace{0.5em}Moving Image}}
    \includegraphics[width=\imagewidth]{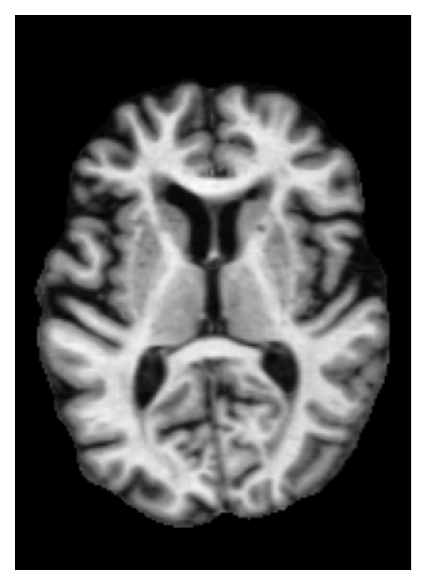}
    \includegraphics[width=\imagewidth]{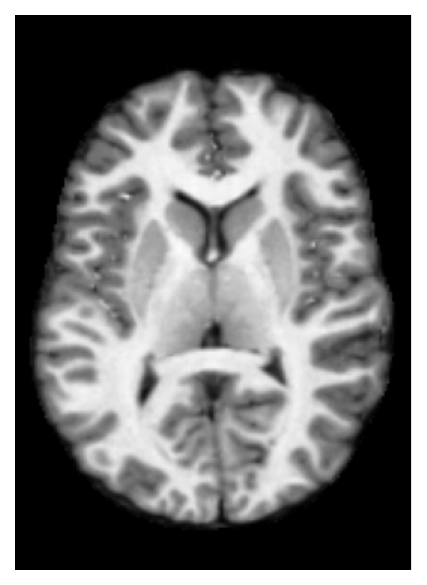}
    \includegraphics[width=\imagewidth]{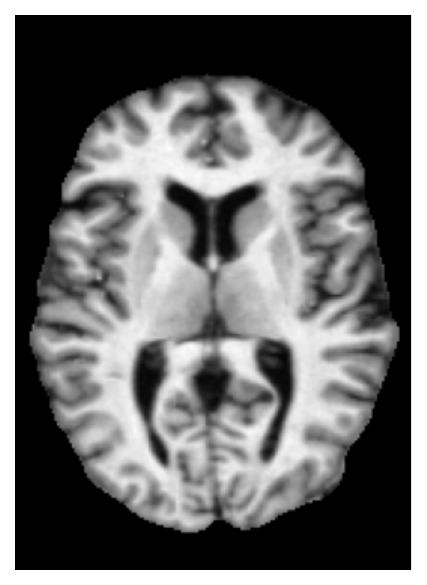}
    \includegraphics[width=\imagewidth]{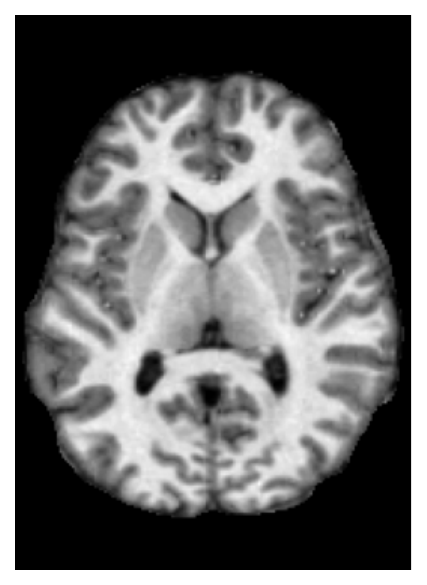}
    \includegraphics[width=\imagewidth]{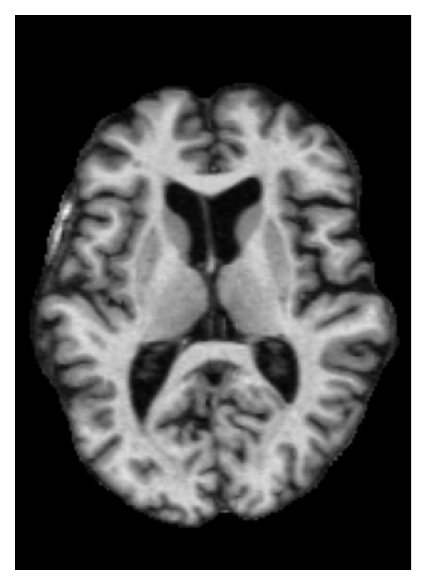}
    \includegraphics[width=\imagewidth]{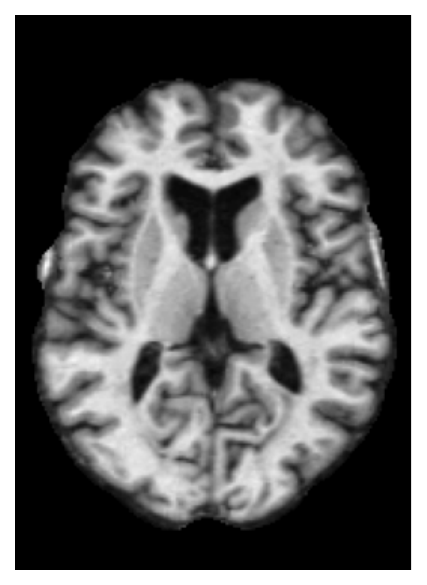}
    \includegraphics[width=\imagewidth]{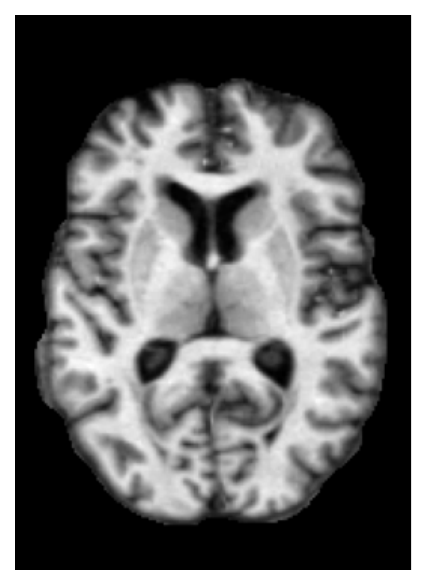}
    \includegraphics[width=\imagewidth]{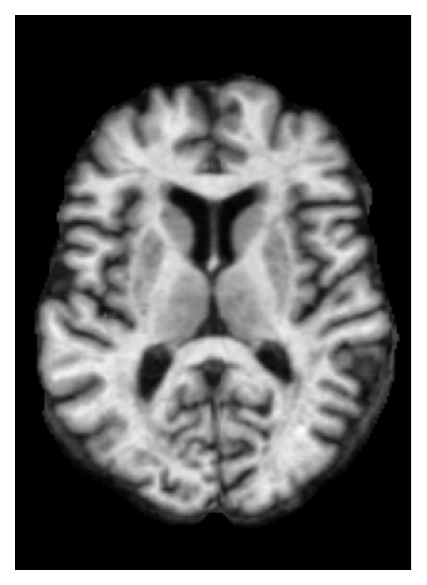}

    \rotatebox{90}{\adjustbox{valign=m}{\hspace{0.5em}Moving Seg}}
    \includegraphics[width=\imagewidth]{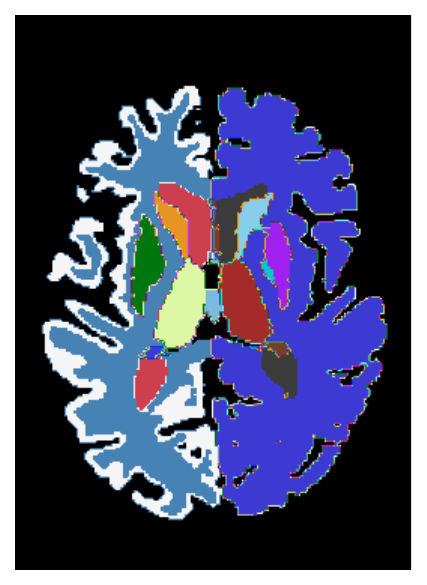}
    \includegraphics[width=\imagewidth]{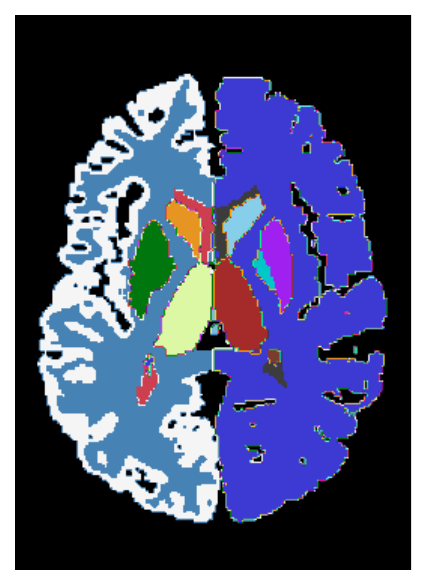}
    \includegraphics[width=\imagewidth]{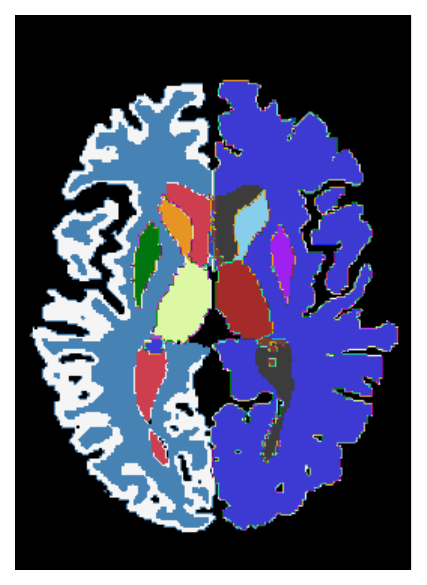}
    \includegraphics[width=\imagewidth]{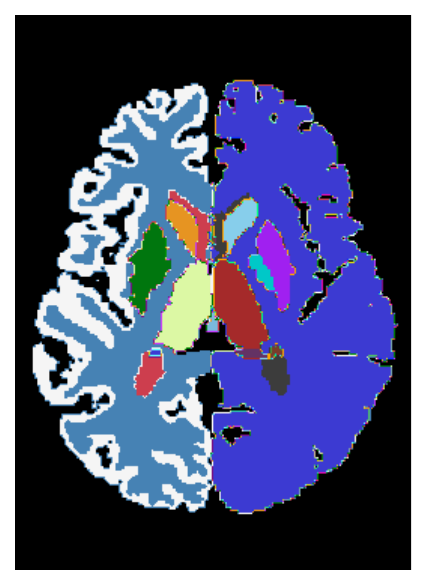}
    \includegraphics[width=\imagewidth]{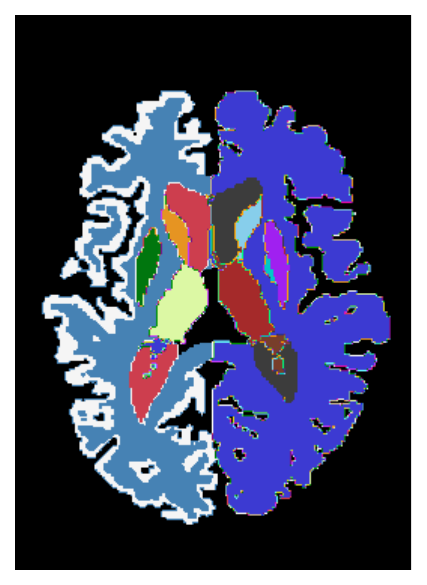}
    \includegraphics[width=\imagewidth]{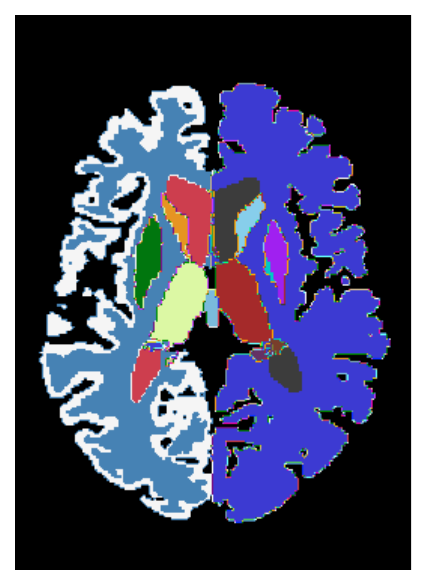}
    \includegraphics[width=\imagewidth]{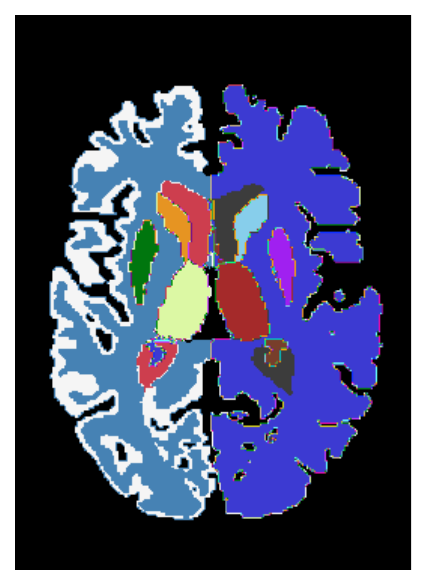}
    \includegraphics[width=\imagewidth]{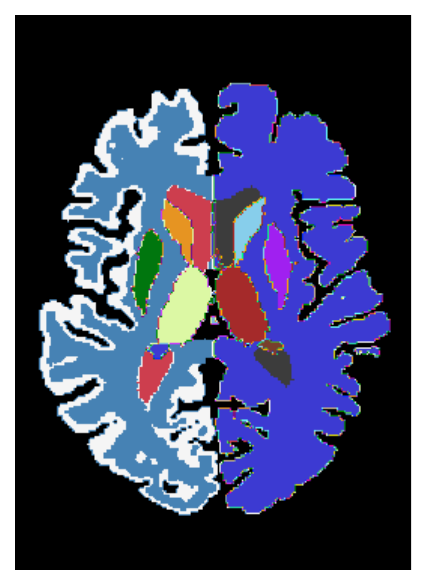}
    
    \caption{\textbf{Qualitative comparison of KeyMorph and our method on OASIS dataset.} 
    Qualitative evaluation of both labelmaps and intensity images shows that dense features from our method are instrumental in being robust and accurately registering complex deformable structures compared to sparse keypoints.
    }
    \label{fig:app-qual-oasis-2}
\end{figure*}

\subsection{Datasets}
\label{app:datasets}

We consider four brain MRI datasets in this paper: OASIS dataset for in-distribution performance, and LPBA40, IBSR18, and CUMC12 datasets for out-of-distribution performance~\cite{shattuck2008construction,ibsr,klein_evaluation_2009,oasisdataset}.
More details about the datasets are provided below.

\begin{itemize}[leftmargin=*]
    \item{\textbf{OASIS}. The Open Access Series of Imaging Studies (OASIS) dataset contains 414 T1-weighted brain images in Young, Middle Aged, Nondemented, and Demented Older adults. The images are skull-stripped and bias-corrected, followed by a resampling and afine alignment to the FreeSurfer's Talairach atlas. Label segmentations of 35 subcortical structures were obtained using automatic segmentation using Freesurfer software.}

    \item{\textbf{LPBA40}. 40 brain images and their labels are used to construct the LONI Probabilistic Brain Atlas (LPBA40) dataset at the Laboratory of Neuroimaging (LONI) at UCLA~\cite{shattuck2008construction}.
    All volumes are preprocessed according to LONI protocols to produce skull-stripped volumes.
    These volumes are aligned to the MNI305 atlas -- this is relevant since existing DLIR methods may be biased towards images that are aligned to the Talairach and Tournoux (1988) atlas which is used to align the images in the OASIS dataset.
    This is followed by a custom manual labelling protocol of 56 structures from each of the volumes.
    Bias correction is perfrmed using the BrainSuite's Bias Field Corrector.}

    \item{\textbf{IBSR18}. the Internet Brain Segmentation Repository contains 18 different brain images acquired at different laboratories as IBSRv2.0. The dataset consists of T1-weighted brains aligned to the Talairach and Tournoux (1988) atlas, and manually segmented into 84 labelled regions. Bias correction of the images are performed using the `autoseg' bias field correction algorithm.}

    \item{\textbf{CUMC12}. The Columbia University Medical Center dataset contains 12 T1-weighted brain images with manual segmentation of 128 regions. The images were scanned on a 1.5T GE scanner, and the images were resliced coronally to a slice thickness of 3mm, rotated into cardinal orientation, and segmented by a technician trained according to the Cardviews labelling scheme.}
\end{itemize}

\begin{figure*}[ht!]
   \centering 
   \includegraphics[width=\linewidth]{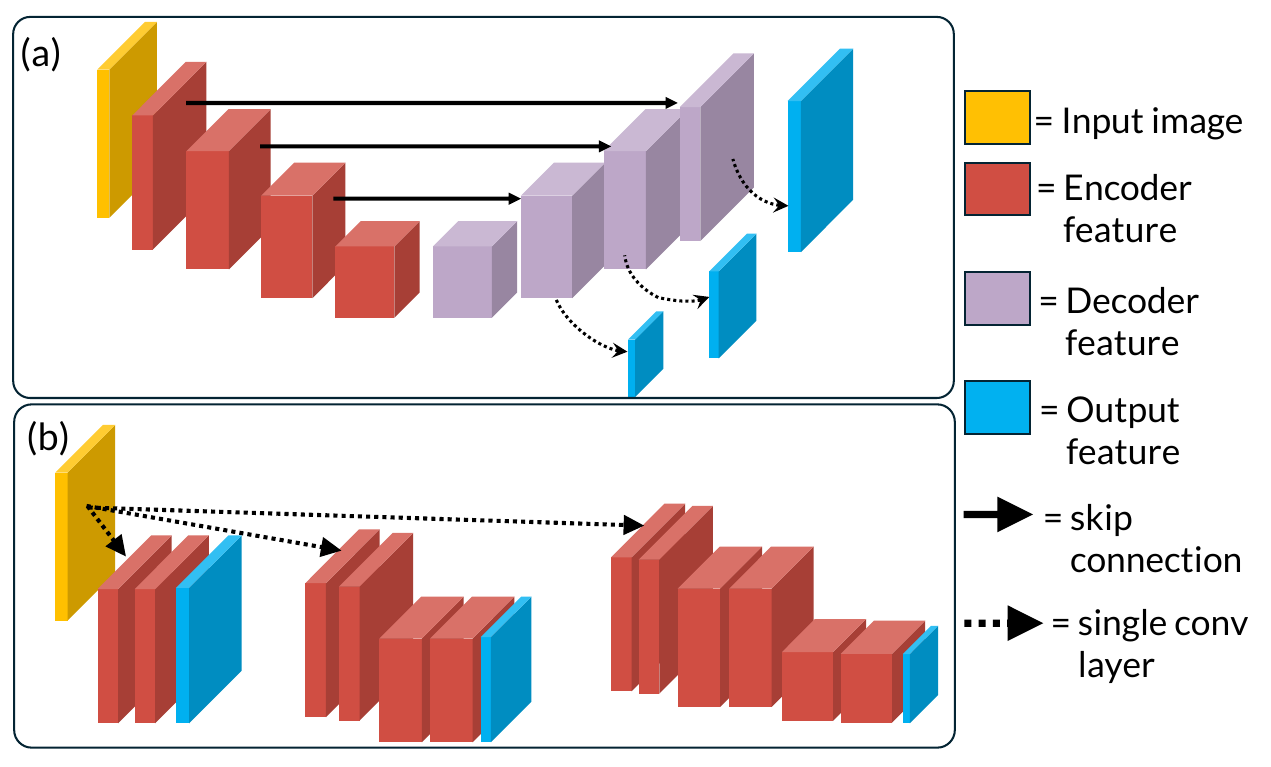}
   \caption{\textbf{Architecture details}. \textbf{(a)} illustrates the UNet and Large Kernel U-Net (LKUNet) architecture designs, which consists of encoder blocks (red) and decoder blocks (purple) linked using skip connections. Multi-scale features are extracted from the intermediate decoder layers using a single convolutional layer. This design leads to shared features across multiple scales. UNet and LKUNet differ in the kernel parameters within each encoder and decoder blocks. \textbf{(b)} illustrates the `Encoder-Only' versions of the same networks. The decoder path is entirely discarded, and each feature image is extracted using a separate encoder. This design enables independent learning of each multi-scale feature.}
   \label{fig:architectures}
\end{figure*}

\subsection{Convergence of KeyMorph on OASIS}
\label{app:km-convergence}
\begin{figure*}
    \centering
    \includegraphics[width=0.48\linewidth]{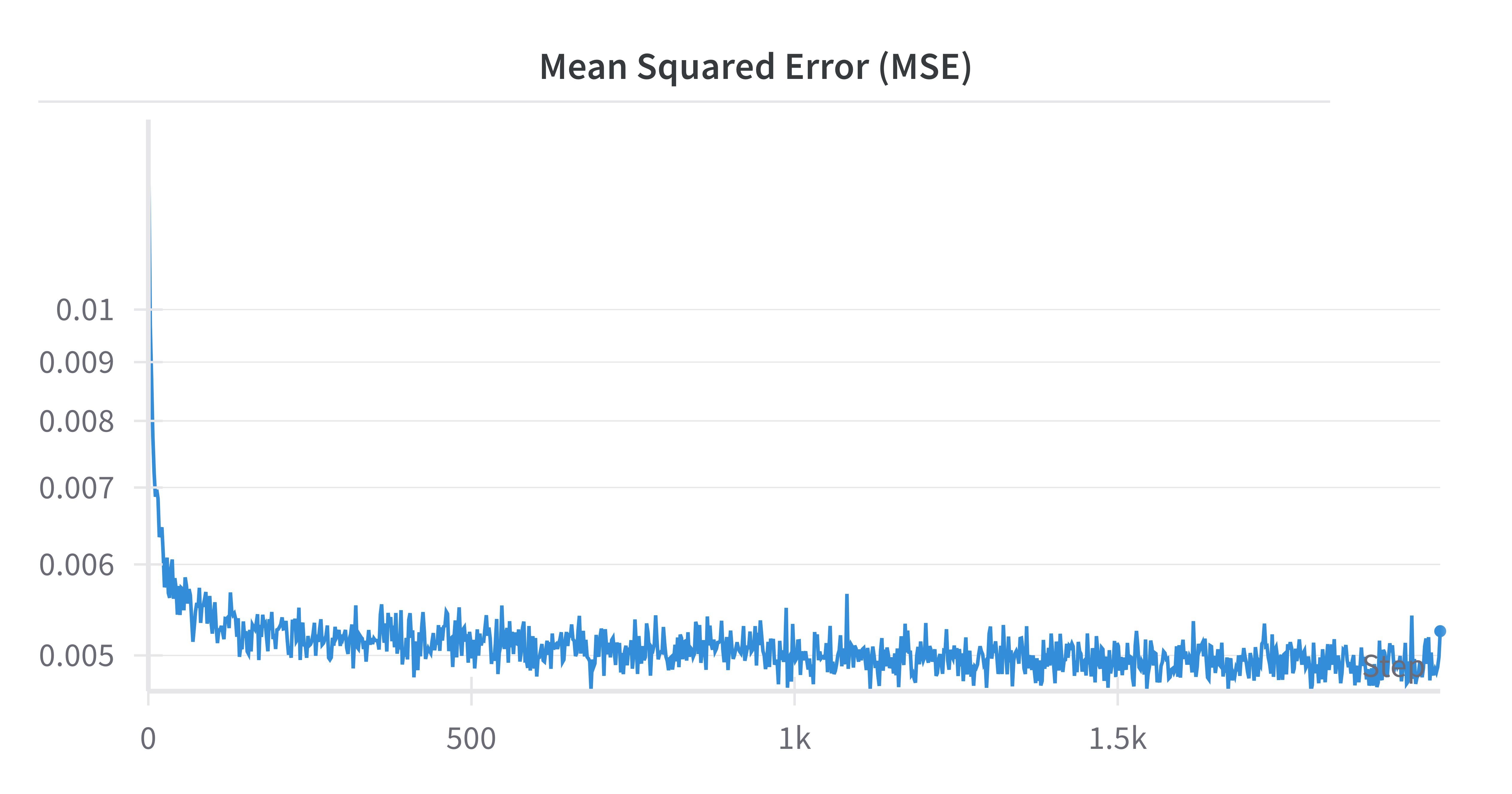}
    \includegraphics[width=0.48\linewidth]{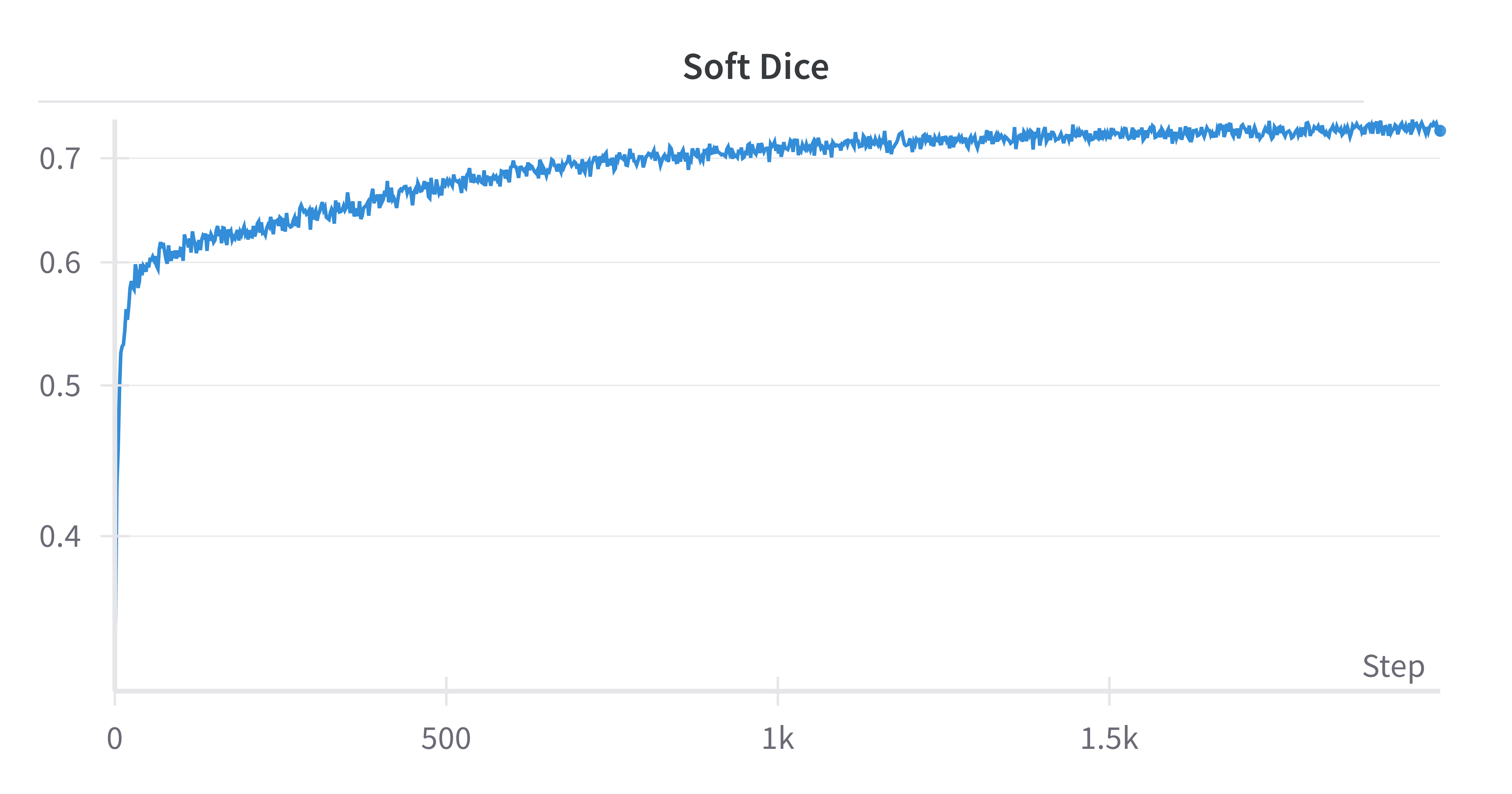}
    \caption{\textbf{Verifying convergence of KeyMorph}. We verify the convergence of KeyMorph (with dice loss) on the OASIS dataset by plotting the Mean Squared Error (left) and Soft Dice (right) on the training set.}
    \label{fig:km-convergence}
\end{figure*}

We run KeyMorph~\cite{wang2023robust} on the OASIS dataset for 2000 epochs. We plot the Soft Dice ($= 1 - \text{diceloss}$) and Mean Squared error between the fixed and moving images in \cref{fig:km-convergence}.
Note that the soft Dice loss starts to plateau at $\sim 0.70$, and the hard dice loss on the validation set is even lower ($\sim 0.64$).
This represents a huge gap in performance compared to unsupervised baselines and our method. %
These numbers are also consistent with those reported in ~\cite{wang2023robust} for deformable registration.
Note that although KeyMorph works in the contrived scenario of arbitrary rotations and translations (most MRI datasets are acquired in standard coordinate systems like RAS), it is not designed to handle the more complex deformations that are present in the brain MRI datasets. 

\end{document}